\documentclass{article}

\usepackage{microtype}
\usepackage{graphicx}
\usepackage{subfigure}
\usepackage{booktabs} %

\usepackage[accepted]{icml2024}

\definecolor{Gray}{gray}{0.9}
\definecolor{mygreen}{rgb}{0.0, 0.5, 0.0}
\definecolor{myred}{rgb}{0.8, 0.25, 0.33}
\definecolor{myblue}{rgb}{0.19, 0.55, 0.91}
\definecolor{uclablue}{rgb}{0.15, 0.45, 0.68}
\definecolor{ucladblue}{rgb}{0.0, 0.33, 0.53}
\definecolor{ucladdblue}{rgb}{0.0, 0.23, 0.36}
\definecolor{uclagold}{rgb}{1.0, 0.82, 0.0}
\definecolor{ucladgold}{rgb}{1.0, 0.78, 0.17}
\definecolor{ucladdgold}{rgb}{1.0, 0.72, 0.11}
\definecolor{boxgreen}{rgb}{0.02, 0.66, 0.02}
\definecolor{boxred}{rgb}{0.66, 0.1, 0.1}
\definecolor{boxblue}{rgb}{0.01, 0.01, 0.73}

\usepackage{etoc}
\usepackage{lipsum}
\usepackage{wrapfig}
\usepackage{multicol}
\usepackage{mdframed}

\usepackage{mathtools}
\usepackage{amsthm}

\usepackage[utf8]{inputenc}   %
\usepackage[T1]{fontenc}      %
\usepackage{url}              %
\usepackage{amsfonts}         %
\usepackage{nicefrac}         %
\usepackage{microtype}
\usepackage{times}
\usepackage{graphicx}
\usepackage{amsmath,amssymb,mathbbol}
\usepackage{acronym}
\usepackage{enumitem}
\usepackage{balance}
\usepackage{xspace}
\usepackage{setspace}
\usepackage[skip=3pt,font=small]{caption}
\usepackage{booktabs,tabularx,colortbl,multirow,array,makecell}
\usepackage{pifont}
\usepackage{xr-hyper}
\usepackage{tcolorbox}
\tcbuselibrary{breakable}

\usepackage[pagebackref,breaklinks,citecolor=ucladblue,colorlinks=true,linkcolor=ucladblue,bookmarks=false]{hyperref}
\usepackage[capitalise]{cleveref}

\usepackage{siunitx}   %
\usepackage{soul}   %
\usepackage{listings}
\definecolor{codegreen}{rgb}{0,0.6,0}
\definecolor{codegray}{rgb}{0.5,0.5,0.5}
\definecolor{codepurple}{rgb}{0.58,0,0.82}
\definecolor{backcolour}{rgb}{0.95,0.95,0.92}

\lstdefinestyle{mystyle}{
    backgroundcolor=\color{backcolour},   
    commentstyle=\color{codegreen},
    keywordstyle=\color{magenta},
    numberstyle=\tiny\color{codegray},
    stringstyle=\color{codepurple},
    basicstyle=\ttfamily\footnotesize,
    breakatwhitespace=false,         
    breaklines=true,                 
    captionpos=b,                    
    keepspaces=true,                 
    numbers=left,                    
    numbersep=5pt,                  
    showspaces=false,                
    showstringspaces=false,
    showtabs=false,                  
    tabsize=2
}

\makeatletter
\DeclareRobustCommand\onedot{\futurelet\@let@token\@onedot}
\def\@onedot{\ifx\@let@token.\else.\null\fi\xspace}
\def\eg{\emph{e.g}\onedot} 

\def\ie{\emph{i.e}\onedot}

\def\etc{\emph{etc}\onedot} 
\def\vs{\emph{vs}\onedot}

\makeatother

\makeatletter
\renewcommand{\paragraph}{%
  \@startsection{paragraph}{4}%
  {\z@}{0ex \@plus 0ex \@minus 0ex}{-1em}%
  {\hskip\parindent\normalfont\normalsize\bfseries}%
}
\makeatother

\crefname{algorithm}{Alg.}{Algs.}
\Crefname{algocf}{Algorithm}{Algorithms}
\crefname{section}{Sec.}{Secs.}
\Crefname{section}{Section}{Sections}
\crefname{table}{Tab.}{Tabs.}
\Crefname{table}{Table}{Tables}
\crefname{figure}{Fig.}{Fig.}
\Crefname{figure}{Figure}{Figure}

\definecolor{gblue}{HTML}{4285F4}
\definecolor{gred}{HTML}{DB4437}
\definecolor{ggreen}{HTML}{0F9D58}

\definecolor{mygray}{gray}{.92}

\newcommand{\sota}{state-of-the-art\xspace}

\newcommand{\cmark}{\ding{51}}
\newcommand{\xmark}{\ding{55}}

\newcommand{\agent}{\textsc{LEO}\xspace}

\acrodef{llm}[LLMs]{large language models}
\acrodef{vla}[VLA]{vision-language-action}
\acrodef{vl}[VL]{vision-language}
\acrodef{ocot}[O-CoT]{Object-centric Chain-of-Thought}
\acrodef{cot}[CoT]{Chain-of-Thought}
\acrodef{lvlm}[LVLM]{large vision-language models}

\definecolor{color1}{rgb}{0.94,0.95,0.95}
\definecolor{color2}{rgb}{0.92,1.0,0.98}
\definecolor{color3}{rgb}{0.89,0.90,0.98}
\definecolor{color4}{rgb}{0.9,0.96,1.0}

\icmltitlerunning{An Embodied Generalist Agent in 3D World}

\begin{document}

\twocolumn[
\icmltitle{An Embodied Generalist Agent in 3D World}

\icmlsetsymbol{equal}{*}
\icmlsetsymbol{lead}{†}

\begin{icmlauthorlist}
\icmlauthor{Jiangyong Huang}{equal,bigai,pku}
\icmlauthor{Silong Yong}{equal,bigai,thu}
\icmlauthor{Xiaojian Ma}{equal,lead,bigai}
\icmlauthor{Xiongkun Linghu}{equal,bigai}
\icmlauthor{Puhao Li}{bigai,thu} \\
\icmlauthor{Yan Wang}{bigai}
\icmlauthor{Qing Li}{bigai}
\icmlauthor{Song-Chun Zhu}{bigai,pku,thu}
\icmlauthor{Baoxiong Jia}{bigai}
\icmlauthor{Siyuan Huang}{lead,bigai}
\end{icmlauthorlist}

\icmlaffiliation{bigai}{State Key Laboratory of General Artificial Intelligence, Beijing Institute for General Artificial Intelligence (BIGAI)}
\icmlaffiliation{pku}{Peking University}
% \icmlaffiliation{cmu}{Carnegie Mellon University}
\icmlaffiliation{thu}{Tsinghua University}

\icmlkeywords{Embodied Generalist Agent, 3D Vision-Language, Grounded 3D Scene Understanding, ICML}

\vskip 0.3in
]

\printAffiliationsAndNotice{\icmlEqualContribution \textsuperscript{†}Research lead} %

\begin{abstract}
Leveraging massive knowledge from \ac{llm}, recent machine learning models show notable successes in general-purpose task solving in diverse domains such as computer vision and robotics. However, several significant challenges remain: (i) most of these models rely on 2D images yet exhibit a limited capacity for 3D input; (ii) these models rarely explore the tasks inherently defined in 3D world, \eg, 3D grounding, embodied reasoning and acting.
We argue these limitations significantly hinder current models from performing real-world tasks and approaching general intelligence. To this end, we introduce \agent, an embodied multi-modal generalist agent that excels in perceiving, grounding, reasoning, planning, and acting in the 3D world. \agent is trained with a unified task interface, model architecture, and objective in two stages: (i) 3D vision-language (VL) alignment and (ii) 3D \ac{vla} instruction tuning. We collect large-scale datasets comprising diverse object-level and scene-level tasks, which require considerable understanding of and interaction with the 3D world. Moreover, we meticulously design an LLM-assisted pipeline to produce high-quality 3D VL data. Through extensive experiments, we demonstrate \agent's remarkable proficiency across a wide spectrum of tasks, including 3D captioning, question answering, embodied reasoning, navigation and manipulation. Our ablative studies and scaling analyses further provide valuable insights for developing future embodied generalist agents. Code and data are available on \href{https://embodied-generalist.github.io/}{project page}.
\end{abstract}
\section{Introduction}\label{sec:intro}

Building one generalist model that can handle comprehensive tasks like humans has been a long-existing pursuit in artificial intelligence and neuroscience~\citep{lake2015human,lake2017building,zhu2020dark, mountcastle1979organizing,schmidhuber2018one,huang2022perceive}. Recent advances in \ac{llm}~\citep{brown2020language} and ``foundation models''~\citep{bommasani2021opportunities} emerge as a promising paradigm in building such generalist models in natural language processing~\citep{openai2022chatgpt,openai2023gpt4}, computer vision~\citep{kirillov2023segment,alayrac2022flamingo}, and robotics~\citep{brohan2022rt,brohan2023rt,reed2022generalist,driess2023palm,li2023multimodal}. The keys to the success of this paradigm lie in large-scale internet-level datasets from numerous tasks and domains, as well as scalable Transformer architectures~\citep{vaswani2017attention} that can absorb generalizable and task-agnostic knowledge from the data. 
Nonetheless, existing generalist models primarily thrive within 2D domains, lacking comprehension of the 3D physical environment that envelops human-level intelligence. This limitation stands as an obstacle that prevents current models from solving real-world tasks and approaching general intelligence. Therefore, we ask a fundamental question: \textit{how to equip the generalist agent with a comprehensive understanding of and the ability to interact with the real 3D world}?

\begin{figure*}[t!]
\centering
\includegraphics[width=\textwidth]{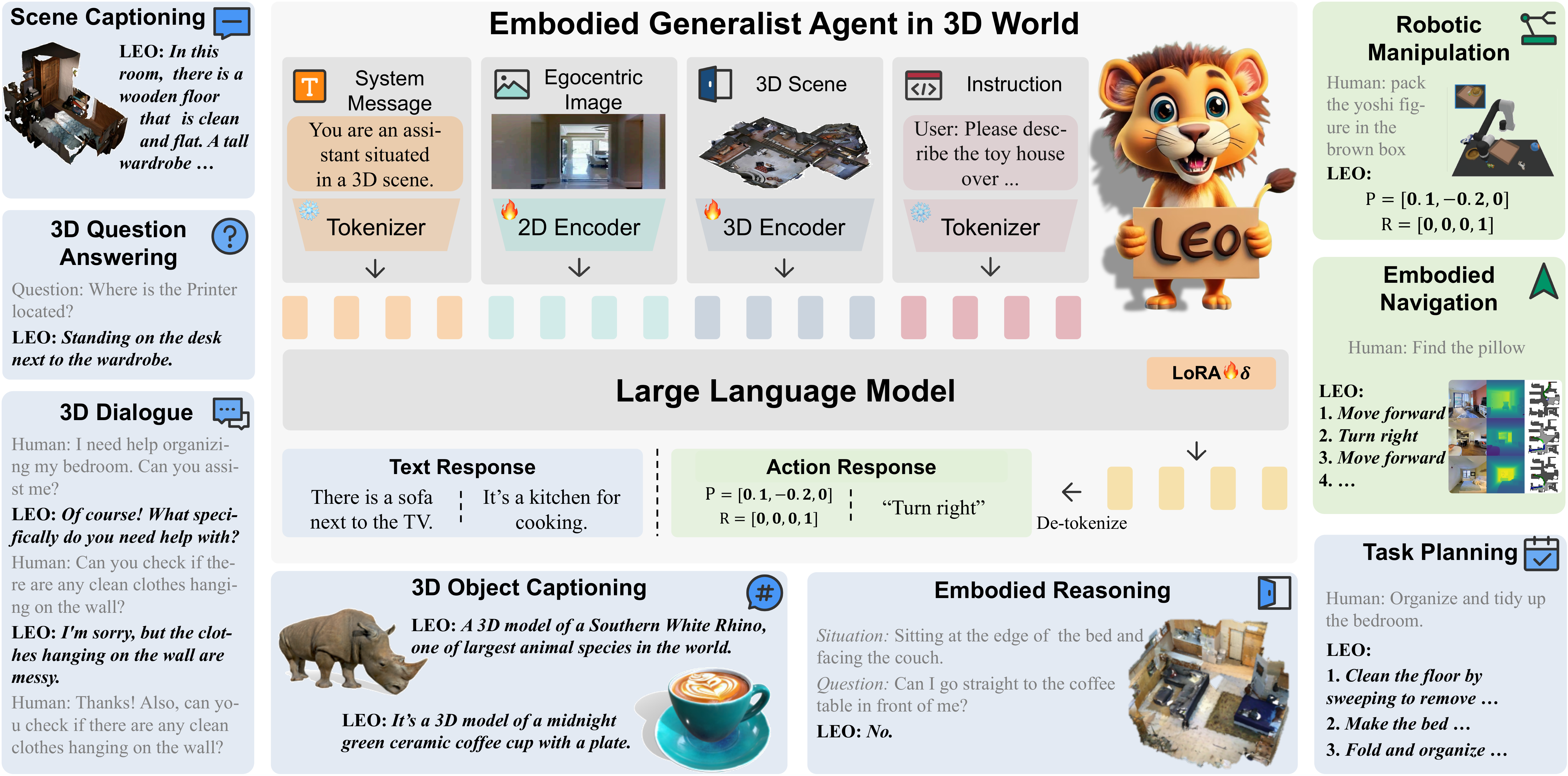}%
  \caption{\textbf{The proposed embodied generalist agent \agent}. It takes egocentric 2D images, 3D point clouds, and texts as input and formulates comprehensive 3D tasks as autoregressive sequence prediction. By instruction-tuning \agent, it extends the capability of \ac{llm} to multi-modal vision-language-action tasks with a unified model.}
  \label{fig:leo}
  \vskip -0.05in
\end{figure*}

The development of such generalist agents encounters three primary challenges: the lack of suitable datasets, unified models, and effective learning strategies. Despite substantial progress in scaling up image-text models~\citep{tsimpoukelli2021multimodal,alayrac2022flamingo} and the curation of corresponding datasets~\citep{radford2021learning,schuhmann2022laion}, advancement in 3D scene-level understanding has significantly lagged behind. This is largely attributed to the limited scale and manual labeling of 3D datasets \citep{dai2017scannet,wald2019rio,chen2020scanrefer}, given the higher cost associated with collecting 3D data compared to 2D data. Furthermore, large-scale unified pretraining and efficient finetuning are under-explored by previous 3D VL models, which are often designed with strong priors \citep{zhao20213dvg,chen2022language}. Notably, recent works \citep{zhu20233d,hong20233d} utilize multi-modal Transformer together with synthetic data to enhance the model's capability in grounded 3D scene understanding. Nevertheless, they fall short in embodied tasks, \eg, acting within 3D environments. Additionally, there are significant yet rarely explored problems, \eg, the potential of \ac{vla} learning and efficient adaptation of \ac{llm} for 3D tasks.

In this work, we introduce the generalist agent \agent, which is generically embodied, multi-modal, and general-purpose. It can take egocentric 2D images, 3D point clouds, and texts as task input and handle comprehensive tasks within the 3D environment. As shown in \cref{fig:leo}, \textit{\agent exhibits the capability of perceiving, grounding, reasoning, planning, and acting with a unified task interface, model architecture, and objective.} \agent perceives through an egocentric 2D image encoder for the embodied view and an object-centric 3D point cloud encoder for the third-person global view. Such perception modules can be flexibly adapted to various embodied environments and enhance 3D reasoning. The encoded visual tokens are interleaved with text tokens to form a unified multi-modal task sequence, which further serves as the input to a decoder-only LLM. Equipped with a vocabulary containing both text and action tokens, the LLM can generate responses to various tasks simultaneously. Consequently, all the tasks are formulated as sequence prediction, thereby accommodating a unified training objective.

Following prior experiences \citep{liu2023visual}, we adopt a two-stage learning scheme, \ie, 3D VL alignment and 3D \ac{vla} instruction tuning. We accordingly collect large-scale comprehensive datasets \agent-align and \agent-instruct, which comprise diverse object-level and scene-level tasks. Notably, we meticulously design an LLM-assisted pipeline to generate high-quality 3D VL data, wherein we propose to prompt \ac{llm} \citep{openai2022chatgpt} with scene graphs and \ac{ocot} method. To further enhance quality control, we devise a series of refinement procedures via regular expression matching and scene graph retrieval. We demonstrate our approach largely enriches the data scale and diversity, meanwhile mitigating hallucination in LLM-generated data.

We quantitatively evaluate and ablate \agent on diverse 3D tasks, including 3D captioning~\citep{chen2021scan2cap}, 3D question answering~\citep{azuma2022scanqa}, situated question answering~\citep{ma2023sqa3d}, embodied navigation~\citep{ramrakhya2022habitat}, and robotic manipulation~\citep{cliport}. The results indicate (i) through task-agnostic instruction tuning with a unified model, \agent achieves state-of-the-art performances on most tasks, particularly surpassing previous task-specific models; (ii) \agent shows proficiency in scene-grounded dialogue and planning, capable of generating flexible and coherent responses; (iii) \agent achieves comparable performances to \sota task-specific models on navigation and manipulation tasks, and exhibits remarkable generalization ability; (iv) \agent's strong performances stem from both data and model aspects, including the alignment stage, data diversity, generalist-style instruction tuning, and object-centric representation; (v) \agent manifests the scaling law that echoes prior findings~\citep{kaplan2020scaling,reed2022generalist,openai2023gpt4}. We also present qualitative results to illustrate \agent's versatility and proficiency in grounded 3D scene understanding.

In summary, our main contributions are as follows: (i) we propose \agent, the first embodied generalist agent capable of following human instructions to perceive, ground, reason, plan, and act in the 3D world; (ii) we propose a simple yet effective framework that connects object-centric 3D representation and LLM to efficiently bridge the gap between vision, language, and embodied action; (iii) we collect large-scale comprehensive datasets for our two-stage generalist training scheme, and particularly propose an LLM-assisted pipeline for the generation of high-quality 3D VL data; (iv) we conduct extensive experiments to demonstrate \agent's proficiency across various tasks, and present in-depth analyses to reveal valuable insights; (v) we release the data, code, and model weights to endow the future research in embodied generalist agents.

\section{Model}\label{sec:model}%
The leading design principles of \agent are two-fold: 1) It should handle the multi-modal input of egocentric 2D, global 3D, and textual instruction, and the output of textual response as well as embodied action commands in a unified architecture; 2) It should leverage pre-trained large language models (LLMs) as a powerful prior for the downstream tasks. We therefore convert all data of different modalities into a sequence of tokens, illustrated below:
\vskip -0.3in
\begin{equation}\label{equ:data}
\begin{split}
    &\underbrace{\text{\texttt{You are...}}}_{\text{system message}}~\underbrace{s_{\text{2D}}^{(1)},..., s_{\text{2D}}^{(M)}}_{\substack{\text{2D image tokens}\\ \text{(optional)}}}\underbrace{s_{\text{3D}}^{(1)},..., s_{\text{3D}}^{(N)}}_{\substack{\text{object-centric}\\\text{3D tokens}}},\\
    &\underbrace{\text{\texttt{USER:... ASSISTANT:}}}_{\text{instruction}}~\underbrace{s_{\text{res}}^{(1)},...s_{\text{res}}^{(T)}}_{\text{response}}.
\end{split}
\end{equation}
\vskip -0.2in
With this representation, we formulate the learning of \agent as GPT-style autoregressive language modeling~\citep{brown2020language} given the \textit{prefix} (from \textit{system message} to \textit{instruction}), \ie prefix language modeling~\citep{raffel2020exploring}. Therefore, a pretrained LLM can be used to process such sequences. Next, we will detail the tokenization of multi-modal data, model architecture, training loss, and inference settings. An overview of our model can be found in \cref{fig:leo}.
\vskip -0.5in

\subsection{Tokenization}\label{sec:model_tokenization}

We follow prior practices in 2D VLM~\citep{liu2023visual,alayrac2022flamingo} and 3D VLM~\citep{zhu20233d} to tokenize the multi-modal data in \agent. 
We use SentencePiece tokenizer~\citep{kudo2018sentencepiece} to encode text with 32k subwords; 2D image tokens for egocentric 2D images; and object-centric 3D tokens extracted over Mask3D-based~\citep{schult2022mask3d} object proposals for 3D point cloud inputs.
For embodied action commands, continuous actions (\eg in manipulation) are discretized (details in \cref{sec:action_tokenization}) to join the discrete actions (\eg navigation) and form a unified discrete action space. We follow~\cite{brohan2023rt} to map these discrete actions to the least used tokens in SentencePiece.
After tokenization, all tokens are ordered into the format in (\ref{equ:data}).

\begin{table*}[t!]
\begin{minipage}{0.65\linewidth}
    \captionof{table}{\label{tab:data_stat}\textbf{Datasets statistics}. We illustrate key statistics of datasets for 3D VL alignment (\agent-align) and 3D VLA instruction tuning (\agent-instruct). \textit{res.} (response) denotes tokens to be predicted, while \textit{prefix} denotes those in the context.} 
    \centering
    \small
    \setlength\tabcolsep{3pt}
    \resizebox{1.0\linewidth}{!}{
    \begin{tabular}{ccccccc}
    \toprule
         Dataset & Task & 2D input & 3D assets & \#data & \makecell{\#token\\(\textit{res.})} & \makecell{\#token\\(\textit{prefix}+\textit{res.})} \\ 
         \midrule
         \multirow{3}{*}{\agent-align} & object captioning & \xmark & Objaverse & 660K & 10M & 27M \\
         & object referring & \xmark & ScanNet + 3RScan &  354K & 15M & 39M  \\
         & scene captioning & \xmark & 3RScan & 20K & 3.3M & 4.4M \\
         \midrule
        \multirow{6}{*}{\agent-instruct} & 3D captioning & \xmark & ScanNet & 37K & 821K & 3M  \\
         & 3D QA & \xmark & ScanNet + 3RScan & 83K & 177K & 4M \\
         & 3D dialogue & \xmark & 3RScan & 11K & 1.1M & 8.3M \\
         & task planning & \xmark & 3RScan & 14K & 1.9M & 2.7M  \\
         & navigation & \cmark & MP3D & 60K & 11.4M & 272M \\
         & manipulation & \cmark & CLIPort & 300K & 7.2M & 734M \\
         \bottomrule
    \end{tabular}
    }
\end{minipage}
\hfill
\begin{minipage}{0.33\linewidth}
    \centering
    \small
    \captionof{table}{Answer accuracy of LLM-generated data on three types of questions.}\label{tab:data_quality_comparison}
    \setlength\tabcolsep{3pt}
    \resizebox{1.0\linewidth}{!}{
    \begin{tabular}{lccc}
        \toprule
         & Counting & Existence & Non-existence \\
        \midrule
        3D-LLM & 56.5 & 96.8 & 40.0 \\
        \midrule
        Ours & 57.4 & 91.3 & 27.4 \\
        + O-CoT & 78.0 & 93.4 & 30.5 \\
        + refinement & 100.0 & 100.0 & 100.0 \\
        \bottomrule
    \end{tabular}
    }
    \captionof{table}{The amount of examined data in \cref{tab:data_quality_comparison}. 3D-LLM data \citep{hong20233d} is much less since we can only access a subset.}\label{tab:data_quality_statistics}
    \setlength\tabcolsep{3pt}
    \resizebox{1.0\linewidth}{!}{
    \begin{tabular}{lccc}
        \toprule
         & Counting & Existence & Non-existence \\
        \midrule
        3D-LLM & 434 & 95 & 10 \\
        Ours & 2666 & 6766 & 3314 \\
        \bottomrule
    \end{tabular}
    }
\end{minipage}
\vspace{-0.28em}
\end{table*}

\subsection{Token Embedding \& LLM}

We apply several token embedding functions to process the tokens in the sequence before sending them to the LLM. The LLM will then align these tokens of different modalities, and produce the response. Most of the responses are text and can be decoded directly. For responses that include embodied actions, we will map the reserved SentencePiece text tokens back to action commands.

\paragraph{Text \& 2D token embedding.}  For text tokens (including embodied actions that have been mapped to the reserved text tokens), an embedding look-up table is used to map them into vectors. While the egocentric 2D image is encoded by a pretrained OpenCLIP ConvNext~\citep{liu2022convnet} for obtaining image token embeddings.
We apply MLP adapters to match the dimensions of all token embeddings.

\paragraph{Object-centric 3D token embedding.} Each 3D object token (\ie, the point cloud of a 3D object) is first encoded by a pretrained point cloud encoder (\eg, PointNet++~\citep{qi2017pointnet++}). We then adopt the Spatial Transformer introduced in~\cite{chen2022language} to further process the point cloud embedding of all objects into object-centric 3D token embeddings. 
In a nutshell, Spatial Transformer biases the standard attention score with relative position and size for capturing 3D relations between objects. 
Due to space limit, the readers are referred to~\cite{chen2022language} and \cref{sec:supp_embedding} for more details. %

\paragraph{Pretrained LLM.} We choose Vicuna-7B~\citep{vicuna2023}
to process the token sequence. In order to tackle the challenging alignment and grounding problem of multi-modal tokens (2D, 3D, text, embodied action) while preserving the LLM pretrained knowledge, we employ LoRA~\citep{hu2022lora} to introduce additional tunable parameters to the frozen pretrained LLM. 

\subsection{Training \& Inference}
We formulate the learning objective of \agent following~\citep{brown2020language,raffel2020exploring} in a prefix language modeling fashion. For a batch $\mathcal{B}$ of token sequence $s$, we optimize \agent via:
\vskip -0.3in
\begin{align}
   \mathcal{L}(\theta, \mathcal{B}) = -\sum^{|\mathcal{B}|}_{b=1}\sum^{T}_{t=1}\log p_{\theta}(s_{\text{res}}^{(b,t)}|s_{\text{res}}^{(b,<t)},s_{\text{prefix}}^{(b)}), 
\end{align}
\vskip -0.15in
where $s_\text{prefix}$ denotes the prefix tokens (from \textit{system message} to \textit{instruction}) in (\ref{equ:data}). 
During training, we freeze the pretrained 3D point cloud encoder and the LLM and finetune the 2D image encoder, the Spatial Transformer, and the LoRA parameters.
In total, \agent has \textasciitilde{}7B parameters and \textasciitilde{}142M of them will be tuned. 
During inference, we use beam search to generate textual responses. For tasks that require action commands, we map the textual outputs to action commands as discussed in \cref{sec:model_tokenization}. More details on the model and training can be found in \cref{app:model}.

\section{Datasets}\label{sec:data}

Since \agent is a generalist agent that receives multi-modal inputs and follows instructions, we adopt the two-stage training proposed by \citet{liu2023visual} and split the data into two sets: (i) \agent-align (\cref{sec:data:align}) that focuses on \textbf{3D \ac{vl} alignment} to bridge the gap between 3D scene representation and natural language; and (ii) \agent-instruct (\cref{sec:sft}) that targets at \textbf{3D VLA instruction tuning} to endow \agent with various capabilities. The statistics and examples of these datasets can be found in \cref{tab:data_stat} and \cref{sec:supp_leo_ds_examples}, respectively. Due to the data scarcity, we adopt LLMs to facilitate the data generation process and outline the details in \cref{sec:data:generation}.

\subsection{\agent-align: 3D Vision-Language Alignment}\label{sec:data:align}
In \agent-align, we focus on 3D \ac{vl} alignment. Similar to BLIP-2~\citep{li2023blip}, we train \agent to generate captions given various 3D inputs. Specifically, we collect three types of 3D captioning data: 1) \textbf{object-level captions}, where we align 3D individual objects with their descriptions \citep{luo2023scalable}; 2) \textbf{object-in-the-scene captions}, where the goal is to generate the referring expressions of objects in a 3D scene context \citep{achlioptas2020referit3d,zhu20233d}; and 3) \textbf{scene-level captions}, which focuses on depicting global 3D scene using natural language. Due to the space limit, we defer details including data source and components to \cref{app:dataset:leo_align}.

\begin{figure*}[t]
\centering
\includegraphics[width=\textwidth, keepaspectratio]{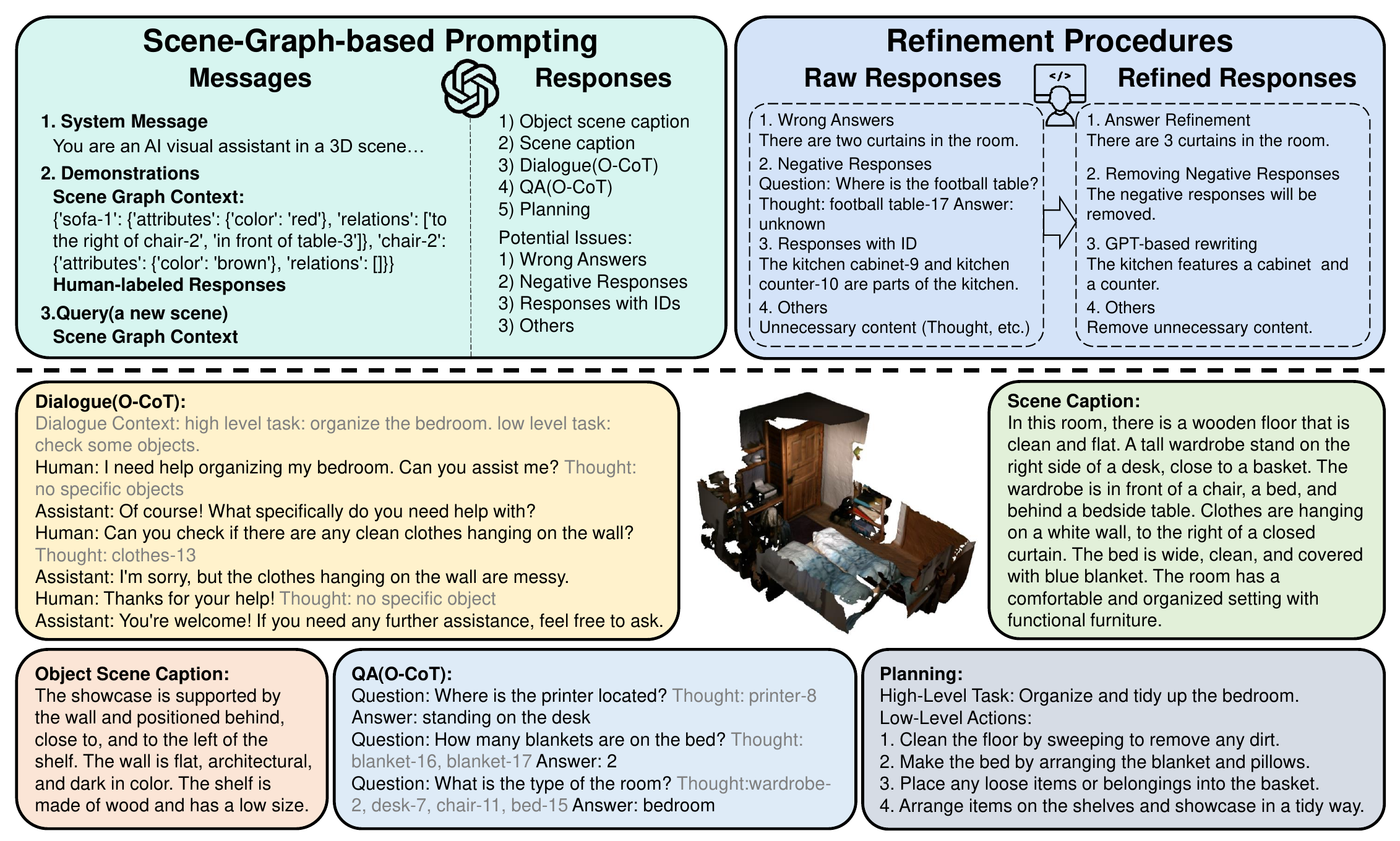}%
  \caption{\textbf{Our proposed LLM-assisted 3D-language data generation pipeline and data examples.}. (Top-left) Messages with 3D scene graphs,
  including object attributes and relations in a phrasal form,
  used for providing scene context when prompting LLM.
  (Top-right) The human-defined refinement procedures were conducted over raw LLM responses to improve data quality.
  (Bottom) Examples of LLM-assisted generation in \agent-align and \agent-instruct. \textcolor{gray}{Thoughts, colored in gray, will be removed after refinements}.}
  \label{fig:data_framework}
  \vskip -0.15in
\end{figure*}

\subsection{\agent-instruct: Instruction Following in 3D world}\label{sec:sft}
In \agent-instruct, \agent will be tuned to follow instructions and accomplish various 3D \ac{vla} tasks. We curate a comprehensive set of tasks that covers a broad spectrum from grounded scene understanding and reasoning~\citep{chen2021scan2cap, ma2023sqa3d}, to dialogue, planning, and embodied acting~\citep{savva2019habitat,cliport}. Specifically, we introduce 1) \textbf{3D captioning and question answering} -- given 3D scene input, the agent needs to generate a natural language response to describe the scene or answer questions; 2) \textbf{3D dialogue and task planning}, where the agent is expected to generate flexible and coherent responses to complex instructions with respect to the given 3D scene, and 3) \textbf{navigation and manipulation}, which require the agent to accomplish a variety of embodied acting tasks in the 3D scene. We defer details to \cref{app:dataset:leo_instruct}.

\subsection{LLM-assisted 3D-language Data Generation}\label{sec:data:generation}
As mentioned above, at the core of producing a large proportion of \agent-align and \agent-instruct is the assistance of LLMs. We now detail the key techniques of prompting LLMs (\ie, ChatGPT) to generate 3D-text paired data. An overview can be found in \cref{fig:data_framework}.

\paragraph{Scene-graph-based prompting.} Our data generation pipeline starts with 3D scene graphs from 3DSSG~\citep{wu2021scenegraphfusion}, which provide scene contexts for prompting. 
Compared to counterparts that utilize object boxes~\citep{yin2023lamm,hong20233d,wang2023chat}, it offers both rich object attributes and accurate spatial relation information among objects, allowing \ac{llm} to generate data with high-quality 3D details (comparisons in \cref{sec:scene graph prompting and bbox prompting}). Next, we manually design some examples as seed tasks~\cite{liu2023visual}, including scene and object captioning, QA, dialogue, and planning, and ask LLM to produce more tasks as well as the responses. Details for designing the seed tasks can be found in~\cref{app:dataset:seed_task}.

\paragraph{Object-centric CoT.} To further combat the \textbf{hallucination} of \ac{llm}~\citep{bang2023hallucination} in open-ended generation as in our pipeline, we propose the object-centric chain of thought (\ac{ocot}) prompting that requires the LLM to explicitly provide the label and ID of object candidates as \textcolor{gray}{thoughts} during text generation. We also utilize subgraph sampling to further enhance the diversity of 3D scene graphs (see details in \cref{app:subgraph_sampling}). We provide examples of \ac{ocot} in~\cref{fig:data_framework}.

\paragraph{Refinement procedures.} Upon the scene graph and O-CoT prompting, we introduce refinement as an additional safeguard to the quality and reliability of our generated data. Specifically, we send raw LLM responses to several human-defined filters based on the 3D scene graphs: negative responses (\eg, lacking the necessary information to answer) will be removed; unnatural narratives will be rewritten, \etc. Further, we detect text that involves logical reasoning (\eg, counting) or hallucination, and manually fix the wrong responses according to the ground truth provided by scene graphs. We provide illustrative examples in \cref{fig:data_framework} and \cref{app:dataset:refine:examples}, and quantitative analysis on the impact of data refinement procedures in \cref{sec:impact_data_refinement}. %

\paragraph{Assess the quality of generated data.} In addition to data examples, we propose to assess the quality of generated data quantitatively. We focus on the LLM-produced question-answer pairs about objects (questions starting with \textit{How many/Is there} and ending with \textit{in the room/bedroom/kitchen/living room/bathroom}). We first divide these pairs into three categories: \textit{counting}, \textit{existence}, and \textit{non-existence}, which examines the number of certain objects/whether an object exists/whether an object does not exist in the scene, respectively. We manually check if the answers in these pairs are correct, and report the overall accuracy. Results in \cref{tab:data_quality_comparison} demonstrate that our proposed scene-graph-based prompting, O-CoT prompting and refinement bring consistent improvement to data quality and the complete data generation pipeline outperforms a recent counterpart (3D-LLM). We also demonstrate how we help address the \textbf{grammatical errors} compared to counterparts in \cref{app:additional_data_comparison}. Finally, we provide the data distribution in \cref{app:dataset statistics} to illustrate the \textbf{diversity} of our generated data.

\begin{table*}
\begin{minipage}{0.635\linewidth}
\centering
\captionof{table}{\textbf{Quantitative comparison with \sota models on 3D VL understanding and embodied reasoning tasks}. ``C'' stands for ``CIDEr'', ``B-4'' for ``BLEU-4'', ``M'' for ``METEOR'', ``R'' for ``ROUGE'', ``Sim'' for sentence similarity, and ``EM@1'' for top-1 exact match. The n-gram metrics for Scan2Cap are governed by IoU@0.5. $^\dagger$ indicates answering questions via prompting GPT-3 with the generated scene caption. \textcolor{gray}{Gray} indicates evaluation results with refined exact-match protocol.}
\vspace{0.1em}
\resizebox{\linewidth}{!}{
\begin{tabular}{lccccccccccc}
    \toprule
     & \multicolumn{5}{c}{Scan2Cap (val)} & \multicolumn{5}{c}{ScanQA (val)} & SQA3D (test) \\
     \cmidrule(lr){2-6} \cmidrule(lr){7-11} \cmidrule(lr){12-12}
     & C & B-4 & M & R & Sim & C & B-4 & M & R & EM@1 & EM@1 \\
    \midrule
    \multicolumn{1}{l}{\small\textbf{\textit{Task-specific models}}} \\
    Scan2Cap & 35.2 & 22.4 & 21.4 & 43.5 & - & - & - & - & - & - & \hspace{2pt} 41.0$^\dagger$ \\
    3DJCG & 47.7 & 31.5 & 24.3 & 51.8 & - & - & - & - & - & - & - \\
    Vote2Cap-DETR & 61.8 & 34.5 & 26.2 & 54.4 & - & - & - & - & - & - & - \\
    ScanRefer+MCAN & - & - & - & - & - & 55.4 & 7.9 & 11.5 & 30.0 & 18.6 & - \\
    ClipBERT & - & - & - & - & - & - & - & - & - & - & 43.3 \\
    ScanQA & - & - & - & - & - & 64.9 & 10.1 & 13.1 & 33.3 & 21.1 & 47.2 \\
    \midrule
    \multicolumn{1}{l}{\small\textit{\textbf{Task-specific fine-tuned}}} \\
    3D-VisTA & 66.9 & 34.0 & 27.1 & 54.3 & 53.8 & 69.6 & 10.4 & 13.9 & 35.7 & 22.4 & 48.5 \\
    3D-LLM (FlanT5) & - & - & - & - & - & 69.4 & 12.0 & 14.5 & 35.7 & 20.5 & - \\
    \midrule

    \agent & \textbf{72.4} & \textbf{38.2} & \textbf{27.9} & \textbf{58.1} & \textbf{55.3} & \textbf{101.4} & \textbf{13.2} & \textbf{20.0} & \textbf{49.2} & \textbf{24.5 \textcolor{gray}{(47.6)}} & \textbf{50.0 \textcolor{gray}{(52.4)}} \\

    \bottomrule
\end{tabular}
}
\label{tab:vl_results}
\end{minipage}
\hfill
\begin{minipage}{0.345\linewidth}
    \vspace{-0.05em}
    \centering   
    \small
    \captionof{table}{\label{tab:test_result_act_cliport} \textbf{Results on robot manipulation}. {\color{mygreen}{seen}} indicates in-domain tasks. {\color{myred}{unseen}} marks OOD tasks with novel colors or objects.}
    \vspace{-0.25em}
    \setlength\tabcolsep{2pt}
    \resizebox{\linewidth}{!}{
    \begin{tabular}{llllccccccccccccccc}
    \toprule
    \multicolumn{4}{l}{} & \multicolumn{4}{c}{\small{separating-piles}} & \multicolumn{4}{c}{\small{\makecell{packing-google\\-objects-seq}}} & \multicolumn{4}{c}{\small{\makecell{put-blocks-in\\-bowls}}} \\
    \cmidrule(lr){5-8}\cmidrule(lr){9-12}\cmidrule(lr){13-16}
    \multicolumn{4}{c}{}& \multicolumn{2}{c}{\color{mygreen}{seen}} & \multicolumn{2}{c}{\color{myred}{unseen}} & \multicolumn{2}{c}{\color{mygreen}{seen}} & \multicolumn{2}{c}{\color{myred}{unseen}} & \multicolumn{2}{c}{\color{mygreen}{seen}} & \multicolumn{2}{c}{\color{myred}{unseen}} \\
    
    \midrule
    
    \multicolumn{4}{l}{CLIP-only}        & \multicolumn{2}{c}{90.2} & \multicolumn{2}{c}{71.0} & \multicolumn{2}{c}{95.8} & \multicolumn{2}{c}{57.8} & \multicolumn{2}{c}{97.7} & \multicolumn{2}{c}{44.5} \\
    
    \multicolumn{4}{l}{CLIPort (single)}          & \multicolumn{2}{c}{98.0} & \multicolumn{2}{c}{\textbf{75.2}} & \multicolumn{2}{c}{\textbf{96.2}} & \multicolumn{2}{c}{71.9} & \multicolumn{2}{c}{\textbf{100}} & \multicolumn{2}{c}{25.0} \\

    \multicolumn{4}{l}{CLIPort (multi)}          & \multicolumn{2}{c}{89.0} & \multicolumn{2}{c}{62.8} & \multicolumn{2}{c}{84.4} & \multicolumn{2}{c}{70.3} & \multicolumn{2}{c}{\textbf{100}} & \multicolumn{2}{c}{\textbf{45.8}} \\
    
    \midrule
    
    \multicolumn{4}{l}{\agent}  & \multicolumn{2}{c}{\textbf{98.8}} & \multicolumn{2}{c}{\textbf{75.2}} & \multicolumn{2}{c}{76.6} & \multicolumn{2}{c}{\textbf{79.8}} & \multicolumn{2}{c}{86.2} & \multicolumn{2}{c}{35.2} \\
    \bottomrule
\end{tabular}
}
    \vfill
\centering
\small
\setlength\tabcolsep{2pt}
\captionof{table}{\label{tab:test_result_act_objnav}\textbf{Results on object navigation.} {$^\dagger$ indicates zero-shot evaluation.}}
\resizebox{\linewidth}{!}{
\begin{tabular}{lcccccc}
\toprule
 \multicolumn{2}{c}{\multirow{2}{*}{}} & \multicolumn{2}{c}{\small{MP3D-val}} & & \multicolumn{2}{c}{\small{HM3D-val}} \\ \cmidrule(lr){3-4}\cmidrule(lr){6-7}
\multicolumn{2}{l}{} &  \small{Success$(\uparrow)$} & \small{SPL$(\uparrow)$} & & \small{Success$(\uparrow)$} &  \small{SPL$(\uparrow)$}\\ \midrule
\multicolumn{2}{c}{Habitat-web (shortest)} & 4.4 & 2.2 & & - &  - \\
\multicolumn{2}{c}{Habitat-web (demo)} & \textbf{35.4} & 10.2 & & - & - \\
\multicolumn{2}{c}{ZSON} & 15.3$^\dagger$ & 4.8$^\dagger$ & & \textbf{25.5} & 12.6 \\
\midrule
\multicolumn{2}{c}{\agent} & 23.1 & \textbf{15.2} & & 23.1$^\dagger$ & \textbf{19.1}$^\dagger$ \\
\bottomrule

\end{tabular}
}  
\end{minipage}
\vskip -0.15in
\end{table*}

\section{Capabilities and Analyses}\label{sec:exp}
We demonstrate \agent's capabilities by a comprehensive evaluation on the full spectrum of embodied 3D tasks encompassing perceiving, grounding, reasoning, planning, and acting.
In \cref{sec:exp_3dvl}, we present quantitative comparisons between \agent and \sota models on various 3D VL tasks, underscoring \agent's proficiency in 3D VL understanding and reasoning. In \cref{sec:exp_dialog}, we highlight \agent's strength in scene-grounded dialogue and task planning. In \cref{sec:exp_eai}, we extend \agent to embodied acting tasks wherein \agent exhibits remarkable versatility. In \cref{sec:ablation}, we conduct ablative studies to reveal more insights into \agent, including data and model aspects. In \cref{sec:exp_scaling}, we probe the scaling effect and manifest the potential for further development.

\subsection{3D Vision-Language Understanding and Reasoning}\label{sec:exp_3dvl}

\paragraph{Overview.} Understanding and reasoning about object attributes, object relations, and other facets of 3D scenes from an agent's egocentric perspective is a fundamental capability of an embodied generalist agent in the 3D world. We investigate \textit{how well can \agent perform 3D VL understanding and embodied reasoning tasks, especially when being compared against task-specific models and existing generalist agents}. Specifically, we consider three renowned 3D tasks: 3D captioning on Scan2Cap~\citep{chen2021scan2cap}, 3D QA on ScanQA~\citep{azuma2022scanqa}, and 3D embodied reasoning on SQA3D~\citep{ma2023sqa3d}. Our evaluation metrics include conventional scores (\eg, CIDEr, BLEU, METEOR, ROUGE) and other metrics adapted for open-ended generation, \eg, sentence similarity \citep{reimers2019sentence} and refined exact-match accuracy (see details in \cref{sec:supp_eval_qa}).
Following 3D-VisTA~\citep{zhu20233d}, \textbf{we use object proposals from Mask3D~\citep{schult2022mask3d} instead of ground-truth object segments for evaluation.}

\paragraph{Baselines.} For quantitative comparisons, we include both task-specific approaches and generalist models: 1) \sota specialists in 3D dense captioning \citep{chen2021scan2cap,cai20223djcg,chen2023end}; 2) \sota specialists in 3D QA \citep{azuma2022scanqa,ma2023sqa3d}; 3) task-specific fine-tuned generalist models like 3D-VisTA \citep{zhu20233d} and 3D-LLM \citep{hong20233d}. To the best of our knowledge, \textit{\agent is the first model that, in stark contrast to prior models, can directly handle the aforementioned 3D VL tasks in a unified architecture without task-specific fine-tuning}. This lends greater credence to \agent's comparative superiority.

\paragraph{Results \& analysis.} As shown in \cref{tab:vl_results}, \agent surpasses both \sota single-task and task-specific fine-tuned models significantly on 3D dense captioning and 3D QA tasks. In contrast to the specialist models that utilize task-specific heads, our LLM-based approach not only affords the flexibility of generating open-ended responses but also exhibits excellent quantitative results. On the other hand, considering the complicated feature aggregation in 3D-LLM, we believe that object-centric 3D representation is a simple yet effective option to connect 3D scenes with LLM while harnessing the inherent knowledge of LLM.

\vspace{-0.3em}
\subsection{Scene-grounded Dialogue and Planning}\label{sec:exp_dialog}

\paragraph{Overview.} Upon the 3D VL understanding and reasoning, we anticipate \agent to support more sophisticated interaction with humans, \eg, responding to complex multi-round user instructions in the 3D world. To verify these capabilities, we conduct qualitative studies on 3D dialogue and planning tasks, with unseen scenarios from the held-out test sets of \agent-instruct. We defer the quantitative results of dialogue and planning to our ablation study in \cref{sec:ablation}. Quantitative comparison with other approaches is infeasible given the absence of comparable benchmarks.

\paragraph{Results \& analysis.} As shown in \cref{fig:qualitative}, \agent is capable of generating high-quality responses, which encompass two features: \textbf{1) Precisely grounded to the 3D scenes.} The task plan proposed by \agent involves concrete objects related to the 3D scene, as well as plausible actions regarding these objects. \textbf{2) Rich informative spatial relations.} The entities in \agent's responses often accompany detailed depictions. Such information helps identify specific objects in complex 3D scenes and affords considerable assistance to humans.

\begin{table*}
\begin{minipage}{0.41\linewidth}
    \centering
    \captionof{table}{Quantitative results of \agent trained with different data configurations. \textit{w/o Align}: without alignment stage. \textit{ScanNet}: tuned on ScanNet scenes only. \textit{w/o Act}: tuned without embodied acting tasks. We report the exact match metrics for QA tasks and sentence similarity for others. \underline{Underlined figures} indicate zero-shot results on novel scenes (3RScan).}
    \setlength\tabcolsep{3pt}
    \resizebox{\linewidth}{!}{
    \begin{tabular}{lcccccc}
        \toprule
     & \multicolumn{3}{c}{ScanNet} & \multicolumn{3}{c}{3RScan} \\
     \cmidrule(lr){2-4} \cmidrule(lr){5-7}
         & Scan2Cap & ScanQA & SQA3D & 3RQA & 3RDialog & 3RPlan \\
        \midrule
        \rowcolor{color1} \textit{w/o Align} & 62.8 & 22.7 \textcolor{gray}{(45.0)} & \textbf{50.9 \textcolor{gray}{(53.2)}} & 49.7 \textcolor{gray}{(53.7)} & 73.0 & 80.3 \\
        \rowcolor{color2} \textit{ScanNet} & 64.0 & 24.4 \textbf{\textcolor{gray}{(49.2)}} & 46.8 \textcolor{gray}{(49.5)} & \underline{35.8 \textcolor{gray}{(50.0)}} & \underline{25.5} & \underline{23.4} \\
        \textit{w/o Act} & \textbf{65.4} & 24.3 \textcolor{gray}{(48.5)} & 50.0 \textcolor{gray}{(52.5)} & \textbf{51.9 \textcolor{gray}{(57.4)}} & \textbf{73.3} & \textbf{81.1} \\
        \rowcolor{color4} \textit{VLA} & 65.3 & \textbf{25.0} \textcolor{gray}{(48.9)} & 46.2 \textcolor{gray}{(48.3)} & 51.3 \textcolor{gray}{(55.8)} & 72.3 & 77.2 \\
        \bottomrule
    \end{tabular}
    }
    \label{tab:data_ablation}
\end{minipage}
\hfill
\begin{minipage}{0.31\linewidth}
\centering
\small
\setlength\tabcolsep{3pt}
\captionof{table}{TrueSkill scores with human preference. \textit{Dialg}: dialogue and planning data.}\label{tab:response_trueskill}
    \resizebox{\linewidth}{!}{
\begin{tabular}{lccc}
\toprule
 & Answerable & Unanswerable & NLP \\
 \midrule
 \textit{w/o Dialg} & 24.4$\pm$1.3 & 23.1$\pm$1.4  & 23.4$\pm$1.4 \\
 \textit{w/ Dialg} & \textbf{25.6}$\pm$\textbf{1.3}  & \textbf{26.8}$\pm$\textbf{1.4} & \textbf{26.6}$\pm$\textbf{1.4} \\
 \bottomrule 
\end{tabular}
}
\vfill
\vspace{0.3em}
\centering
\small
\captionof{table}{Answer accuracy (EM) on object-existence questions. \textit{Aug}: augmented data.}\label{tab:data_balance}
\resizebox{\linewidth}{!}{
\begin{tabular}{lcccccccc}
    \toprule
     & \multicolumn{3}{c}{3RScan} & \multicolumn{3}{c}{ScanNet (0-shot)} \\
     \cmidrule(lr){2-4} \cmidrule(lr){5-7}
     & Yes & No & Overall & Yes & No & Overall \\
     \midrule
    \textit{w/o Aug} & \textbf{1.00} & 0.01 & 0.34 & \textbf{0.98} & 0.16 & 0.43 \\
    \textit{w/ Aug} & 0.72 & \textbf{0.91} & \textbf{0.85} & 0.88 & \textbf{0.81} & \textbf{0.83} \\
    \bottomrule
\end{tabular}
}
\end{minipage}
\hfill
\begin{minipage}{0.263\linewidth}
    \centering
    \captionof{figure}{\label{fig:scaling_law}\agent-instruct test loss with the growth of data and model scale, manifesting the scaling law.}
    \vspace{0.2em}
    \includegraphics[width=\linewidth]{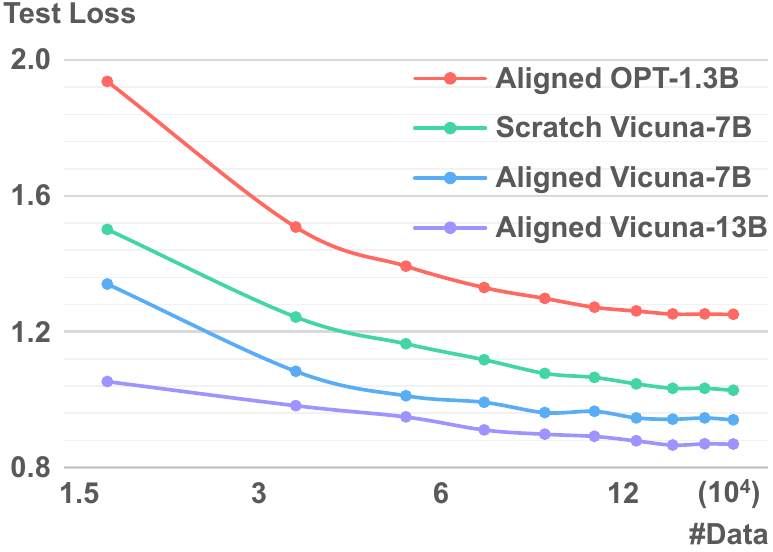}
\end{minipage}
\end{table*}

\subsection{Embodied Action in 3D World}\label{sec:exp_eai}

\paragraph{Overview.} To probe \agent's capacity of bridging vision-language-acting in the 3D world, we select two canonical embodied AI tasks: embodied navigation (\texttt{ObjNav}) on AI Habitat~\citep{ramrakhya2022habitat} and robotic manipulation on CLIPort~\citep{cliport}.
Specifically, for CLIPort robotic manipulation, we evaluate \agent on the three tasks listed in~\cref{tab:test_result_act_cliport} including their unseen counterparts, and report the success scores. For \texttt{ObjNav}, 
we evaluate \agent on the original MP3D \texttt{ObjNav} validation split. Additionally, we test generalization to the validation split of the newly introduced HM3D \texttt{ObjNav} task~\citep{ramakrishnan2021habitat}. We report the success rate and SPL metrics following \citet{ramrakhya2022habitat}. We consider both Habitat-web \citep{ramrakhya2022habitat} (fully supervised) and ZSON \citep{majumdar2022zson} (zero-shot) as baselines.

\paragraph{Results \& analysis.} We present the results of CLIPort manipulation and object navigation in~\cref{tab:test_result_act_cliport,tab:test_result_act_objnav}. Our findings are as follows: 1) In robotic manipulation, \agent is comparable to \sota performances and even better on some challenging {\color{myred}{unseen}} tasks. In particular, \agent directly produces motor commands without inductive bias (\eg, heatmap) that benefit previous methods, showcasing \agent's considerable capacity for learning embodied actions. 2) In \texttt{ObjNav}, \agent achieves a success rate that is comparable to the baselines and has a better SPL on MP3D-val, suggesting that \agent can leverage the object-centric 3D scene input (potentially offering a coarse global map) and take a shorter path to the target. Furthermore, results on HM3D-val confirm \agent's zero-shot generalization to novel scenes. Notably, all baselines are equipped with recurrent modules while \agent only incorporates truncated past actions, which could account for a lower success rate (see discussion in \cref{sec:supp_eai_split}). 3) Overall, the two-stage learning scheme endows \agent with semantic-level generalization (novel objects, \etc) in both manipulation and navigation tasks. We demonstrate the efficacy of tackling embodied acting tasks with a general framework from 3D VL.

\paragraph{Additional results.} We further investigate the perception modules, data regime, and generalization to unseen objects in \texttt{ObjNav} task. See the results in \cref{sec:result_objnav_additional}.

\vskip -0.15in
\subsection{More Insights into \agent}\label{sec:ablation}

\paragraph{Overview.} In this section, we aim to offer deeper insights into \agent's characteristics, mainly from the data perspective (model perspective is deferred to \cref{sec:model_ablation}). Specifically, we evaluate \agent when trained with different data configurations, including exact match, sentence similarity, and human rating. We regard \agent instruction-tuned without embodied acting tasks (\textit{w/o Act}) as the default setting. Following \citet{achlioptas2020referit3d}, we use ground-truth object segments in these analyses. We present additional analyses on data in \cref{sec:data_comparison} and model in \cref{sec:model_comparison}.

\paragraph{Alignment stage.} In contrast to complete two-stage training (\textit{w/o Act}), we direct instruction-tune a model without alignment stage (\textit{w/o Align}). The results in \cref{tab:data_ablation} show the consistent impact of alignment. In particular, the benefit of alignment is significant on Scan2Cap since it concerns detailed scene understanding and captioning, which is a primary focus of alignment training. %

\paragraph{Specialist \vs generalist.} We train a specialist on ScanNet scenes (\textit{ScanNet}). As shown in \cref{tab:data_ablation}, \textit{ScanNet} performs slightly worse than \textit{w/o Act} even on ScanNet tasks, and particularly struggles at generalization across scenes (3RQA) and tasks (3RDialog and 3RPlan). This demonstrates the advantage of generalist-style instruction tuning with broad coverage of scenes and tasks.

\paragraph{VL \vs VLA.} We compare \textit{w/o Act} and \textit{VLA}, which differ in whether embodied acting tasks are included for training. The results in \cref{tab:data_ablation} show that incorporating embodied acting tasks could lead to performance drops on 3D VL tasks. This may stem from 1) the gap between language generation and embodied action prediction, and 2) the imbalanced data scale of embodied acting tasks. In contrast to the finding that VL data benefits embodied acting tasks in VLA co-training \citep{brohan2023rt}, our observation implies that embodied acting tasks may harm VL capabilities in turn. How to continually bridge the gap between VL and embodied acting tasks is an important direction for further exploration.

\paragraph{Dialogue and planning data.} In contrast to the default model (\textit{w/ Dialg} in \cref{tab:response_trueskill}), we train \agent without dialogue and planning data (\textit{w/o Dialg}). We design an evaluation set with three types of questions (Answerable, Unanswerable, and NLP) and evaluate with TrueSkill~\citep{graepel2007bayesian} according to human preference (see details in \cref{sec:dialog_planning}). The results in \cref{tab:response_trueskill} confirm more hallucinations (less preferred by users on ``Unanswerable'') and worse NLP skills for \textit{w/o Dialg}. This is probably because 1) the diverse conversations in our dialogue data can help cultivate flexible responses to complex instructions, and 2) our planning data can offer scene-grounded commonsense knowledge and also encourage detailed coherent text.

\paragraph{Data balancing.} We find imbalanced data could induce hallucination in \agent, \eg, it tends to respond with ``Yes'' when asked ``Is there something in this room?''. To address this, we augment the 3RScanQA data with more negative samples where non-existent objects are queried. We also design an evaluation set with different types (Yes and No) of object-existence questions (see details in \cref{sec:data_balancing}). Results in \cref{tab:data_balance} demonstrate that we can effectively mitigate the hallucination problem by balancing the tuning data. Moreover, the benefit of augmenting 3RScan data can transfer to ScanNet scenes in a zero-shot manner.

\subsection{Scaling Law Analysis}\label{sec:exp_scaling}

\paragraph{Settings.} 
We study the scaling effect~\citep{kaplan2020scaling,reed2022generalist} of data and model in \agent by tracking the instruction-tuning loss on the test set with the growth of data scale. In addition to the default Vicuna-7B, we incorporate two LLMs at different scales: OPT-1.3B \citep{zhang2022opt} and Vicuna-13B \citep{vicuna2023}. For Vicuna-7B, we also probe the influence of alignment (Scratch \vs Aligned).

\paragraph{Results \& analysis.} From the test loss curves in \cref{fig:scaling_law}, we have the following findings: \textbf{1) The instruction tuning of \agent conforms to the scaling law}~\citep{kaplan2020scaling,reed2022generalist}. We observe that all curves decrease log-linearly with the data scale. \textbf{2) Scaling up LLM leads to consistent improvements.} Aligned Vicuna-7B shows significantly lower losses than Aligned OPT-1.3B. In contrast, despite the consistent improvements, the gap between Aligned Vicuna-7B and Vicuna-13B appears less significant, suggesting potential saturation if we continue to scale up the LLM. This indicates the scalability of \agent and the necessity of scaling up data to match the model capacity. \textbf{3) Alignment leads to consistent improvements.} Aligned Vicuna-7B shows consistently lower losses than Scratch Vicuna-7B, which corresponds to the inferior performances of \textit{w/o Align} in \cref{tab:data_ablation}.

\section{Related Work}\label{app:sec:related}

\paragraph{Generalist agents.} The AI community has witnessed the rising generalist models in both vision~\citep{lu2023unified,wang2023images,kirillov2023segment} and language~\citep{openai2022chatgpt,openai2023gpt4} domains. A generalist agent requires additional embodiment knowledge to interact with the environment and complete embodied acting tasks. Existing efforts towards generalist agents include: grounded reasoning and task planning in the real world~\citep{ahn2022can,huang2022inner}, skill generalization in open-world environment~\citep{fan2022minedojo,cai2023open,wang2023describe,wang2023voyager,cai2023groot,jxma_llm_vla_vlm_mas_multiagent_2023}, general robotic manipulation~\citep{brohan2022rt,jiang2023vima,gong2023arnold}, and unified vision-language-action (VLA) models such as Gato~\citep{reed2022generalist}, PaLM-E~\citep{driess2023palm}, EmbodiedGPT~\citep{mu2023embodiedgpt}, and RT-2~\citep{brohan2023rt}. \agent belongs to the \ac{vla} model, however, its goal is to build a generalist agent that can understand the real 3D world beyond 2D images, which is absent in existing works.

\paragraph{Multi-modal instruction tuning.} Pre-trained LLMs demonstrated practical for solving vision-language tasks~\citep{tsimpoukelli2021multimodal,alayrac2022flamingo,guo2023images,li2023blip,jxma_vlm_multimodal_2023}. Meanwhile, the instruction-tuning paradigm exhibited strong zero-shot generalization in NLP tasks~\citep{wei2022finetuned,sanh2022multitask,ouyang2022training,chung2022scaling}. The two streams merged into instruction-tuned LVLMs~\citep{liu2023visual,zhu2023minigpt,ye2023mplug,gao2023llama,li2023otter,gong2023multimodal,dai2023instructblip}. Despite the burst, these models are confined to 2D visual modalities, \eg, image or video. Concurrent works~\citep{yin2023lamm,hong20233d,wang2023chat,xu2023pointllm} extend to 3D vision tasks, but these models either lack the acting capability or unified efficient architecture.

\paragraph{Grounded 3D scene understanding.}
One key obstacle to building \agent is grounding the 3D world with natural languages. There exist diverse methods of grounded scene understanding, \eg, spatial relation modeling \citep{zhao20213dvg,chen2022language,zhu20233d} and fine-grained open-scene understanding \citep{peng2023openscene,kerr2023lerf}. However, due to data scarcity, how to utilize \ac{llm} to ground the 3D scene is rarely explored. Recently, 3D-LLM \citep{hong20233d} leverages multi-view images and Chat-3D~\citep{wang2023chat} uses object-centric point clouds to enable the \ac{llm} with 3D grounding. In this work, we devise both 2D and 3D encoders for grounding various visual representations and employ LoRA~\citep{hu2022lora} to efficiently fine-tune the \ac{llm}.

\paragraph{3D data prompting from LLMs.} LLMs exhibit extraordinary capabilities of text generation and serve as a source for collecting diverse instruction-following data~\citep{wang2023self,alpaca,peng2023instruction}. However, the lack of access to visual modalities makes it troublesome to collect visual instruction-tuning data. To address this issue, existing methods provide bounding boxes~\citep{liu2023visual} and add dense captions~\citep{li2023mimic,liu2023aligning} as image descriptions or directly use off-the-shelf \ac{lvlm}~\citep{zhu2023chatgpt,luo2023scalable} to help collect such data. Unlike concurrent attempts~\citep{yin2023lamm,hong20233d,wang2023chat} in collecting 3D instruction-tuning data, our approach features a scene-graph-based prompting and refinement method to prompt and correct the data.

\section{Conclusions}\label{sec:conclusion}
The proposed agent \agent extends the current generalist ability of \ac{llm} from text towards the 3D world and embodied tasks. It is a crucial initial step towards building embodied generalist agents. Nonetheless, there are also limitations, \eg, generalization to novel scenes, and a notable gap between VL learning and embodied action control. In light of this work, we identify several promising directions that hold the potential for substantial advancement: (1) enhancing the 3D \ac{vl} understanding capability by leveraging larger-scale VL data from richer 3D domains; (2) continually bridging the gap between 3D \ac{vl} and embodied action, as our experiments reveal the efficacy of their joint learning; (3) investigating the issues of safety and alignment in the context of embodied generalist agents, particularly given that our scaling law analysis suggests significant enhancements through scaling on data and model.

\section*{Impact Statement}
This work introduces LEO, an embodied multi-modal generalist agent designed to extend machine learning capabilities into the 3D realm, marking a significant advance in the field. The potential societal implications of LEO are manifold, touching on robotics, AR/VR, assistive technologies, and environmental planning. Ethically, it underscores the importance of responsible AI development, emphasizing safety, privacy, and fairness in automated decision-making. As LEO ventures into new territories of human-machine interaction, it prompts a re-evaluation of ethical frameworks to ensure that advancement contributes positively to society. While the immediate societal consequences of our work align with the goals of advancing machine learning, we acknowledge the necessity of ongoing ethical consideration as applications of LEO evolve.

\section*{Acknowledgements}
This work is supported in part by the National Science and Technology Major Project (2022ZD0114900).

\bibliography{ref}
\bibliographystyle{icml2024}

\clearpage
\onecolumn
\appendix
\clearpage

\renewcommand{\thefigure}{A.\arabic{figure}}
\renewcommand{\thetable}{A.\arabic{table}}
\renewcommand{\theequation}{A.\arabic{equation}}
\setcounter{figure}{0}
\setcounter{table}{0}
\setcounter{equation}{0}

\section{Qualitative Results}
\begin{figure*}[h!]
\centering
\includegraphics[width=\textwidth, keepaspectratio]{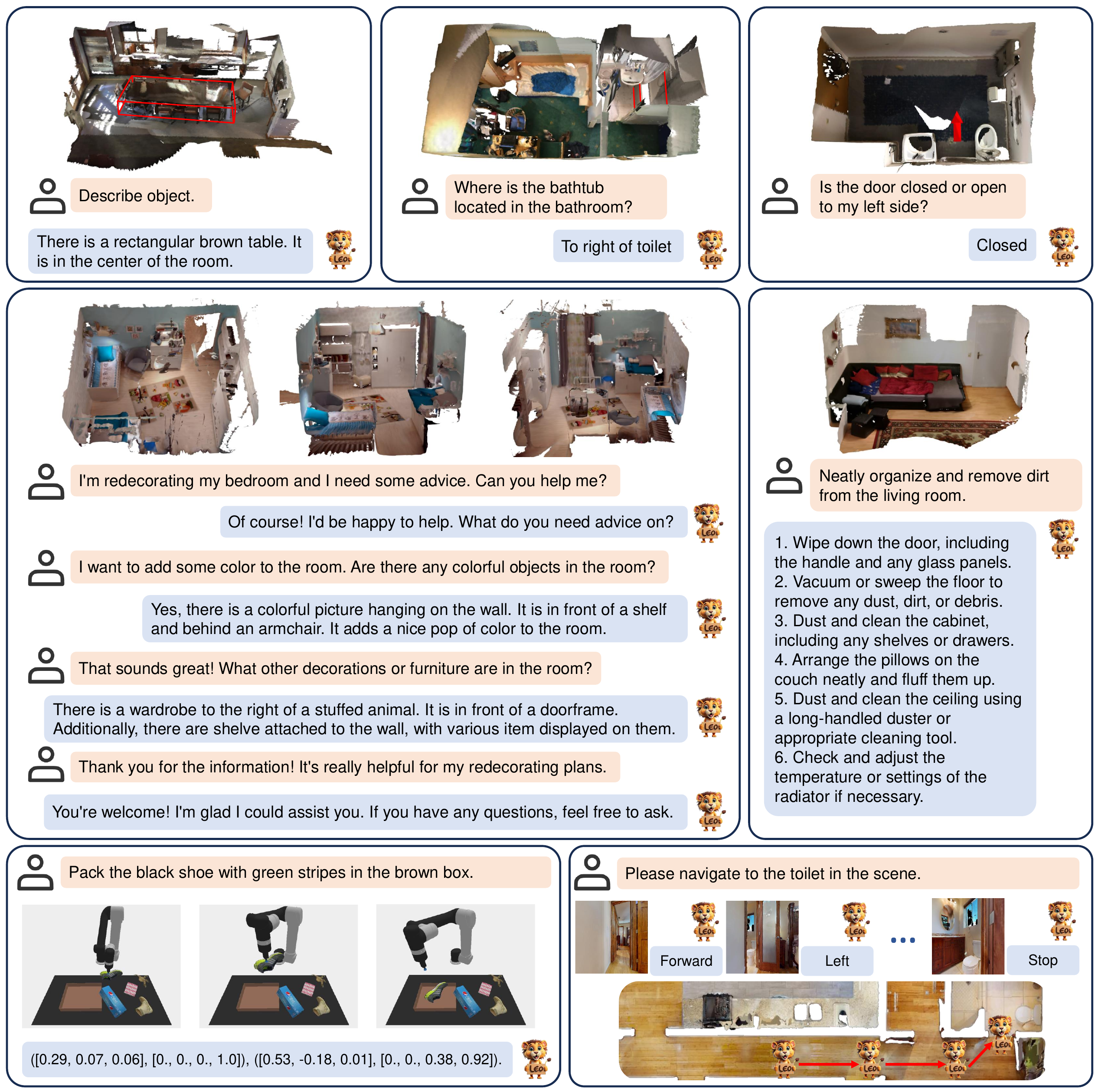}%
\caption{\label{fig:qualitative} \textbf{Qualitative results of interacting with \agent} on unseen scenarios from a held-out test set of \agent-instruct. \agent's responses and actions can be grounded in novel scenes.}
  \vspace{-16pt}
\end{figure*}

\section{Data}\label{app:dataset}

\subsection{More Details on LEO-align}\label{app:dataset:leo_align}
\paragraph{Object-level caption.} To facilitate object-level grounding of detailed object attributes, we leverage Cap3D~\citep{luo2023scalable}, which contains language descriptions for objects in Objaverse~\citep{deitke2023objaverse}. Given a single 3D object as input, \agent will be asked to predict its caption.

\paragraph{Object-in-the-scene caption.} For a better understanding of how an object can be related to others (spatial relations, \etc) when situated in a 3D scene, we collect referring expressions of objects in scenes from existing datasets, including ScanScribe~\citep{zhu20233d} and ReferIt3D~\citep{achlioptas2020referit3d}. Further, we generate additional object-referring expressions on 3RScan~\citep{wald2019rio} scenes by prompting \ac{llm} (details in \cref{app:dataset:prompt}).
During alignment, \agent needs to predict these referring expressions given the object-centric 3D input of the scene and the referred object.

\paragraph{Scene-level caption.} Finally, we encourage \agent to capture scene-level descriptions of a 3D scene. These scene-level captions focus on global information depicting key objects in the scene as well as their attributes and functionalities, relations among multiple objects, and room types and styles. We leverage scene graph annotations~\citep{wald2019rio} and prompt \ac{llm} to produce a total of \textasciitilde{}20K captions. 
To further increase caption diversity, we propose a subgraph sampling strategy to prevent LLMs from always attending to certain notable facets of the scene (details in \cref{app:subgraph_sampling}). Similar to previous settings, \agent needs to predict these captions given the corresponding 3D input.

\subsection{More Details on LEO-instruct}\label{app:dataset:leo_instruct}

Below, we provide a comprehensive illustration of the data preparation process for these tasks and an overview of generated data in~\cref{fig:data_framework}. We list the corresponding instructions in \cref{sec:supp_leo_ds_examples}.

\paragraph{3D captioning.} The task is to produce a generic caption given 3D input. We adopt the Scan2Cap dataset~\citep{chen2021scan2cap}, which is based on the ScanNet~\citep{dai2017scannet} 3D scenes and covers various levels (object-level and scene-level) and aspects (attributes, relations, \etc) of scene details.

\paragraph{3D question answering.} The 3D-QA task is an extension of VQA~\citep{antol2015vqa} to 3D scenes with a focus on 3D knowledge, ranging from spatial relations to functionalities of objects. For this task, we first aggregate two existing 3D-QA datasets: ScanQA~\citep{azuma2022scanqa} and SQA3D~\citep{ma2023sqa3d}. To further generate questions concerning rich 3D knowledge, we prompt LLMs to generate \textasciitilde{}35K QA pairs on 3RScanQA with our quality refinement techniques discussed in~\cref{sec:data:generation}.

\paragraph{3D dialogue.} The goal of this task is to support natural conversations between \agent and users about a given 3D scene. 
This task necessitates coherence and continuity across multiple rounds of conversational interactions.
We build such dialogues on 3RScan scenes by prompting LLMs with a variant of the Chain-of-Thought prompting method discussed in~\cref{sec:data:generation} to facilitate diverse dialogues about relevant and accurate details about the 3D scene. 
In total, \textasciitilde{}11K dialogues are collected.

\paragraph{Scene-aware task planning.} In this task, \agent is required to decompose high-level tasks into step-by-step low-level plans given 3D scenes. 
We expect \agent to generate feasible plans based on the current 3D scene and ground its inherent common sense knowledge about procedures to the scene configurations, including, objects, their attributes, relations, and functional characteristics, \etc. By prompting LLMs, we end up collecting \textasciitilde{}14K task-plan pairs on 3RScan scenes.

\paragraph{Embodied navigation.} We follow imitation learning setting in Habitat-web~\citep{ramrakhya2022habitat} for the embodied navigation task. We choose \texttt{ObjNav}, where \agent needs to map navigation instructions (\eg ``find bed''), object-centric 3D input, and an egocentric 2D input into discrete habitat motor commands. For simplicity, we use shortest path navigation trials rather than human demonstrations for learning as they are less noisy and therefore easier to learn when provided with the 3D scene. In total, we generate \textasciitilde{}60K navigation episodes out of the MP3D \texttt{ObjNav} training scenes~\citep{savva2019habitat} for this task.

\paragraph{Robotic manipulation.} We employ a subset of the manipulation tasks introduced in CLIPort~\citep{cliport}. The input of this task includes instructions, egocentric 2D observations, and object-centric 3D information. As discussed in~\cref{sec:model_tokenization}, we discretize the continuous action space of CLIPort into bins to unify the action decoding of navigation and manipulation (more details in~\cref{sec:action_tokenization}). We generate 100K demonstrations for each selected manipulation task. 

\subsection{Design of Seed Tasks for LLM-assisted 3D Data Generation}\label{app:dataset:seed_task}

\paragraph{Object Scene Caption \& Scene Caption.} To align the 3D scene/object with language, we prompt ChatGPT to curate these two types of caption data. Object Scene Caption includes the spatial relationships of the object with some adjacent objects in the scene. Scene Caption is the comprehensive description for the whole 3D scene, including some key objects and their spatial relationships.

\paragraph{QA \& Dialogue.} For QA, we design several question-answer pairs given a scene graph. A diverse set of questions are asked about the 3D scene, including the object attributes, object counting, object existence, spatial relationships between the objects, object types, object affordance, room type and so on. For dialogue, we design a conversation between the assistant and a person asking questions about this scene. The answers are in a tone as if the assistant is understanding the scene and helping the person. Different from single-round QA, dialogue has some high-level tasks such as 'searching for specific objects' that require multi-round conversations.

\paragraph{Planning.} To include a deeper understanding of the global 3D scene information, we prompt ChatGPT to generate a high-level task and 5-10 action steps(interaction between the assistant and the objects in the scene) to finish the task.

\subsection{Prompts for LLM-assisted 3D Data Generation}\label{app:dataset:prompt}
\begin{figure}[t!]
\centering
\includegraphics[width=\textwidth, keepaspectratio]{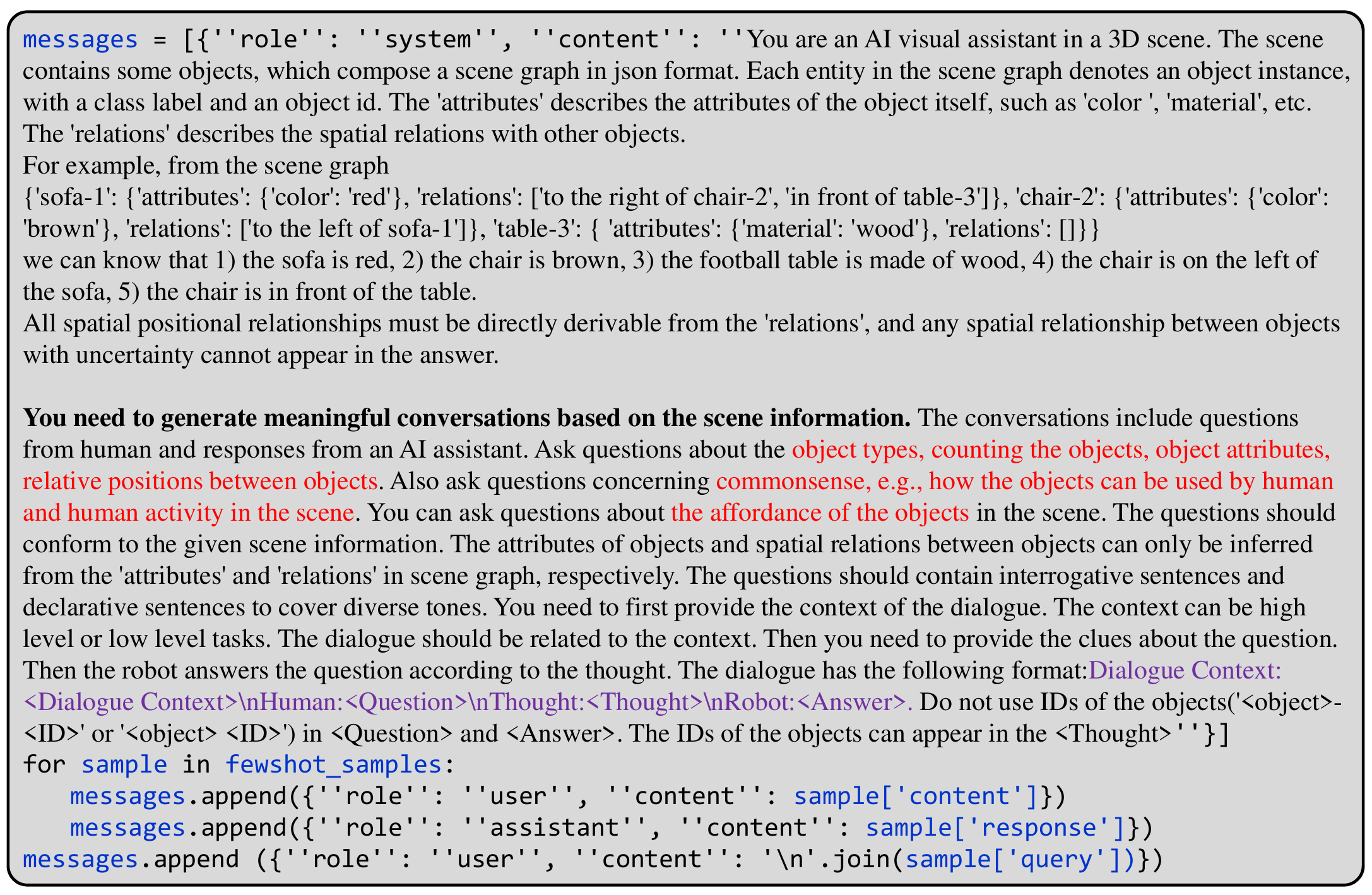}%
  \caption{The prompt for generating 3D Dialogue.}
  \label{fig:prompt:dialogue}
\end{figure}

\begin{figure}[t!]
\centering
\includegraphics[width=\textwidth, keepaspectratio]{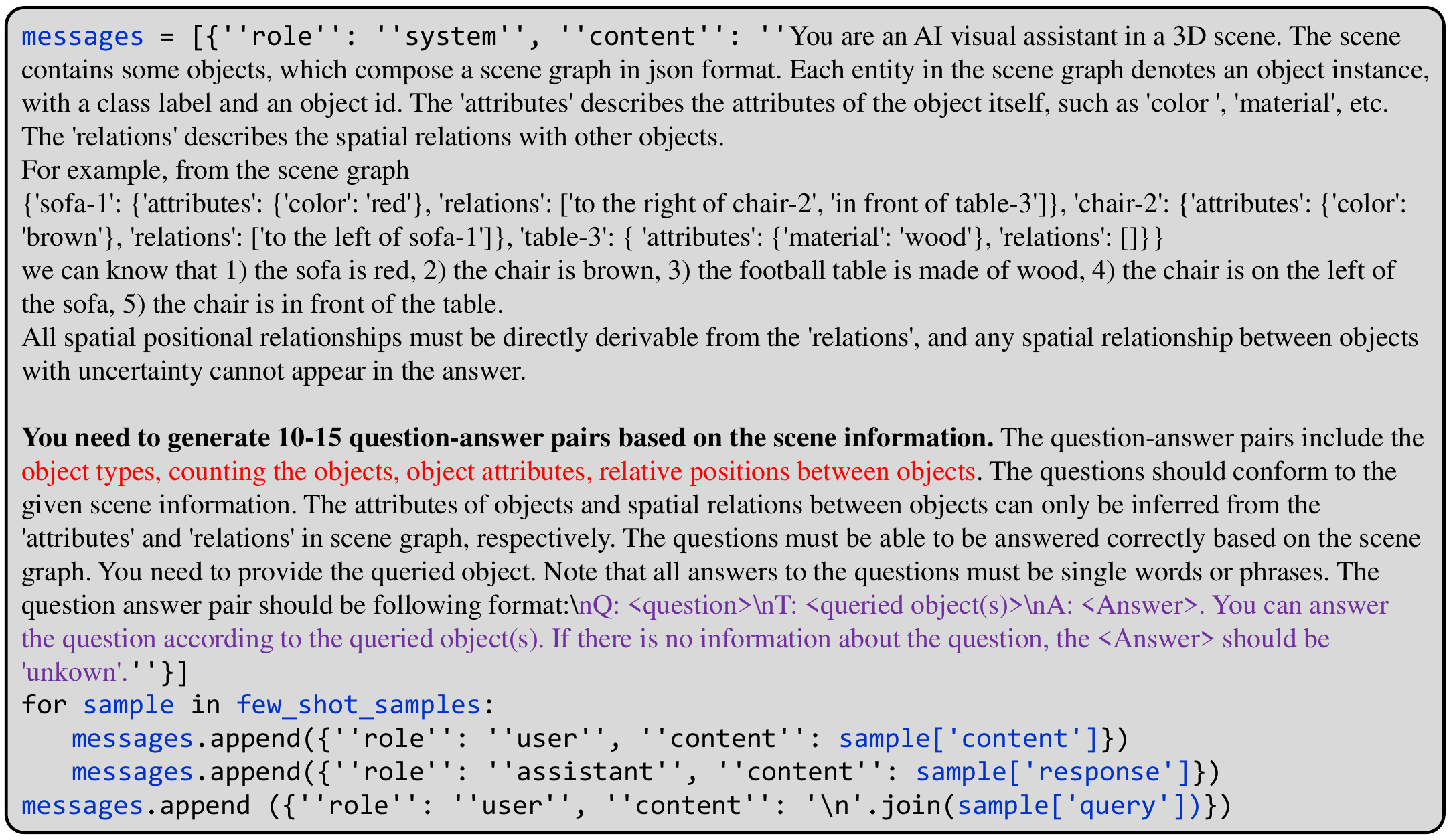}%
  \caption{The prompt for generating 3D QA.}
  \label{fig:prompt:QA}
\end{figure}

\begin{figure}[t!]
\centering
\includegraphics[width=\textwidth, keepaspectratio]{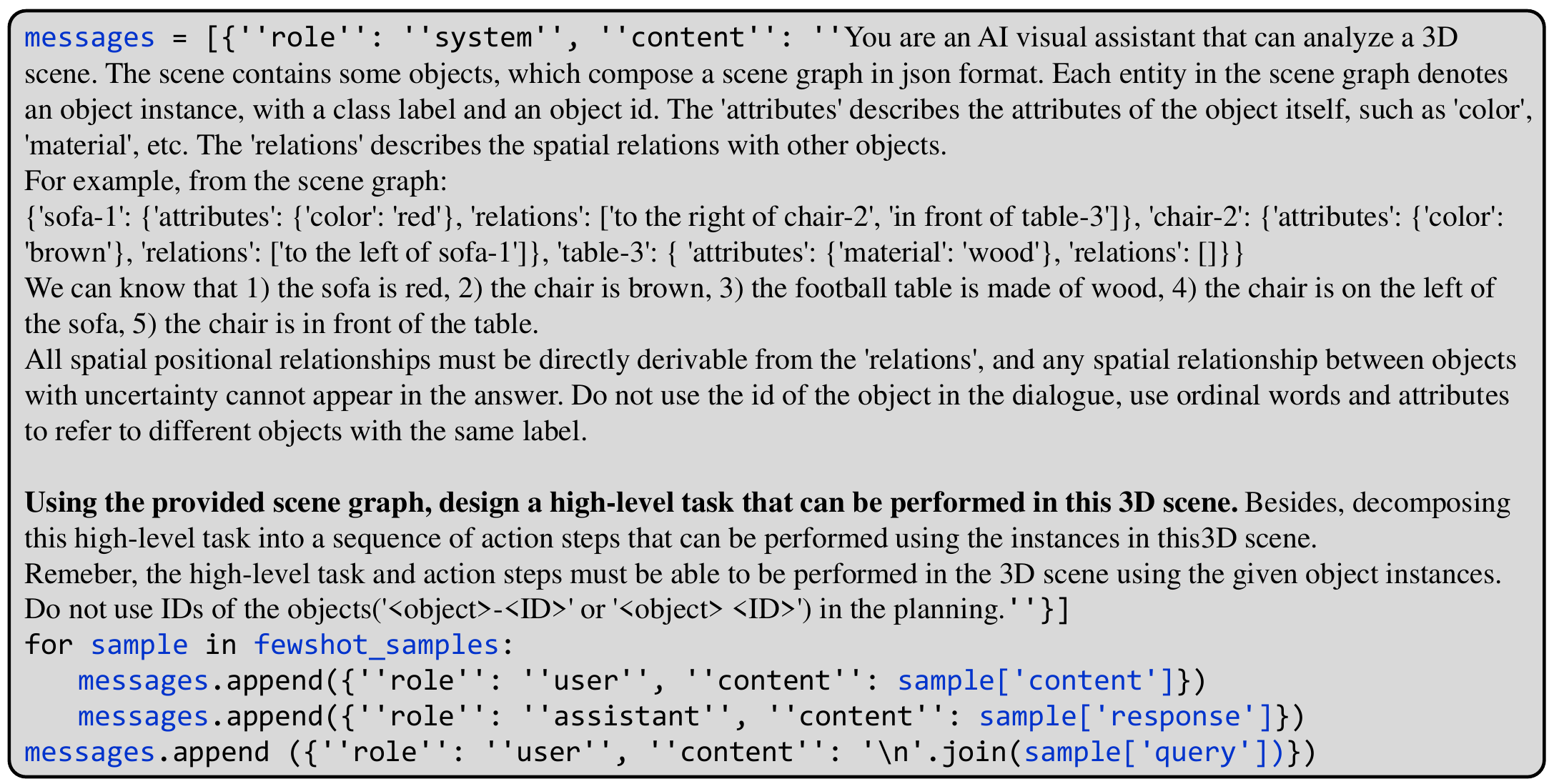}%
  \caption{The prompt for generating 3D planning.}
  \label{fig:prompt:planning}
\end{figure}

\begin{figure}[t!]
\centering
\includegraphics[width=\textwidth, keepaspectratio]{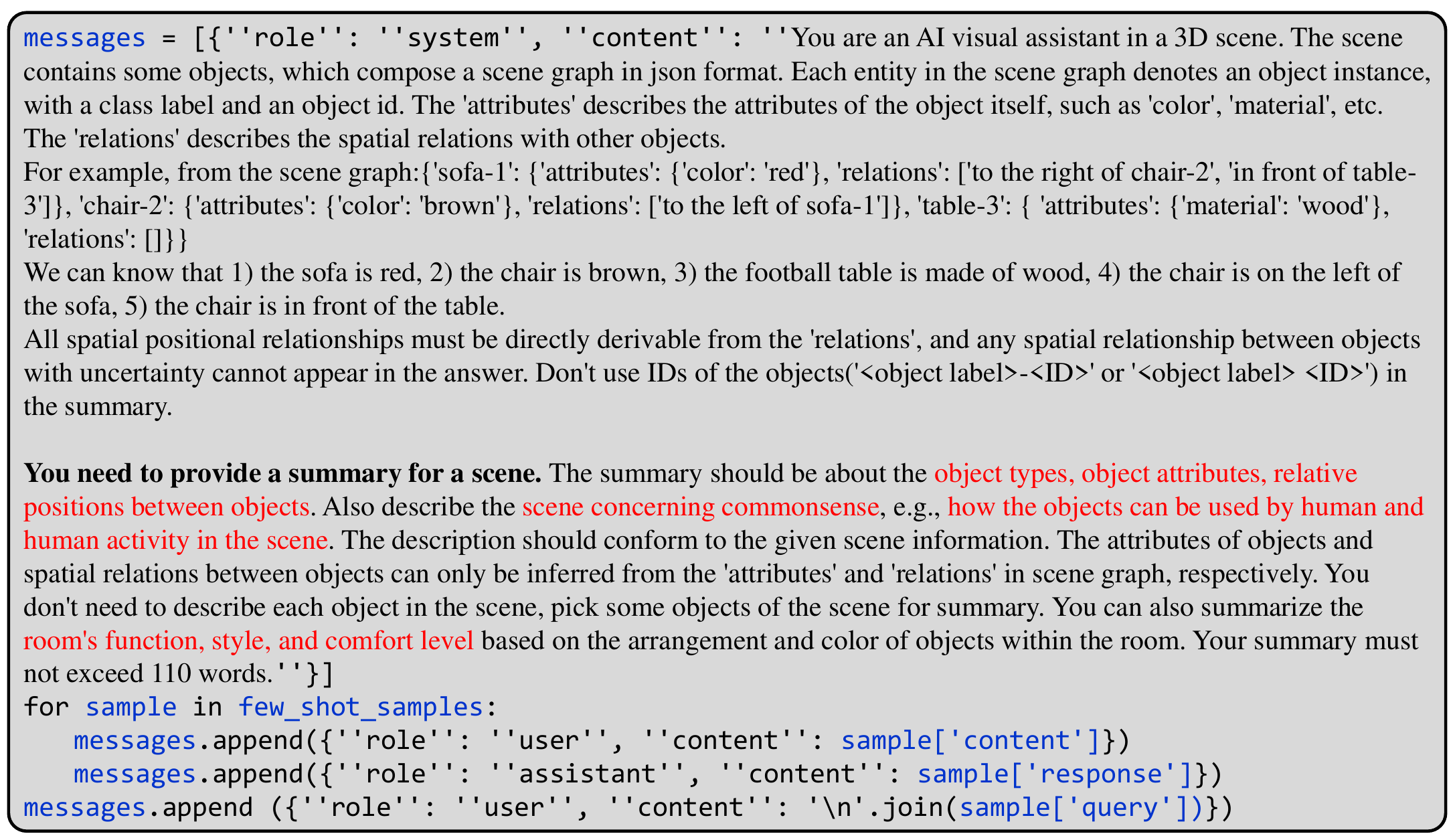}%
  \caption{The prompt for generating 3D scene caption.}
  \label{fig:prompt:scene_caption}
\end{figure}

\begin{figure}[t!]
\centering
\includegraphics[width=\textwidth, keepaspectratio]{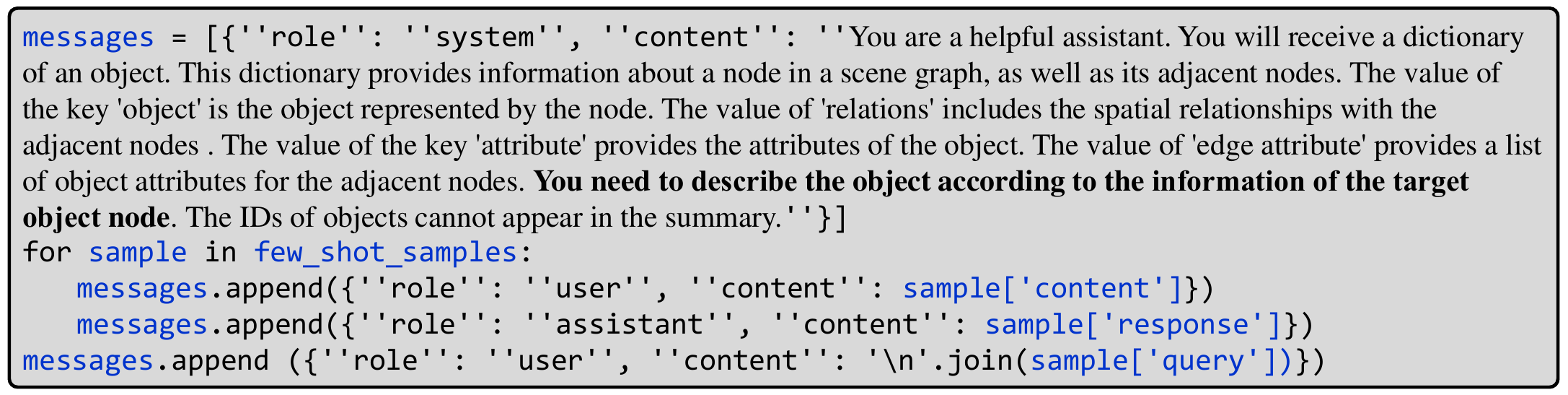}%
  \caption{The prompt for generating 3D object-in-the-scene caption. }
  \label{fig:prompt:object_caption}
\end{figure}

In \crefrange{fig:prompt:dialogue}{fig:prompt:object_caption}, we show the prompts for five types of LLM-assisted 3D-language data generation. We provide few-shot examples as the context. In each example, the ``content'' contains a scene graph, and the ``response'' refers to a human-labeled response. The query is a new scene graph, based on which ChatGPT \citep{openai2022chatgpt} generates responses.

\cref{fig:prompt:dialogue} shows the prompt for generating 3D dialogue data. {\color{red}{Red fonts}} outline our requirements of the dialogue content, including object attributes, spatial relations, and commonsense topics. \textcolor{purple}{Purple fonts} formulate the template of the response. We require the response generated by the ChatGPT should include the dialogue context as well; the ``thought'' contains the involved objects in the question, which is used to enhance the reliability of the answer. These two components will be removed after the refinement procedures.

\subsection{Analysis of the Object-Centric Chain-of-Thought}\label{app:dataset:ocot}
\begin{table}[t!]
    \centering
    \small
    \caption{The effect of \ac{ocot} on the answer accuracy for Object Counting questions.}
    \begin{tabular}{l|c c c c c c }
        \toprule
        \textbf{Settings} & \textbf{Seed 1} & \textbf{Seed 2} & \textbf{Seed 3} & \textbf{Seed 4} & \textbf{Average} & \textbf{Avg. Gain} \\
        \midrule
        w/o \ac{ocot} & 0.5838 & 0.5349 & 0.5962 & 0.5816 & 0.5741 & \multirow{2}[2]{*}{0.2061}\\
        \cmidrule{1-6}
        O-CoT & 0.7647 & 0.8117 & 0.7778 & 0.7667 & 0.7802 \\
        \bottomrule
    \end{tabular}
    \label{tab:ocot_ablation}
\end{table}
To further investigate the impact of Object-centric Chain-of-Thought (\ac{ocot}) on data quality, we analyze the answer accuracy for Object Counting questions. Specifically, we collect several demonstrations, and for each run, we select two of them as the prompt seed. With these seeds, we generate dialogues across all scenes in 3DSSG \citep{wu2021scenegraphfusion} and then assess the answer accuracy for Object Counting questions. The results are presented in~\cref{tab:ocot_ablation}.

The results in \cref{tab:ocot_ablation} indicate that \ac{ocot} consistently improves the answer accuracy for Object Counting questions. Though there remain errors after applying \ac{ocot}, we will conduct refinement to fix them. Examples of Object Counting questions are provided in \cref{app:dataset:refine:examples}.

\subsection{Refinement Details}\label{app:dataset:refine:examples}
We conduct refinement by passing raw LLM-generated responses into several human-defined filtering procedures based on the 3D scene graph. The refinement considers five raw response categories:
\begin{itemize}[leftmargin=*,nolistsep]
    \item Object Counting. The question concerns counting the target object.
    \item Object Existence. The response claims the existence of objects, which can be actually either existent or non-existent.
    \item Object Non-existence. The response claims the non-existence of objects, which can be actually either existent or non-existent.
    \item Negative Response. The scene graph cannot provide a solid response to the question, which means the question cannot be answered and will be discarded.
    \item Response with ID. The response contains unexpected object IDs.
\end{itemize} 

Specifically, we employ regular expression matching to detect errors in these five categories. We also employ this method to correct the responses except for Response with ID, which will be rewritten by ChatGPT instead. The QA pair will be eliminated if multiple rounds of rewriting fail to remove the IDs. \cref{tab:dialogue_refinement} and \cref{tab:qa_refinement} show some examples of the responses subject to the above five categories as well as the effect of our refinement. 
\begin{table}[t!]
    \centering
    \small
    \caption{\textbf{Examples of dialogue refinement}.}
    \begin{tabular}{l|p{4.5cm}|p{4.5cm}}
        \toprule
        \textbf{Types} & \textbf{Raw Responses} & \textbf{Refined Responses}\\
        \midrule
        \textbf{Object Counting} & There are 3 chairs in the room.
        
        I see there are two washing machines in the bathroom.
        & There are 4 chairs in the room.

        I see there are 4 washing machines in the bathroom. \\
        \midrule
        \textbf{Object Existence} & Yes, there is a cutting board in the kitchen.
        
        Yes, there is a computer and a monitor on the desk. However, the monitor is currently off.
        &  No, there is no cutting board in the room.
        
        No, there is no computer in the room.
        \\
        \midrule
        \textbf{Object Non-existence} & No, there is no stereo equipment in the room.
        
        I'm sorry, but I couldn't find a hair dryer in the bathroom.
        &Yes, there is a stereo equipment in the room.
        
        Yes, I found a hair dryer in the room.
        \\
        \midrule
        \textbf{Negative Response} &  
        No, there is nothing else mentioned in the scene graph. 
        
        I'm sorry, but there is no mention of a mirror in the scene graph for the bathroom.
        &

        \textit{The negative responses will be removed.} \\
        \midrule
        \textbf{Response with ID}  & You can place your backpack on the floor, to the left of the dining table-33. As for your bag, you can place it on the floor, to the left of the bed-10.
        &
        You can place your backpack on the floor, to the left of the dining table. As for your bag, you can place it on the floor, to the left of the bed.
        \\
        \bottomrule
    \end{tabular}
    \label{tab:dialogue_refinement}
\end{table}

\begin{table}[t!]
    \centering
    \small
    \caption{\textbf{Examples of QA refinement}.}
    \begin{tabular}{l|p{4.5cm}|p{4.5cm}}
        \toprule
        \textbf{Types} & \textbf{Raw Responses} & \textbf{Refined Responses}\\
        \midrule
        \textbf{Object Counting} & Q: How many chairs are in the room?

        A: 3
        & Q: How many chairs are in the room?

        A: four \\
        \midrule
        \textbf{Object Existence} & Q: Is there a mirror in the room?
        
        A: yes
        & Q: Is there a mirror in the room?
        
        A: no
        \\
        \midrule
        \textbf{Object Non-existence} & Q: Is there an ironing board in the room?
        
        A: no
        &Q: Is there an ironing board in the room?
        
        A: yes
        \\
        \midrule
        \textbf{Negative Response} &  
        Q: What is the material of the bathtub?
        A: unknown

        Q: Where is the shampoo dispenser?
        A: unknown
        &
        \textit{The negative responses will be removed.} \\
        \midrule
        \textbf{Response with ID}  & Q: Where is the mirror located?

        A: attached to wall-3, behind heater-18, to the left of shelf-19
        &
        Q: Where is the mirror located?

        A: attached to a wall, behind a heater, to the left of a shelf
        \\
        \bottomrule
    \end{tabular}
    \label{tab:qa_refinement}
\end{table}

\subsection{Subgraph Sampling}\label{app:subgraph_sampling}
\begin{figure}[t!]
\centering
\includegraphics[width=0.6\textwidth,height=5.5cm]{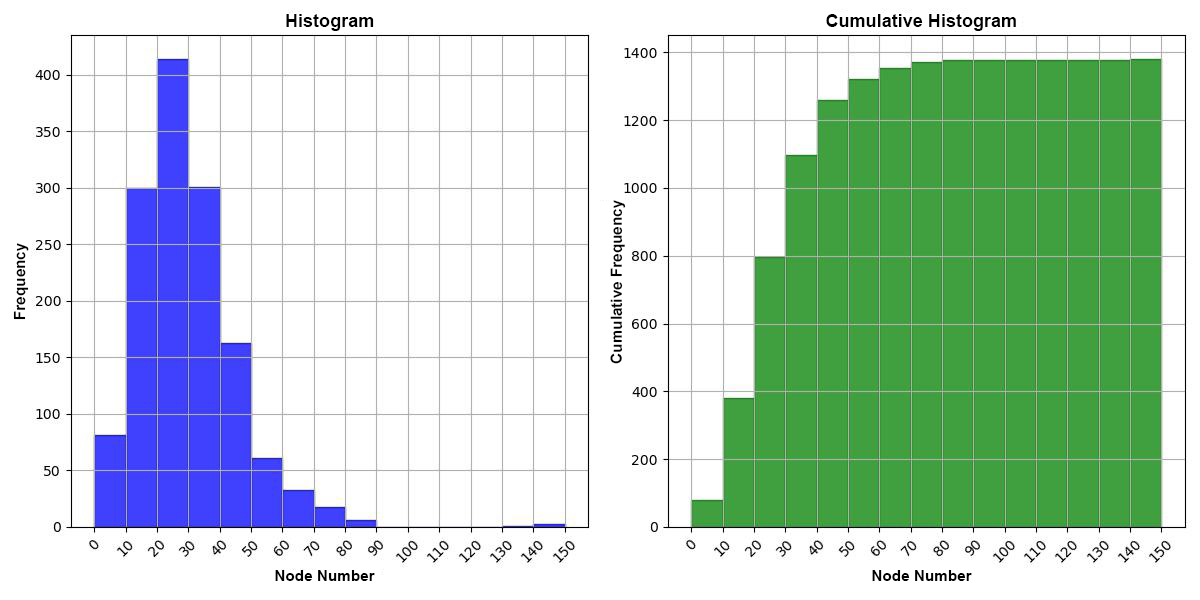}%
  \caption{\textbf{The distribution of node numbers for 3DSSG scenes.} The node number represents the number of objects in a scene.
  }
  \label{fig:node_distribution}
\end{figure}

To enhance the diversity of the 3D scene graphs used for prompting, we perform subgraph sampling on the 3DSSG according to a sampling rate, which denotes the ratio of preserved nodes. The sampled subgraphs are used for generating scene captions and planning data. We analyze the distribution of node numbers across the 3DSSG dataset in \cref{fig:node_distribution} and set different sampling rates for scenes with different numbers of nodes in \cref{tab:sampling_rate}. For each sampling rate, we set 4 random prompt seeds to further enhance the diversity of prompted data.

\begin{table}[t!]
    \centering
    \small
    \caption{\textbf{Sampling rates for scenes with different node numbers.} The hyphen denotes a sweep of sampling rates, \eg, ``0.7-0.9'' means ``0.7,0.8,0.9''.}
    \begin{tabular}{l|c c c c c c c}
        \toprule
        \textbf{Node Number} & \textbf{10-20} & \textbf{20-30} & \textbf{30-40} & \textbf{40-50} & \textbf{50-60} &\textbf{60-70} &\textbf{>70} \\
        \midrule
        Sampling Rate & 0.8,0.9 & 0.7-0.9 & 0.6-0.9 & 0.6-0.9 & 0.5-0.9 & 0.5-0.9 & 0.4-0.9\\
        \bottomrule
    \end{tabular}
    \label{tab:sampling_rate}
\end{table}

To verify whether the subgraph sampling strategy can maintain the consistency and diversity of scene captions, we generate scene captions for the same scene using both the full graph and subgraph. We then employ GPT-4 \citep{openai2023gpt4} to evaluate the similarities and differences between the two captions.
The results in~\cref{tab:subgraph_analysis} indicate that our subgraph sampling strategy can maintain both consistency and diversity.

\begin{table}[t!]
    \centering
    \small
    \caption{\textbf{The comparative analysis of full graph prompting and subgraph prompting.}}
    \begin{tabular}{l|p{10cm}}
        \toprule
        \textbf{Settings} & \textbf{Generated Scene Captions}\\
        \midrule
        \textbf{Full Graph} & In this scene, there is a colorful and flat white wall. A small TV stand is positioned in front of a low upholstered couch, to the left of a brown lamp, and to the right of a wooden table. The couch is covered with a pink padded square pillow and a patterned blanket. A light hangs on the wall and another light is supported by the table. There is a rectangular glass window attached to the wall, behind which is a yellow curtain. The room has a comfortable and cozy atmosphere, suitable for relaxation and entertainment. \\
        \midrule
        \textbf{Sampling Rate 0.6} & In this scene, there is a colorful white wall attached to a brown wooden floor. The couch is upholstered and low in size, located to the right of a small TV stand and a table. A red blanket covers the couch, and there are patterned, padded square pillows on the couch. A black glass window is attached to the wall, and there are artificial lamps standing on the floor and close to the couch. The lamps are also near a table and to the right of the TV stand. The scene suggests a comfortable living room with a cozy couch, pillows, and a blanket, suitable for relaxation and entertainment. \\
        \midrule
        \textbf{GPT-4 Evalutaion} &  \textbf{Similarities:} 1. Both scenes describe a living room setting, emphasizing comfort and coziness suitable for relaxation and entertainment.
2. Both rooms have a white wall.
3. Each scene features a couch and a TV stand.
4. Both scenes describe a blanket and a padded square pillow on the couch.

\textbf{Differences:} 1. The first scene has a brown wooden floor, while the floor is not mentioned in the second scene.
2. The first scene has a red blanket on the couch; the second has a patterned blanket but doesn't specify the color.
3. The first scene describes the lamps as ``standing on the floor'', while the second mentions one light hanging on the wall and another supported by the table.
4. The second scene includes a yellow curtain behind the window, which the first scene does not mention.

\textbf{Summary:} Overall, both summaries provide a similar thematic view of a comfortable living room but differ in the specific arrangement and color details of the items within the room.
\\
        \bottomrule
    \end{tabular}
    \label{tab:subgraph_analysis}
\end{table}

\subsection{Scene-graph-based Prompting \vs Box-based Prompting}
\label{sec:scene graph prompting and bbox prompting}

\begin{figure}[t!]
\centering
\includegraphics[width=\textwidth, keepaspectratio]{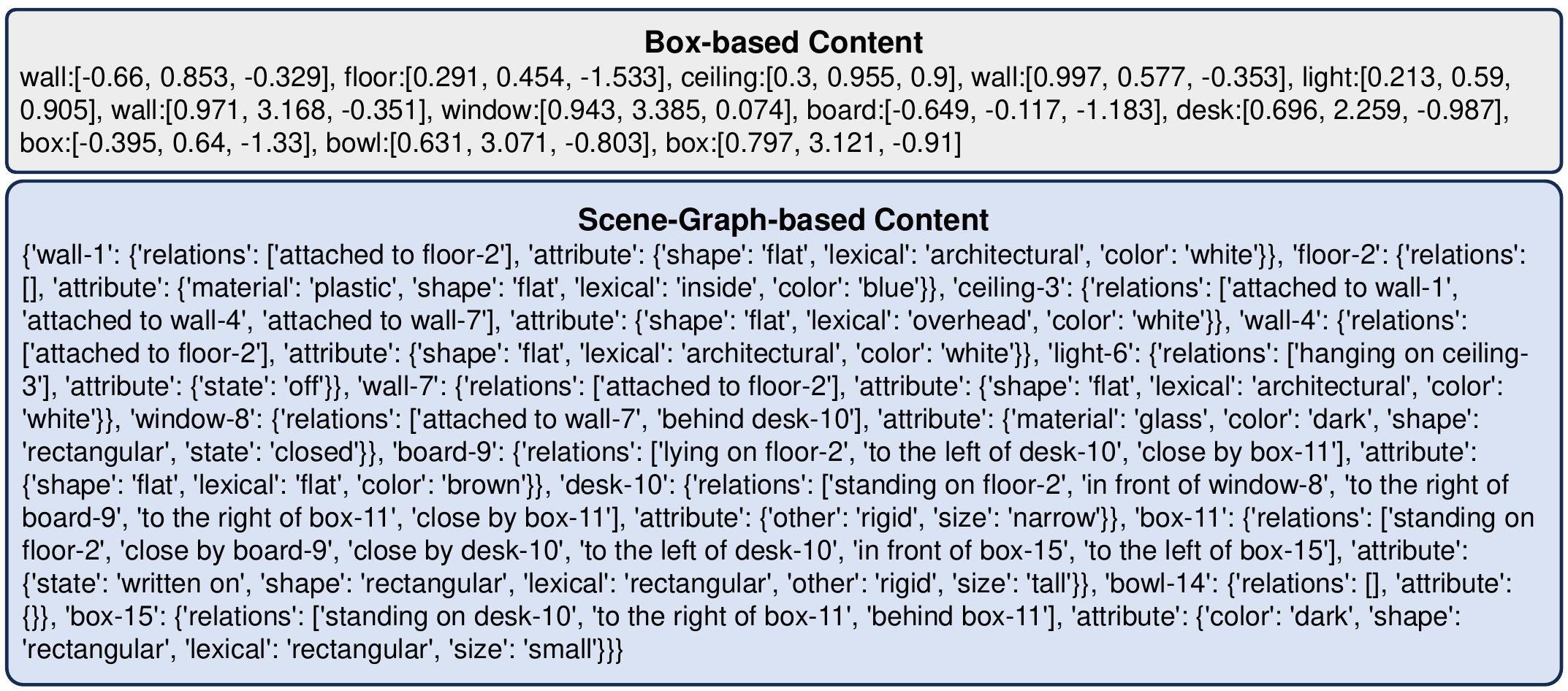}%
  \caption{Comparison of the content between box-based and scene-graph-based prompting.}
  \label{fig:prompt:compare_content}
\end{figure}

\begin{figure}[t!]
\centering
\includegraphics[width=\textwidth, keepaspectratio]{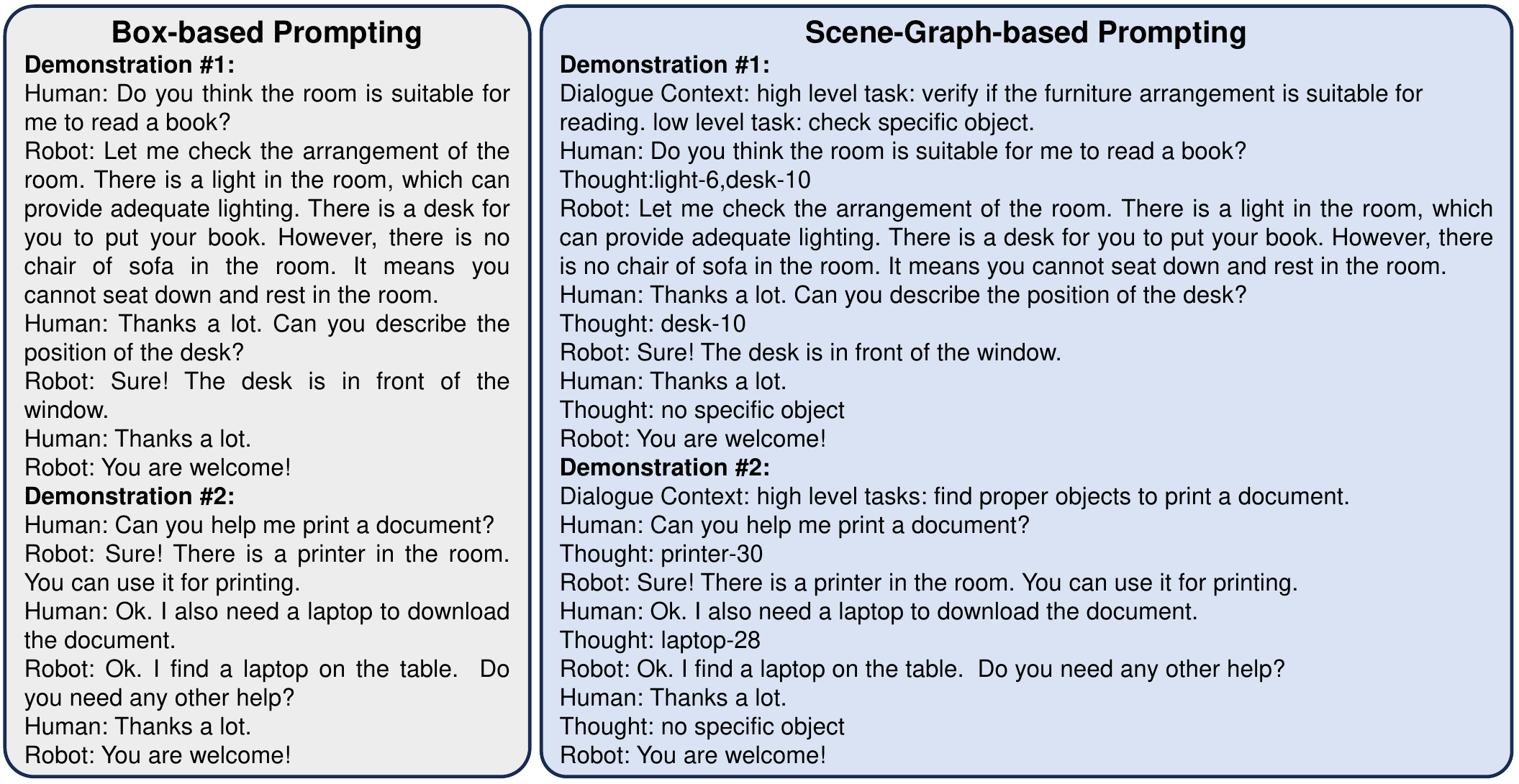}%
  \caption{The demonstrations for prompting, which are designed to be similar for a fair comparison.}
  \label{fig:prompt:compare_demonstrations}
\end{figure}

\begin{figure}[!t]
\centering
\includegraphics[width=\textwidth, keepaspectratio]{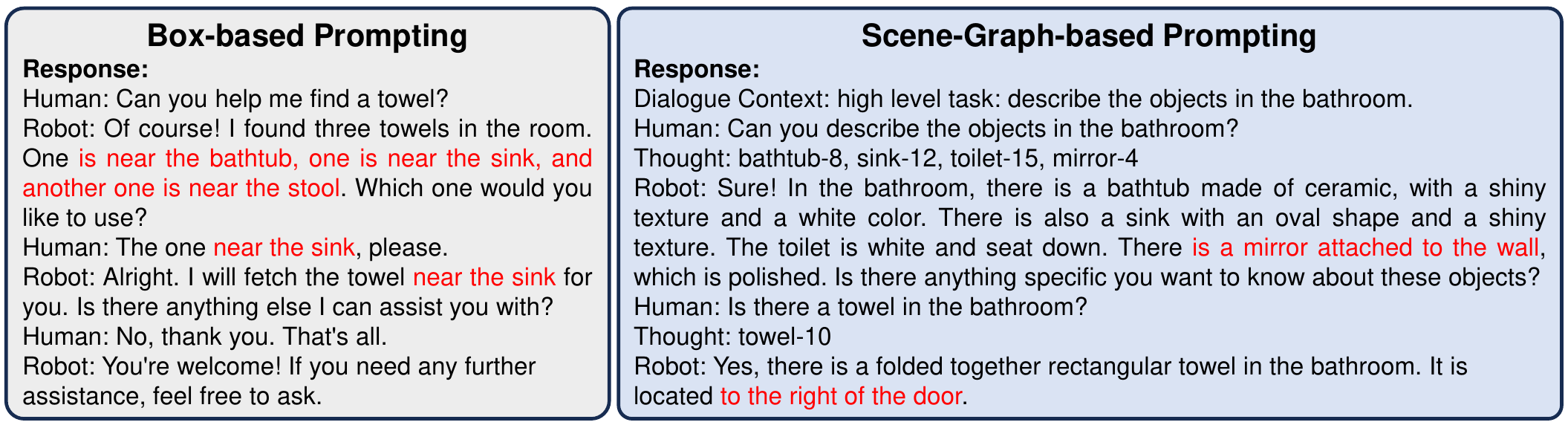}%
  \caption{The responses of two prompting methods. Descriptions highlighted in {\color{red}{red}} show our method leads to more flexible and reliable spatial relations.}
  \label{fig:prompt:compare_responses}
\end{figure}

In this section, we provide a comparative analysis of scene-graph-based prompting and box-based prompting \citep{hong20233d}. We refer the readers to Figure 6 in 3D-LLM \citep{hong20233d} for details of the box-based prompting method. \cref{fig:prompt:compare_content} shows the contents of two methods. To present a fair comparison between the two methods, we prompt with 1) demonstrations that have similar content under the same scene (see \cref{fig:prompt:compare_demonstrations}) and 2) identical new scene queries. Since 3D-LLM does not elaborate on attribute-related prompts, we mainly compare the spatial relations in the responses. As shown in \cref{fig:prompt:compare_responses}, we highlight some spatial relations in {\color{red}{red}}. The comparison shows that our method provides more diverse and reliable spatial relations, which are important for 3D scene understanding.

\subsection{Additional Comparision Regarding Dataset Quality}
\label{app:additional_data_comparison}
In addition to assessing the factual accuracy of responses compared to 3D-LLM, we also compared the grammatical correctness of the responses with ScanScribe\cite{zhu20233d}, a template-based synthetic dataset that focuses on 3D object caption. We observed that their dataset exhibited some grammar errors, whereas our dataset did not manifest such issues. We provide some data examples in \cref{tab:grammar_compare_1} and \cref{tab:grammar_compare_2}. We highlighted the grammar errors present in ScanScribe dataset in {\color{red}{red}}. Through comparison, it is evident that our sentences exhibit accurate and natural syntax, and also surpasses ScanScribe in the diversity and complexity of object descriptions.

\begin{table}[htbp]
\centering
\caption{Object captions in the 3Rscan scene 8f0f144b-55de-28ce-8053-2828b87a0cc9.}
\label{tab:grammar_compare_1}
\begin{tabular}{@{}l|l|l|p{10cm}}
\toprule
object label-id & method     & response id & caption                                                                                      \\ \midrule
microwave-8     & ours       & 1           & The microwave is black in color and is standing on the table. It is located behind and to the left of the toaster. \\ \cmidrule(lr){3-4}
                &            & 2           & The microwave is black in color. It is standing on a table and is positioned behind and to the left of a toaster.   \\ \cmidrule(lr){2-4}
                & ScanScribe & 1           & This is a black microwave. It is standing on the table. It is close by the toaster. A black microwave is in front of {\color{red}{the this}} microwave. \\ \cmidrule(lr){3-4}
                &            & 2           & This is a black microwave. It is behind the toaster. It is standing on the table. A black microwave is in front of {\color{red}{the this}} microwave.   \\ \bottomrule
\end{tabular}
\end{table}

\begin{table}[htbp]
\centering
\caption{Object captions in the 3Rscan scene 0cac753c-8d6f-2d13-8e27-e0664fc33bb9.}
\label{tab:grammar_compare_2}
\begin{tabular}{@{}l|l|l|p{10cm}}
\toprule
object label-id & method     & response id & caption                                                                                      \\ \midrule
counter-4    & ours       & 1           & The counter is standing on a wooden floor. It is close to 4 stools and an armchair. The stools are simple and wide in size, while the armchair is narrow in size. The floor is wooden and flat, and it is clean. \\ \cmidrule(lr){3-4}
                &            & 2           & The counter is standing on the wooden floor. It is close to 4 stools and an armchair. The floor is made of wooden material and is flat in shape. It is also clean. The armchair is narrow in size while the stools are simple in design. Three of the stools are wide in size.   \\ \cmidrule(lr){2-4}
                & ScanScribe & 1           & This is a counter. It is close by the armchair. It is close by the stool. A counter is close by the this counter. A counter is close by {\color{red}{the this}} counter.
                         \\ \bottomrule
\end{tabular}
\end{table}

\subsection{Dataset Statistics}
\label{app:dataset statistics}

\begin{figure}[t!]
    \centering
    \begin{minipage}[b]{0.45\textwidth}
        \includegraphics[width=1\textwidth]{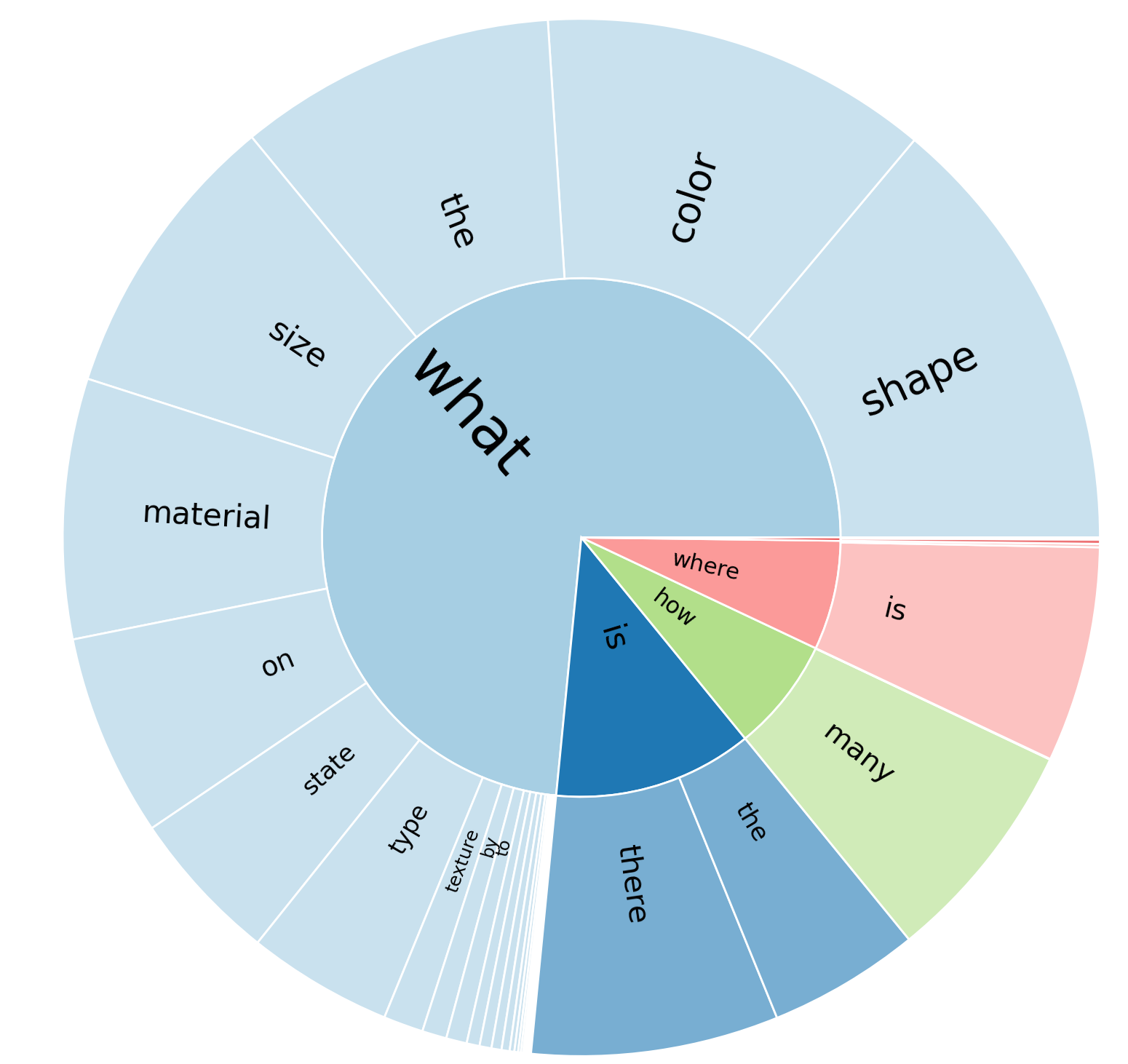}
        \caption{Question types: 3RQA.}
        \label{fig:RscanQA_statistics}
    \end{minipage}
    \hfill
    \begin{minipage}[b]{0.467\textwidth}
        \includegraphics[width=1\textwidth]{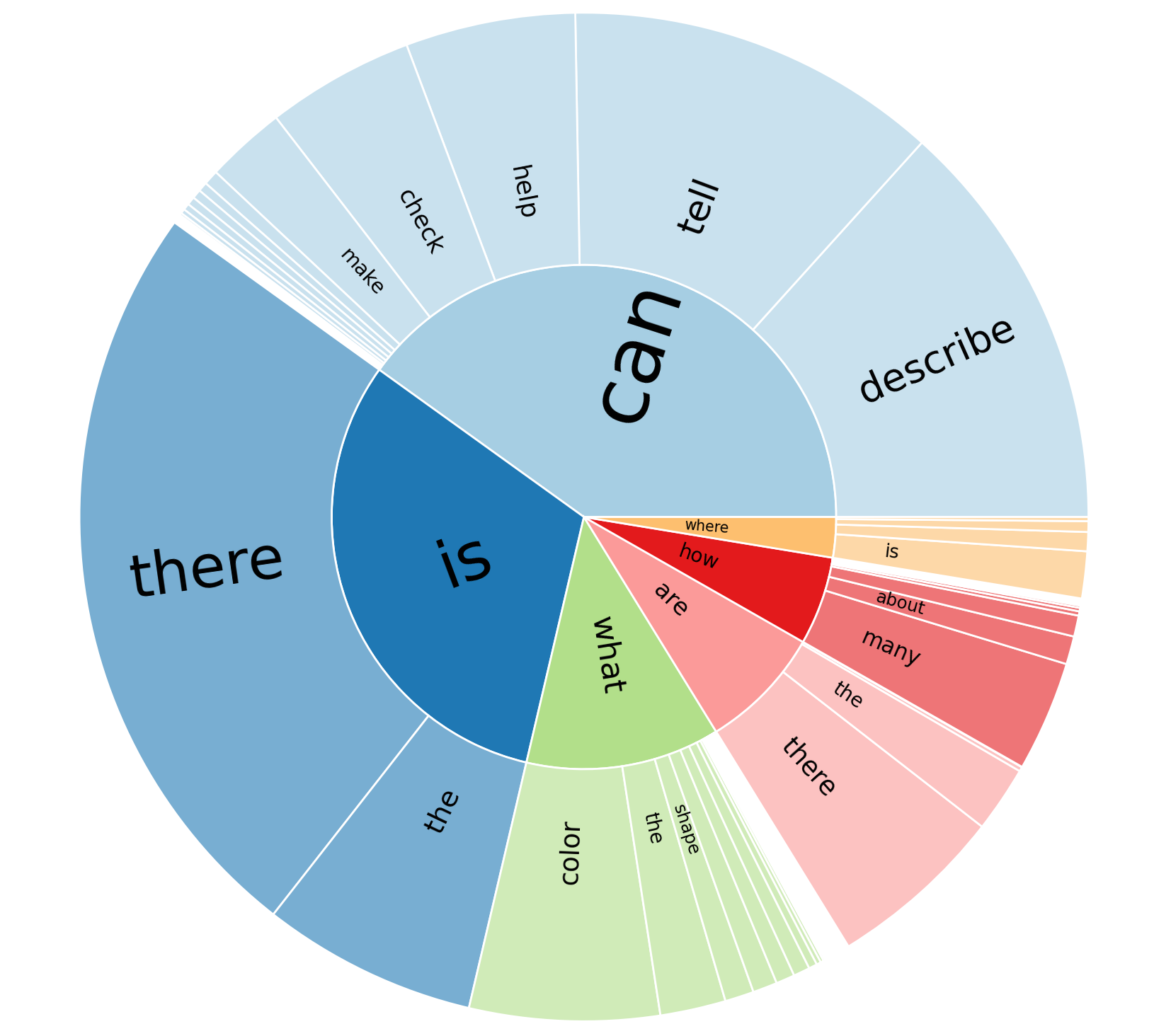}
        \caption{Question types: 3RDialog.}
        \label{fig:RscanDialog_Q_statistics}
    \end{minipage}
\end{figure}

\begin{figure}[t!]
    \centering
    \begin{minipage}[b]{0.45\textwidth}
        \includegraphics[width=1\textwidth]{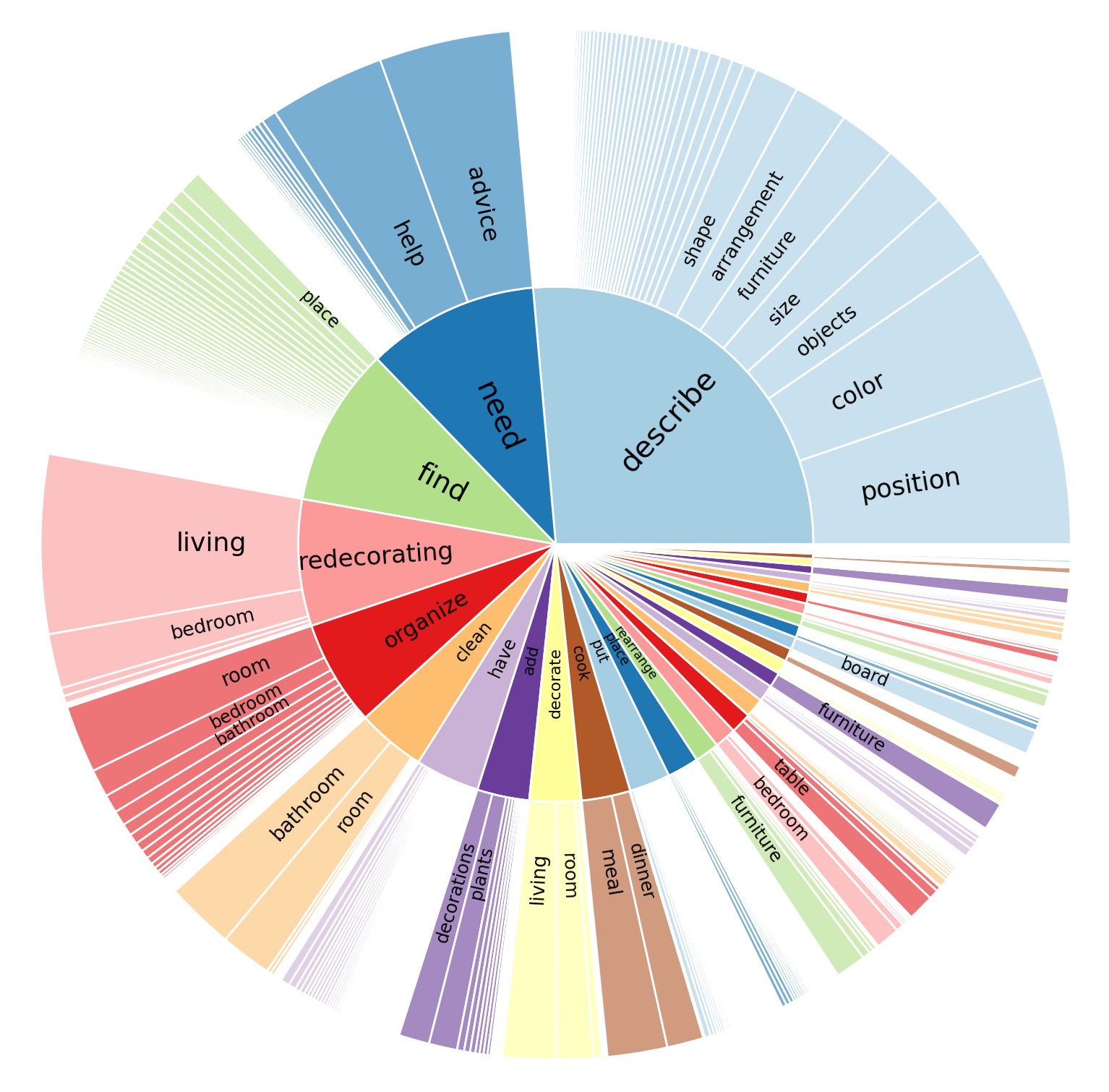}
        \caption{Noun-verb pairs: 3RDialog instruction.}
        \label{fig:RscanDialog_noun_verb_instruction}
    \end{minipage}
    \hfill
    \begin{minipage}[b]{0.46\textwidth}
        \includegraphics[width=1\textwidth]{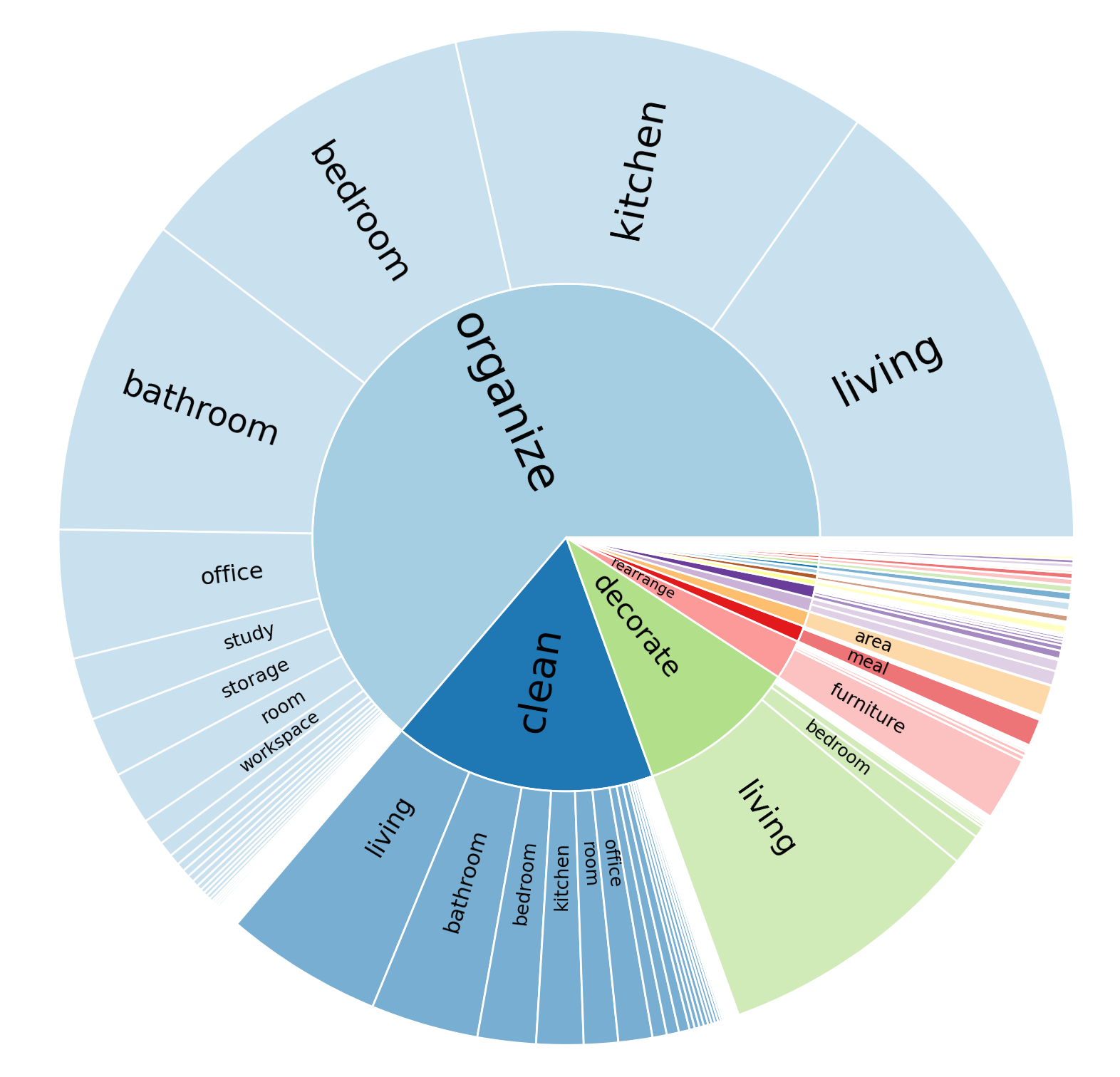}
        \caption{Noun-verb pairs: 3RPlan instruction.}
        \label{fig:RscanPlan_noun_verb_instruction}
    \end{minipage}
\end{figure}   

\begin{figure}[t!]
    \centering
    \begin{minipage}[b]{0.45\textwidth}
        \includegraphics[width=1\textwidth]{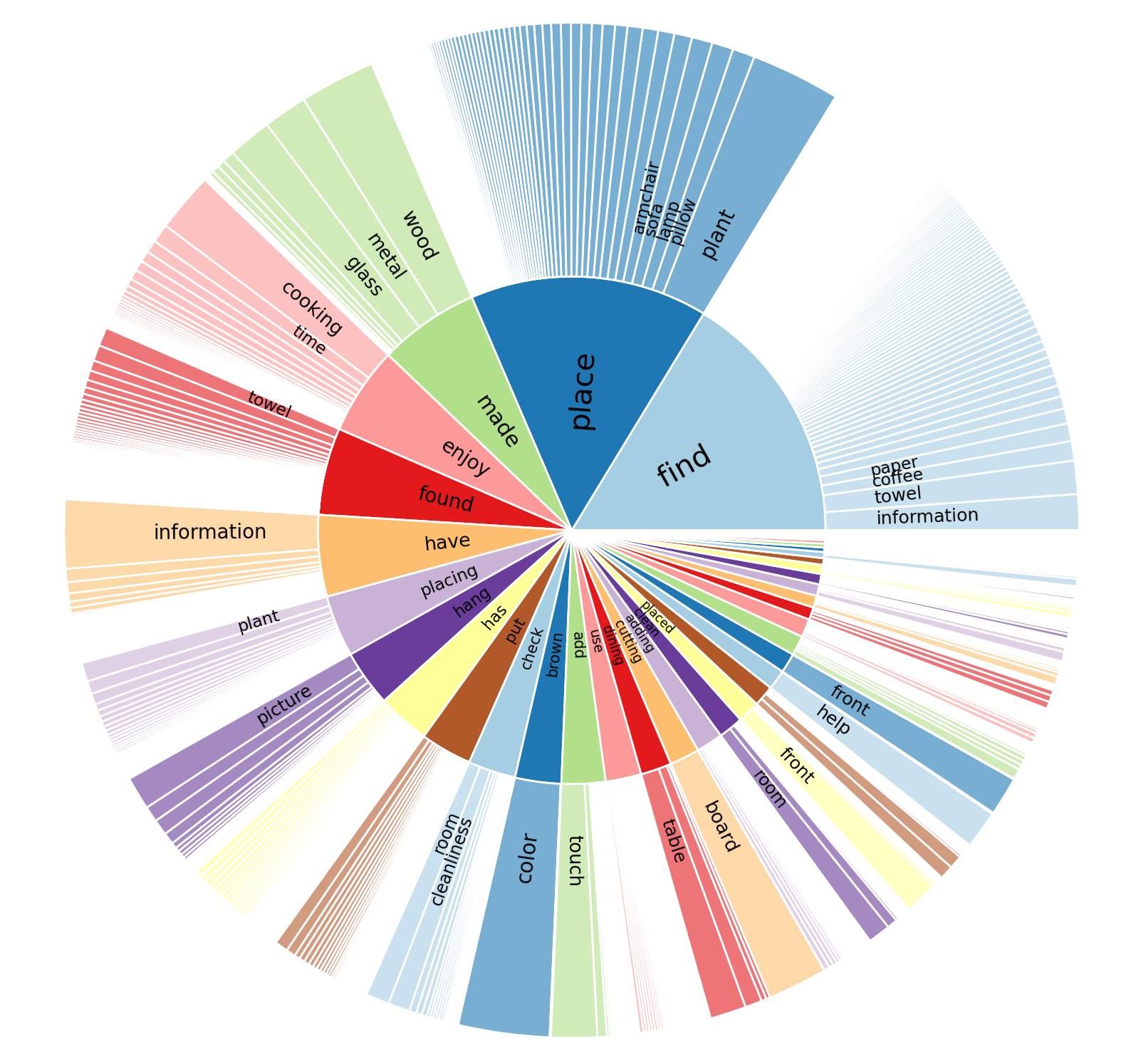}
        \caption{Noun-verb pairs: 3RDialog response.}
        \label{fig:RscanDialog_noun_verb_response}
    \end{minipage}
    \hfill
    \begin{minipage}[b]{0.433\textwidth}
        \includegraphics[width=1\textwidth]{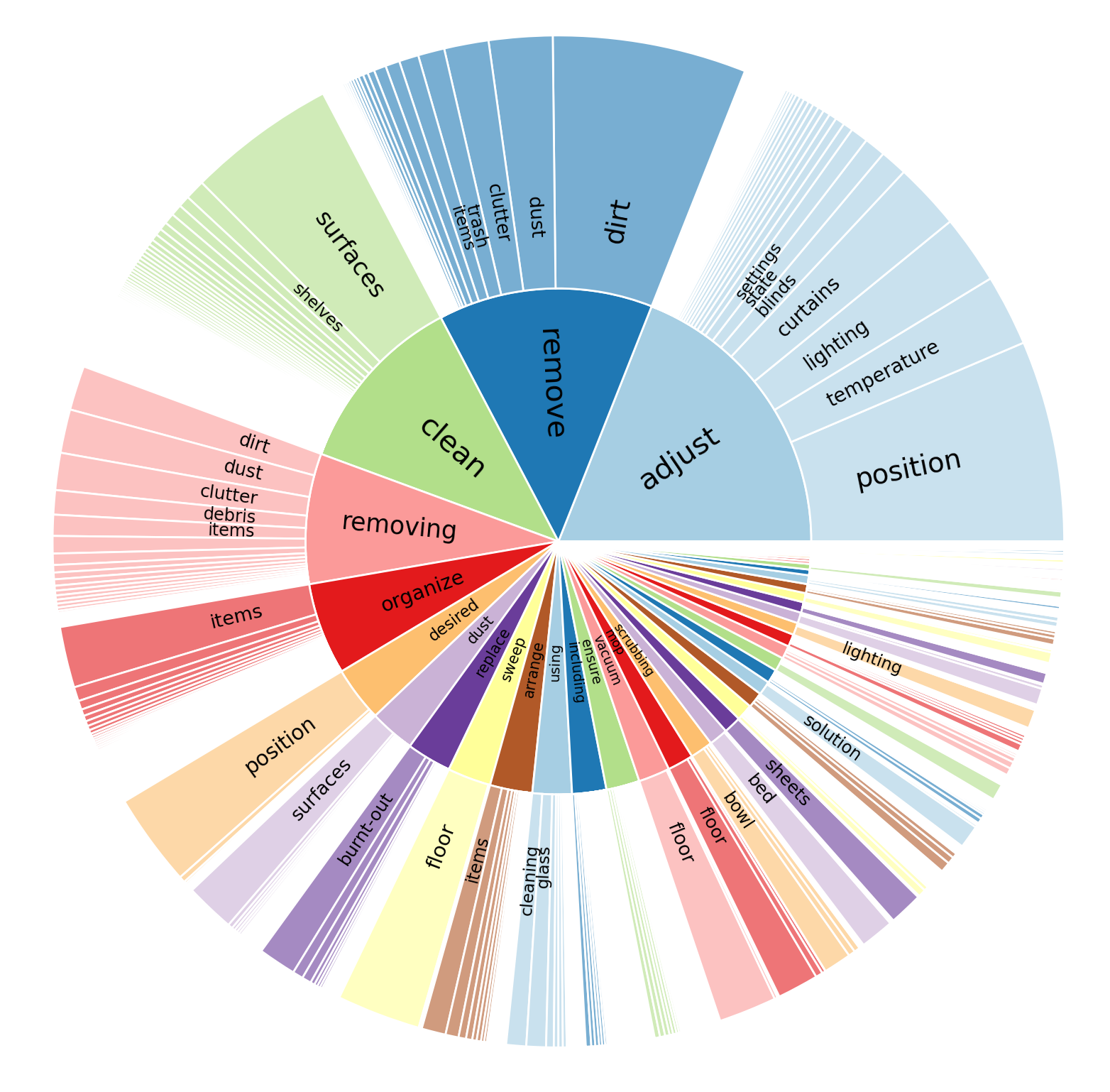}
        \caption{Noun-verb pairs: 3RPlan response.}
        \label{fig:RscanPlan_noun_verb_response}
    \end{minipage}
\end{figure}   

We provide statistics on the instruction-tuning datasets. We visualize the distribution of the question types in 3RQA (\cref{fig:RscanQA_statistics}) and 3RDialog (\cref{fig:RscanDialog_Q_statistics}). The pie chart's inner circle represents the first word of the questions, while the outer circle accounts for the second or third word in the corresponding questions. The results show that the questions cover the attributes and spatial relations of the objects, as well as high-level topics such as room types and functionalities.

We also provide statistics of the root noun-verb pairs for instructions and responses in 3RDialog and 3RPlan, as shown in \crefrange{fig:RscanDialog_noun_verb_instruction}{fig:RscanPlan_noun_verb_response}.

\section{Data Examples}\label{sec:supp_leo_ds_examples}
Please refer to \crefrange{tab:supp_data_example}{tab:supp_data_example_cont2} for examples of our dataset.

\section{Model Details}\label{app:model}
\subsection{Prompts}
The first portion of prompts sent into the LLM is a \textbf{system message}. It consists of two parts: a role prompt and a situation prompt. The role prompt is the same for all tasks:
\begin{tcolorbox}
\begin{minipage}{\linewidth}
You are an AI visual assistant situated in a 3D scene. You can perceive (1) an ego-view image (accessible when necessary) and (2) the objects (including yourself) in the scene (always accessible). You should properly respond to the USER's instructions according to the given visual information.
\end{minipage}
\end{tcolorbox}

The situation prompt begins with a common sentence:

\begin{tcolorbox}
\begin{minipage}{\linewidth}
You are at a selected location in the 3D scene.
\end{minipage}
\end{tcolorbox}

For SQA3D~\citep{ma2023sqa3d}, the situation prompt is further extended with the situation description in the dataset. The situation prompt is only used jointly with the embodiment token to support tasks that require information about the embodiment. Details can be found in \cref{sec:supp_embodiment}.

Next are the \textbf{visual tokens}, including \textbf{2D image tokens} and \textbf{object-centric 3D tokens}. Each token sequence is interleaved within text tokens and starts with a text prefix.
\begin{tcolorbox}
\begin{minipage}{\linewidth}
Ego-view image: \{\texttt{IMAGE\_TOKENS}\} \\
Objects (including you) in the scene: \{\texttt{OBJECT\_TOKENS}\}
\end{minipage}
\end{tcolorbox}

The last portion of prompts is a \textbf{task-specific instruction}. For \textbf{object-level caption} and \textbf{object-in-the-scene caption}, we randomly chose one sentence from 151 sentences to be the instruction. Some examples can be found in \cref{tab:obj-level_cap}. For \textbf{scene-level caption}, we randomly choose one from 183 instructions. Examples can be found in \cref{tab:scene-level_cap}. For \textbf{3D question answering} task, we simply use the question as the instruction. The dialog history is used as the instruction for \textbf{3D dialogue} to provide continuity across multiple rounds of interactions. A planning instruction pool consisting of 202 instructions is introduced for \textbf{scene-aware task planning} and we randomly choose one from it as done in the caption tasks. Examples from the pool can be found in \cref{plan cap}. The chosen instruction is further followed by an instruction that specifies the task, \eg, \textit{set up a home office}.

With past action tokens \{\texttt{PAST\_ACTIONS}\} appended at the end, the instruction for \textbf{embodied navigation} is as follows, where \{\texttt{GOAL}\} stands for the goal specified by the target object name:
\begin{tcolorbox}
\begin{minipage}{\linewidth}
The task is navigation. Your goal is to find \{\texttt{GOAL}\} by moving around in the scene. Past actions: \{\texttt{PAST\_ACTIONS}\}.
\end{minipage}
\end{tcolorbox}

The instruction for \textbf{robotic manipulation} is similar to the one in \textbf{embodied navigation}. Here \{\texttt{GOAL}\} is the task description in CLIPort:
\begin{tcolorbox}
\begin{minipage}{\linewidth}
The task is manipulation. Your goal is to \{\texttt{GOAL}\}. Past actions: \{\texttt{PAST\_ACTIONS}\}.
\end{minipage}
\end{tcolorbox}

\begin{table}[t!]
\caption{\textbf{Examples from our object-level caption instruction set.}}
\vspace{-4pt}
\begin{tcolorbox}
\begin{minipage}{\linewidth}
"Produce a description for the object at the chosen spot in the 3D scene.",\\
    "How would you depict the object located at the selected point in the 3D environment?",\\
    "Formulate a description of the item at the picked position within the 3D scene.",\\
    "How would you describe the entity at the designated location in the 3D backdrop?",\\
    "Can you detail the object situated at the selected point in the 3D setting?",\\
    "Compose a narrative for the object at the chosen locale within the 3D environment.",\\
    "What does the object at the specified position in the 3D visualization look like?",\\
    "Provide a description for the item located at the marked site in the 3D world.",\\
    "How would you illustrate the object placed at the selected spot in the 3D landscape?",\\
    "Craft a depiction of the object at the pinpointed location within the 3D territory.",\\
    "What kind of object is illustrated at the identified site in the 3D tableau?",\\
    "Develop a description of the object at the specified position in the 3D backdrop.",\\
    "What is the entity's detail at the highlighted site in the 3D view?",\\
    "Write up a description of the entity at the selected spot in the 3D realm.",\\
    "What does the object look like at the pinpointed location in the 3D space?",\\
    "Detail the entity located at the chosen position within the 3D scene.",\\
    "Can you explain the essence of the object at the selected spot in the 3D zone?",
\end{minipage}
\end{tcolorbox}

\label{tab:obj-level_cap}
\end{table}
\begin{table}[t!]
\caption{\textbf{Examples from our scene-level caption instruction set.}}
\vspace{-4pt}
\begin{tcolorbox}
\begin{minipage}{\linewidth}
    "Describe this scene.",\\
    "Generate a description of this scene.",\\
    "Generate a caption of this scene.",\\
    "Can you describe the scene?",\\
    "Can you generate a description of the scene?",\\
    "Can you generate a caption of the scene?",\\
    "Summarize this scene.",\\
    "Provide an outline of this 3D scene's characteristics.",\\
    "How would you describe the 3D scene?",\\
    "How would you summarize this scene?",\\
    "Convey a summary of the 3D structure of this scene.",\\
    "How would you interpret this 3D scene?",\\
    "Offer a summary of the 3D scene.",\\
    "Can you describe this scene in detail?",\\
    "I'm interested in this scene, can you explain?",\\
    "What is this scene made of?",\\
    "Could you provide more info about this scene?",
\end{minipage}
\end{tcolorbox}

\label{tab:scene-level_cap}
\end{table}

\begin{table}[t!]
\caption{\textbf{Examples from our planning instruction pool.}}
\vspace{-4pt}
\begin{tcolorbox}
\begin{minipage}{\linewidth}
    "Plan for the task",\\
    "Can you come up with a plan for this task",\\
    "How can we do this task, provide a step-by-step plan",\\
    "Draft a plan for completing this task",\\
    "Detail a strategy for the task",\\
    "What's the best plan for this task",\\
    "Draw out a procedure for the task",\\
    "Lay out the steps for this task",\\
    "Could you devise a plan for the task",\\
    "Show me a plan for this task",\\
    "I need a plan for the task",\\
    "Sketch a plan for the task at hand",\\
    "Set up a plan for this",\\
    "Recommend a plan for this task",\\
    "Offer a strategy for this task",\\
    "Design a blueprint for the task",\\
    "Outline the approach for this task",
\end{minipage}
\end{tcolorbox}

\label{plan cap}
\end{table}

\subsection{Feature Encoding}\label{sec:supp_embedding}
We have several modules to encode the multi-modal features.
\begin{itemize}[leftmargin=*]
    \item \textbf{Object-centric 3D token embedding.} The encoder for 3D object-centric point clouds is a PointNet++~\citep{qi2017pointnet++} pre-trained on ScanNet~\citep{dai2017scannet} with object-classfication task. We sample 1024 points for every object as in~\cite{chen2022language}. The architecture parameters all remain the same with~\cite{chen2022language}. We freeze the PointNet++ for empirically better results. %

    \item \textbf{Spatial Transformer~\citep{chen2022language}.} Spatial Transformer is a modified transformer architecture that explicitly encodes spatial relations between object pairs. Specifically, consider the vanilla self-attention~\citep{vaswani2017attention} mechanism which takes as input a feature matrix $X\in \mathbf{R}^{N\times d}$, where $N$ stands for the number of tokens and $d$ is the feature dimension. Vanilla self-attention first compute $Q=XW_Q, K=XW_K, V=XW_V$ from $X$ using learnable projection matrices $W_Q, W_K, W_V\in \mathbf{R}^{d\times d_h}$ where $d_h$ stands for the output feature dimension. Then the attention weight matrix is computed by $(\omega^o_{ij})_{N\times N} = \Omega^o = softmax(\frac{QK^T}{\sqrt{d_h}})$ and finally used for re-weighting $\Omega^oV$. The intuition of Spatial Transformer is that we can re-scale the elements $\omega_{ij}^o$ in the weight matrix $\Omega^o$.
    
    In the object-centric reasoning setting, the input feature matrix is $O\in \mathbf{R}^{N\times d}$. Consider an object pair $(O_i, O_j)$ with their geometric centers $c_i, c_j$. Spatial Transformer~\citep{chen2022language} computes the Euclidean distance $d_{ij} = ||c_i-c_j||_2$ and the horizontal and vertical angles $\theta_h, \theta_v$ of the line connecting $c_i$ and $c_j$. The spatial feature between the two objects $(O_i, O_j)$ is a 5-dimensional vector $f_{ij} = [d_{ij}, \sin{(\theta_h)}, \cos{(\theta_h)}, \sin{(\theta_v)}, \cos{(\theta_v)}]$. To combine this feature with objects, the spatial attention computes $\omega^s_{ij} = g_i f_{ij}$ where $g_i=W_S^To_i$ is a 5-dimensional vector. The spatial attention further reweights the original self-attention weight matrix as
    $$
    \omega_{ij}=\frac{\sigma(\omega^s_{ij})exp(\omega^o_{ij})}{\sum_{l=1}^N\sigma(\omega^s_{il})exp(\omega^o_{il})}.
    $$
    Readers are referred to \cite{chen2022language} for more details. In summary, Spatial Transformer explicitly computes pairwise spatial relations and fuses them with vanilla self-attention to provide better spatial reasoning ability. We use a three-layer Spatial Transformer with 8 heads to process the object-centric features produced by PointNet++ and output object tokens for LLM. For other settings, We follow all the default hyperparameters in \cite{chen2022language}.
    
    \item \textbf{2D token embedding.} We use OpenCLIP ConvNext-base model~\citep{liu2022convnet} pre-trained on LAION2B~\citep{schuhmann2022laion} to process the egocentric 2D image.
    \item \textbf{CLIP semantic guidance.} To inject more semantics into visual tokens, we use the text encoder from CLIP~\citep{radford2021learning} to process the instruction tokens to obtain a global semantics feature. Next, we update the visual tokens via element-wise product between the CLIP semantics feature and each image \& object token embedding.
\end{itemize}

\subsubsection{Embodiment Encoding}\label{sec:supp_embodiment}
In addition to the egocentric 2D input, we introduce an embodiment token to help \agent reason in an embodiment-aware fashion. We find it useful to use it together with the situation prompt and 2D egocentric input. Specifically, an embodiment token $e$ is introduced in \textbf{embodied navigation}, \textbf{embodied reasoning}, and \textbf{object-in-the-scene caption} tasks. Specifically, $e$ is a learnable embedding that will be inserted into the 3D object list.

So what does embodiment information mean in these tasks? In \textbf{embodied navigation}, it means the agent's position and orientation in the scene, which can be derived from a GPS and a compass sensor. The orientation of the agent is further represented by a rotation which is Fourier-embedded and mapped to a feature vector $r$ by a linear layer. It is the same in \textbf{embodied reasoning} task.
In the \textbf{object-in-the-scene caption} task, we assume the agent is situated at the location of the object that is being referred to. Therefore, embodiment information also means the location of the referred object. We obtain this location by randomly choosing a spot inside the referred object bounding box. To sum up, we could simply treat the embodiment token as a special \textit{self object}, where its object embedding is learnable, and its location/orientation corresponds to the actual or assumed ``agent''.

After inserting the embodiment token, we obtain a new 3D object token list: $e, s_{\text{3D}}^{(1)}, s_{\text{3D}}^{(2)}, \dots, s_{\text{3D}}^{(N)}$, where $s_{\text{3D}}^{(i)}, i\in \{1, 2, \dots, N\}$ are 3D object token embeddings produced by PointNet++, along with location specified for each object (including the \textit{self-object}). We can concatenate them together to get a feature matrix $O\in \mathbf{R}^{(N+1)\times d}$ and send them to the Spatial Transformer to explicitly fuse the spatial information of all the 3D objects and the self-object.

\subsection{Action Tokenization}\label{sec:action_tokenization}

To empower \agent to exert control over an embodiment or a robot, we encode all actions within the context of Object Navigation~\citep{ramrakhya2022habitat} and CLIPort~\citep{cliport} tasks using the least frequently employed language tokens. Specifically, for the Object Navigation task, we allocate 4 tokens to represent actions of \textit{move forward}, \textit{turn right}, \textit{turn left}, and \textit{stop}. For the CLIPort task, we use a total of 516 tokens to discretize action poses, with 320 tokens dedicated to the x-axis pose bins, 160 tokens for the y-axis pose bins, and 36 tokens for the z-rotation bins.

\subsection{LLM Hyperparameters}
We set the maximum output length of our Vicuna-7B to be 256. The maximum context length is also set to 256 and if the length of the input is greater than 256, we truncate it to 256 by deleting tokens from the left (\ie, only the rightmost 256 tokens are preserved). We set rank and $\alpha$ in LoRA~\citep{hu2022lora} to be 16 and the dropout rate to be 0. LoRA is implemented for all the projection matrices in the LLM, \ie, $(W_q, W_k, W_v, W_o)$ in attention modules and $(W_{gate}, W_{up}, W_{down})$ in MLPs.

The hyperparameters for inference are listed in \cref{tab:parameter_beam}.

\section{Alignment Setup}
The hyperparameters for 3D VL alignment are presented in \cref{tab:param_align}.

\begin{table*}[t!]
\centering
\small
\caption{Hyperparameters for \agent inference.}\label{tab:parameter_beam}
\begin{tabular}{@{}ll@{}}
\toprule
\textbf{Hyperparameters}  & \textbf{Value}   \\ \midrule
Number of beams          & 5     \\
Maximum output length    & 256   \\
Minimum output length    & 1     \\
Top $p$                   & 0.9   \\ 
Repetition penalty       & 3.0   \\
Length penalty           & 1.0   \\
Temperature              & 1.0   \\
\bottomrule
\end{tabular}
\end{table*}

\begin{table*}[t!]
\centering
\small
\caption{Hyperparameters for the alignment stage.} \label{tab:param_align}
\begin{tabular}{@{}ll@{}}
\toprule
\textbf{Hyperparameter}  & \textbf{Value}   \\ \midrule
Optimizer                & AdamW     \\
Weight decay             & 0.05     \\
Betas                    & [0.9, 0.999]      \\
Learning rate            & $3\times 10^{-4}$ \\
Warmup steps             & 400      \\
Number of workers        & 4         \\
Parallel strategy        & DDP       \\
Type of GPUs             & NVIDIA A100       \\
Number of GPUs           & 4 \\
Accumulate gradient batches & 5      \\
Batch size per GPU (total)   & 4 (80)    \\ 
Training precision       & bfloat16 \\
Gradient norm            & 5.0        \\
Epochs                   & 5        \\
\bottomrule
\end{tabular}
\end{table*}

\section{Instruction-tuning Setup}
The hyperparameters for 3D VLA instruction tuning are presented in \cref{tab:param_sft}.
\begin{table*}[t!]
\centering
\small
\caption{Hyperparameters for the instruction-tuning stage.} \label{tab:param_sft}
\begin{tabular}{@{}ll@{}}
\toprule
\textbf{Hyperparameter}  & \textbf{Value}   \\ \midrule
Optimizer                & AdamW     \\
Weight decay             & 0.05     \\
Betas                    & [0.9, 0.999]      \\
Learning rate            & $3\times 10^{-5}$ \\
Warmup steps             & 400      \\
Number of workers        & 4         \\
Parallel strategy        & DDP       \\
Type of GPUs             & NVIDIA A100       \\
Number of GPUs           & 4 \\
Accumulate gradient batches & 5      \\
Batch size per GPU (total)   & 4 (80)    \\ 
Training precision       & bfloat16      \\
Gradient norm            & 5.0        \\
Epochs                   &  10        \\
\bottomrule
\end{tabular}
\end{table*}

\section{Ablation Details}\label{sec:supp_ablation}

\subsection{Object-centric Mask}
\paragraph{Ground truth \vs object proposals.} As we adopt an object-centric 3D representation, the object-centric masks are necessary to segment the scene point cloud. For scenes that lack annotations of object-centric masks, we can utilize off-the-shelf detection or segmentation models to generate object proposals and thus obtain the masks. We compare the performances of \agent (\textit{w/o Act}) between using ground-truth masks and Mask3D \citep{schult2022mask3d} proposals. The results in \cref{tab:mask3d_gap} indicate that using Mask3D proposals leads to a moderate performance drop on Scan2Cap (mainly due to the IoU@0.5 metrics) and comparable performances on QA tasks.

\begin{table}[t!]
\centering
\captionof{table}{Quantitative comparison between \agent (\textit{w/o Act}) using ground-truth masks and Mask3D proposals. Metrics follow \cref{tab:vl_results}.}
\resizebox{0.83\linewidth}{!}{
\begin{tabular}{lccccccccccc}
    \toprule
     & \multicolumn{5}{c}{Scan2Cap (val)} & \multicolumn{5}{c}{ScanQA (val)} & SQA3D (test) \\
     \cmidrule(lr){2-6} \cmidrule(lr){7-11} \cmidrule(lr){12-12}
     & C & B-4 & M & R & Sim & C & B-4 & M & R & EM@1 & EM@1 \\
    \midrule

    \textit{w/o Act} (Mask3D) & 72.4 & 38.2 & 27.9 & 58.1 & 55.3 & 101.4 & 13.2 & 20.0 & 49.2 & \textbf{24.5} \textcolor{gray}{(47.6)} & 50.0 \textcolor{gray}{(52.4)} \\

    \textit{w/o Act} (GT) & \textbf{87.4} & \textbf{44.5} & \textbf{30.8} & \textbf{65.7} & \textbf{65.4} & \textbf{103.0} & \textbf{14.6} & \textbf{20.1} & \textbf{49.7} & 24.3 \textbf{\textcolor{gray}{(48.5)}} & 50.0 \textbf{\textcolor{gray}{(52.5)}} \\
    \bottomrule
\end{tabular}
}
\label{tab:mask3d_gap}
\end{table}

\subsection{Model Ablation}\label{sec:model_ablation}
\paragraph{LLM.} Following the setting of \agent (\textit{w/o Act}), we ablate the default LLM (Vicuna-7B) with OPT-1.3B \citep{zhang2022opt} and Vicuna-13B \citep{vicuna2023}, respectively. We report the evaluation results on ScanNet and 3RScan tasks in \cref{tab:llm_ablation}. The results show a significant gap between OPT-1.3B and Vicuna-7B and comparable performances between Vicuna-7B and Vicuna-13B. This indicates the notable improvements when scaling from smaller LLM to 7B scale and the potential saturation if we continue to scale up, resembling the finding in \cref{sec:exp_scaling}.

\paragraph{Point cloud backbone.} We have tried substituting PointNet++ \citep{qi2017pointnet++} with Point-BERT \citep{yu2022point} as the point cloud backbone. Specifically, we utilize the Point-BERT checkpoint from PointLLM \citep{xu2023pointllm}, which has adapted Point-BERT to 6-channel (XYZRGB) input and learned a language-aligned representation for 3D objects. We have not observed notable difference between the performances of using Point-BERT and PointNet++ so we omit the results here.

\begin{table}[t!]
    \centering
    \captionof{table}{Quantitative results of \agent equipped with LLMs at different scales. Metrics follow \cref{tab:data_ablation}.}
    \setlength\tabcolsep{3pt}
    \resizebox{0.65\linewidth}{!}{
    \begin{tabular}{lcccccc}
        \toprule
     & \multicolumn{3}{c}{ScanNet} & \multicolumn{3}{c}{3RScan} \\
     \cmidrule(lr){2-4} \cmidrule(lr){5-7}
         & Scan2Cap & ScanQA & SQA3D & 3RQA & 3RDialog & 3RPlan \\
        \midrule
        \textit{w/o Act} (OPT-1.3B) & 64.6 & 20.3 \textcolor{gray}{(44.2)} & 45.5 \textcolor{gray}{(47.6)} & 50.0 \textcolor{gray}{(54.5)} & 71.1 & 78.3 \\
        \textit{w/o Act} (Vicuna-7B) & \textbf{65.4} & \textbf{24.3} \textcolor{gray}{(48.5)} & \textbf{50.0 \textcolor{gray}{(52.5)}} & 51.9 \textcolor{gray}{(57.4)} & \textbf{73.3} & \textbf{81.1} \\
        \textit{w/o Act} (Vicuna-13B) & 65.2 & 23.4 \textbf{\textcolor{gray}{(48.9)}} & 49.7 \textcolor{gray}{(52.3)} & \textbf{56.2 \textcolor{gray}{(60.4)}} & 72.5 & 80.5 \\
        \bottomrule
    \end{tabular}
    }
    \label{tab:llm_ablation}
\end{table}

\subsection{Dialogue and Planning Data}\label{sec:dialog_planning}
To evaluate \textit{w/o Dialg}, we design an evaluation set with three types of questions: 1) \textbf{Answerable}: general questions that can be answered based on the given 3D scenes; 2) \textbf{Unanswerable}: questions that cannot be answered given the 3D scenes due to a lack of information, \eg, ``Tell me about the elephant in the room''; 3) \textbf{NLP}: questions that solely examine the language functionality of \agent in term of factual knowledge, reasoning, and text coherence. We collect 30 representative questions for each subset and generate \agent's responses for each question. We then ask humans to choose their preferred responses between \textit{w/o Dialg} and \textit{w/ Dialg} Based on the human preferences, we evaluate the two models with TrueSkill~\cite{graepel2007bayesian}, which is an algorithm that quantifies players’ rating scores by Bayesian inference. The scores are estimated by Gaussian distribution and expressed as $\mu\pm\sigma$.

\subsection{Data Balancing}\label{sec:data_balancing}
To investigate the hallucination problem, we collect 150 questions querying object existence on 3RScan and ScanNet respectively. We split three subsets according to the category of queried object. The queried object can exist in the given scene (Yes),
exist in other scenes instead of the given scene (No-1), or not exist in all the scenes (No-2).
Each subset comprises 50 questions. We merge No-1 and No-2 when reporting the exact-match accuracy, as shown in \cref{tab:data_balance}.

\section{Evaluation Details}

\subsection{3D Question Answering}\label{sec:supp_eval_qa}
\paragraph{Rationality of QA evaluation protocol.} We argue that exact match (EM), as a conventional metric for 3D QA, is unsuitable for evaluating the open-ended answer generated by LLMs. For example, given the question ``\textit{On what side of the towel is a bathroom curtain}?'' with ground-truth answer ``\textit{left side of towel}'', it is never wrong to answer ``left''. However, this will be deemed incorrect if we adopt the strict exact match protocol. Such a misjudgment is quite likely to occur when evaluating the answers from LLMs. By contrast, the classifier heads for QA (\eg, MCAN) are less affected because they collect all possible answers in advance to formulate the QA as a close-set classification problem. Hence, we refine the strict exact match protocol as follows.

\begin{lstlisting}[language=Python]
"""
code for QA protocols
pred: str
gts: List[str]
"""

def strict_em(pred, gts):
    for gt in gts:
        if pred == gt:
            # case 1
            return True


def refined_em(pred, gts):
    for gt in gts:
        if pred == gt:
            # case 1
            return True
        elif ''.join(pred.split()) in ''.join(gt.split()):
            # case 2
            return True
        elif ''.join(gt.split()) in ''.join(pred.split()):
            # case 3
            return True
    return False
\end{lstlisting}

In a nutshell, we squeeze the \texttt{pred} and \texttt{gt}, and then check whether one is a subset of the other. To justify our refined exact match protocol, in \cref{tab:em_protocol_cases} we provide some representative examples in the ScanQA validation set. Despite the improvements, we speculate such a simple refinement is still insufficient for a sound evaluation metric considering the flexibility of human language.
\begin{table}[t!]
    \centering
    \caption{Examples from ScanQA validation set manifest the rationality of our refined exact match protocol.}
    \resizebox{\linewidth}{!}{
    \begin{tabular}{lllcc}
        \toprule
        Question & Ground-truth answer & Predicted answer & Strict EM & Refined EM \\
        \midrule
        What color is the chair in the kitchen? & dark brown & brown & \xmark & \cmark (case 2) \\
        What is under the long kitchen counter? & kitchen cabinets & brown rectangular kitchen cabinets & \xmark & \cmark (case 2) \\
        What type of refrigerator is on the right of a kitchen counter? & stainless steel refrigerator & stainless steel & \xmark & \cmark (case 2) \\
        Where is the beige wooden desk placed? & up against wall & against wall & \xmark & \cmark (case 2) \\
        What color does the sofa look? & it looks black & black & \xmark & \cmark (case 2) \\
        Where is the black office chair located? & in front of desks & in front of desk & \xmark & \cmark (case 2) \\
        What is in the corner by windows? & book shelf & bookshelf & \xmark & \cmark (case 2) \\
        Where is the chair pulled into? & table & under table & \xmark & \cmark (case 3) \\
        How many chairs are to the left of the table? & 4 & 4 chairs & \xmark & \cmark (case 3) \\
        What objects are sitting on the black couch? & pillow & pillows & \xmark & \cmark (case 3) \\
        Where are the two different size tables located in room? & in center & in center of room & \xmark & \cmark (case 3) \\
        Where is the laptop located? & desk & on desk & \xmark & \cmark (case 3) \\
        Where is the soap dispenser mounted & above sink & on wall above sink & \xmark & \cmark (case 3) \\
        \bottomrule
    \end{tabular}
    }
    \label{tab:em_protocol_cases}
\end{table}

\subsection{Embodied Navigation}\label{sec:supp_eai_split}
To construct our training set, we adopt all 57 scenes in the MP3D \texttt{ObjNav} training split~\citep{savva2019habitat,ramrakhya2022habitat} and generate \textasciitilde{}60K shortest-path navigation episodes. The evaluation is conducted on the original validation split of the MP3D \texttt{ObjNav} task and the newly introduced HM3D \texttt{ObjNav} task~\citep{ramakrishnan2021habitat}. 

In contrast to most \texttt{ObjNav} agents that utilize recurrence through either RNN~\citep{ramrakhya2022habitat} or DT-style Transformer~\citep{suglia2021embodied}, \agent only employs a simplistic feed-forward policy, \ie, the Transformer in \agent only takes in the instruction, current state (2D and 3D observation), and past 4 actions, and predicts the next action, similar to RT-2~\citep{brohan2023rt}. Therefore, the only information relayed from the past is past actions. The absence of recurrence in \agent's acting policy is indeed the result of a trade-off between better performances and training efficiency. We will commit to exploring the possibility of looping in more sophisticated policy architectures (\eg, recurrence) in future work.

\section{Additional Results}\label{sec:additional_results}

\subsection{Impact of Data Refinement}\label{sec:impact_data_refinement}
\paragraph{Settings.} We investigate the impact of data refinement by comparing the downstream performances between pretraining on the generated data before/after refinement. Specifically, since our generated data (where the refinement occurs) pertains to 3RScan scenes, we first pretrain the \agent after the alignment stage on a mix of 3RScan datasets, and then train on a mix of ScanNet datasets (Scan2Cap, ScanQA, and SQA), where we report the quantitative results as downstream performances.

\begin{table}[t!]
\centering
\captionof{table}{Quantitative comparison between \agent pretrained on the generated data before/after refinement. Metrics follow \cref{tab:vl_results}.}
\resizebox{0.83\linewidth}{!}{
\begin{tabular}{lccccccccccc}
    \toprule
     & \multicolumn{5}{c}{Scan2Cap (val)} & \multicolumn{5}{c}{ScanQA (val)} & SQA3D (test) \\
     \cmidrule(lr){2-6} \cmidrule(lr){7-11} \cmidrule(lr){12-12}
     & C & B-4 & M & R & Sim & C & B-4 & M & R & EM@1 & EM@1 \\
    \midrule

    Before refinement & 84.1 & \textbf{45.8} & 30.9 & 66.1 & 65.3 & 99.4 & 12.6 & 19.4 & 48.6 & 24.5 \textcolor{gray}{(49.1)} & 48.2 \textcolor{gray}{(50.5)} \\

    After refinement & \textbf{87.1} & 45.2 & \textbf{31.1} & 66.1 & \textbf{65.7} & \textbf{105.7} & \textbf{14.9} & \textbf{20.5} & \textbf{50.7} & \textbf{24.7 \textcolor{gray}{(49.8)}} & \textbf{52.4 \textcolor{gray}{(55.0)}} \\
    \bottomrule
\end{tabular}
}
\label{tab:impact_data_refinement}
\end{table}

The results in \cref{tab:impact_data_refinement} demonstrate that data refinement elicits consistent improvements. In particular, data refinement primarily benefits reasoning (QA) tasks, probably because the refinement operation mainly concerns QA and dialogue data.

\subsection{Data Comparison}\label{sec:data_comparison}
\paragraph{Settings.} We collect the training data of LL3DA \citep{chen2024ll3da} to train \agent and compare the quantitative results with \agent trained with our original data to showcase the impact of training data. We report the performances on Scan2Cap and ScanQA, where their data overlaps ours.

\begin{table}[t!]
\centering
\captionof{table}{Quantitative comparison between \agent trained on the LL3DA data and our data. Metrics follow \cref{tab:vl_results}.}
\resizebox{0.7\linewidth}{!}{
\begin{tabular}{lcccccccccc}
    \toprule
     & \multicolumn{5}{c}{Scan2Cap (val)} & \multicolumn{5}{c}{ScanQA (val)} \\
     \cmidrule(lr){2-6} \cmidrule(lr){7-11}
     & C & B-4 & M & R & Sim & C & B-4 & M & R & EM@1 \\
    \midrule

    LL3DA data & 73.9 & 43.5 & 30.2 & 65.0 & 63.4 & 99.7 & \textbf{14.8} & 19.7 & 47.8 & 22.9 \textcolor{gray}{(46.4)} \\

    Our data & \textbf{86.4} & \textbf{44.4} & \textbf{30.9} & \textbf{65.8} & \textbf{65.6} & \textbf{104.9} & 13.8 & \textbf{20.4} & \textbf{50.3} & \textbf{24.5 \textcolor{gray}{(49.2)}} \\
    \bottomrule
\end{tabular}
}
\label{tab:data_comparison}
\end{table}

The results in \cref{tab:data_comparison} exhibit a consistent performance gap between training on LL3DA data and our original data, underscoring the advantage of our collected training data.

\subsection{Model Comparison}\label{sec:model_comparison}
\paragraph{Settings.} \agent adopts an object-centric 3D representation to encode 3D scenes, which is a novel approach compared with recent works. For example, 3D-LLM \citep{hong20233d} leverages 2D foundation models to obtain dense semantic features and lift them to 3D space, and LL3DA \citep{chen2024ll3da} adopts scene-level encoding. They both use learnable queries to extract 3D features. Here we investigate the influence of model design with the same training data. For a fair comparison, we use Mask3D \citep{schult2022mask3d} object proposals instead of ground-truth masks for the evaluation results of \agent.

\paragraph{LL3DA \vs \agent.} We train \agent on the LL3DA training data and compare the performances with LL3DA generalist results (without task-specific fine-tuning). From the results in \cref{tab:ll3da_leo}, we highlight two takeaways: 1) with the same training data, \agent outperforms LL3DA on most metrics; 2) the gap between LL3DA and \agent is significant on ScanQA, which indicates a major advantage of object-centric 3D representation lies in handling the reasoning task.

\begin{table}[t!]
\centering
\captionof{table}{Quantitative comparison between LL3DA and \agent when both trained on LL3DA data. Metrics follow \cref{tab:vl_results}.}
\resizebox{0.73\linewidth}{!}{
\begin{tabular}{lcccccccccccc}
    \toprule
     & \multicolumn{4}{c}{Scan2Cap (val)} & \multicolumn{4}{c}{Nr3D (val)} & \multicolumn{4}{c}{ScanQA (val)} \\
     \cmidrule(lr){2-5} \cmidrule(lr){6-9} \cmidrule(lr){10-13}
     & C & B-4 & M & R & C & B-4 & M & R & C & B-4 & M & R \\
    \midrule

    LL3DA & 63.0 & 36.0 & 25.7 & 54.7 & \textbf{23.9} & \textbf{13.4} & 22.3 & 45.8 & 75.7 & 13.3 & 15.4 & 37.0 \\

    \agent & \textbf{64.9} & \textbf{37.2} & \textbf{27.4} & \textbf{57.5} & 22.1 & 10.9 & \textbf{22.9} & \textbf{46.3} & \textbf{99.2} & \textbf{14.9} & \textbf{19.4} & \textbf{47.3} \\
    \bottomrule
\end{tabular}
}
\label{tab:ll3da_leo}
\end{table}

\paragraph{3D-LLM \vs \agent.} As LL3DA collects a subset (ScanNet part) of 3D-LLM training data, we leverage this subset to pretrain \agent and compare the downstream performances with 3D-LLM. In contrast to the task-specific fine-tuning results of 3D-LLM, we report \agent's evaluation results after instruction tuning without task-specific fine-tuning. The results in \cref{tab:3dllm_leo} show that \agent consistently outperforms 3D-LLM when adopting the same training data. Notably, the magnitude of this subset is much smaller than their original training data, which further underscores the efficiency of our model.

\begin{table}[t!]
\centering
\captionof{table}{Quantitative comparison between 3D-LLM and \agent when both trained on 3D-LLM data. Metrics follow \cref{tab:vl_results}.}
\resizebox{0.53\linewidth}{!}{
\begin{tabular}{lcccccc}
    \toprule
     & \multicolumn{5}{c}{ScanQA (val)} & SQA3D (test) \\
     \cmidrule(lr){2-6} \cmidrule(lr){7-7}
     & C & B-4 & M & R & EM@1 & EM@1 \\
    \midrule

    3D-LLM & 74.5 & 12.9 & 15.1 & 37.5 & 21.2 & 49.8 \\

    \agent & \textbf{97.4} & \textbf{14.6} & \textbf{19.1} & \textbf{46.8} & \textbf{23.2 \textcolor{gray}{(45.4)}} & \textbf{50.6 \textcolor{gray}{(52.9)}} \\
    \bottomrule
\end{tabular}
}
\label{tab:3dllm_leo}
\end{table}

\subsection{Embodied Acting}\label{sec:result_objnav_additional}

\textbf{Quantitative results of \texttt{ObjNav}.} We provide additional results of \agent 1) generalizing to unseen objects on MP3D (below is a list of the objects used during training ({\color{mygreen}{seen}}) and for OOD evaluation ({\color{myred}{unseen}})), 2) learning with 70K human demonstrations provided by Habitat-web~\citep{ramrakhya2022habitat} instead of shortest path, and 3) learning without one modality (full vs. w/o 3D vs. w/o 2D). Evaluation results are shown in \cref{tab:result_objnav_ood_human}. Note that the baseline Habitat-web is unable to generalize to novel objects as it uses categorical embedding rather than natural language to represent object goals.
\begin{tcolorbox}
\begin{minipage}{\linewidth}
\# \texttt{Objects ({\color{mygreen}{seen}})}\\ \texttt{``gym\_equipment'', ``tv\_monitor'', ``picture'', ``counter'', ``chair'', ``cabinet'', ``table'', ``stool'', ``plant'', ``towel'', ``sofa'', ``cushion'', ``sink'', ``fireplace'', ``toilet'', ``seating'', ``chest\_of\_drawers'', ``bed'', ``shower'', ``bathtub'', ``clothes''} \\
\\
\# \texttt{Objects ({\color{myred}{unseen}})}\\ \texttt{``shelf'', ``pillow'', ``lamp'', ``box'', ``desk'', ``refrigerator'', ``vase'', ``armchair''} 
\end{minipage}
\end{tcolorbox}

\begin{table}[t!]
\centering
\caption{\textbf{Results on object navigation with OOD objects and human demonstrations.} Note that the baseline Habitat-web is unable to generalize to MP3D-{\color{myred}{unseen}} as it uses categorical embedding rather than natural language to represent object goals.}
\small
\setlength\tabcolsep{2pt}
\begin{tabular}{llccccc}
\toprule
 \multicolumn{2}{c}{\multirow{2}{*}{}} & \multicolumn{2}{c}{MP3D-{\color{mygreen}{seen}}} & & \multicolumn{2}{c}{MP3D-{\color{myred}{unseen}}} \\ \cmidrule(lr){3-4}\cmidrule(lr){6-7}
\multicolumn{2}{l}{} &  \small{Success$(\uparrow)$} & \small{SPL$(\uparrow)$} & & \small{Success$(\uparrow)$} &  \small{SPL$(\uparrow)$}\\ \midrule
\multicolumn{2}{l}{Habitat-web (shortest)} & 4.4 & 2.2 & & - &  - \\
\multicolumn{2}{l}{Habitat-web (70k demo)} & \textbf{35.4} & 10.2 & & - & - \\
\midrule
\multicolumn{2}{l}{\agent (shortest, w/o 2D)} & 7.8 & 4.6 & & - & - \\
\multicolumn{2}{l}{\agent (shortest, w/o 3D)} & 8.6 & 6.8 & & - & - \\
\multicolumn{2}{l}{\agent (shortest)} & 23.1 & \textbf{15.2} & & \textbf{11.1} &  \textbf{9.6} \\
\multicolumn{2}{l}{\agent (70k demo)} & 7.1 & 5.3 & & 8.9 & 8.6 \\
\bottomrule
\end{tabular}
\label{tab:result_objnav_ood_human}
\end{table}

The results show that \agent can generalize to novel objects. On the other hand, human demonstrations include more explorations, compared with shortest-path data. Therefore, it will be much harder for agents without a recurrent module (\eg, \agent) to learn from human demonstrations (see \cref{sec:supp_eai_split}), leading to significantly weaker performances. Finally, more modalities (2D + 3D) are indeed crucial to the success of mastering object navigation skills.

\textbf{Qualitative results}.~~We provide more qualitative results of robotic manipulation and embodied navigation in the supplementary video.

\subsection{Scan2Cap}
We provide additional qualitative results on Scan2Cap validation set in \cref{tab:Scan2Cap_add}. The results show that \agent can correctly refer to the queried object and provide accurate descriptions, including spatial relationships with other objects. However, \agent's responses are confined to simple formats that lack diversity. How to unlock more flexible responses while maintaining accuracy can be a direction for future research. 

\begin{table}[t]
    \centering
    \caption{\textbf{Additional qualitative results on Scan2Cap validation set}.}
    \small
    \begin{tabular}{c|p{8em}|p{8em}|p{10em}}
    \toprule
         Scene  & Instruction & Response & Ground Truth  \\
         \midrule
            \raisebox{-1\height}
            {\includegraphics[width=0.2\textwidth]{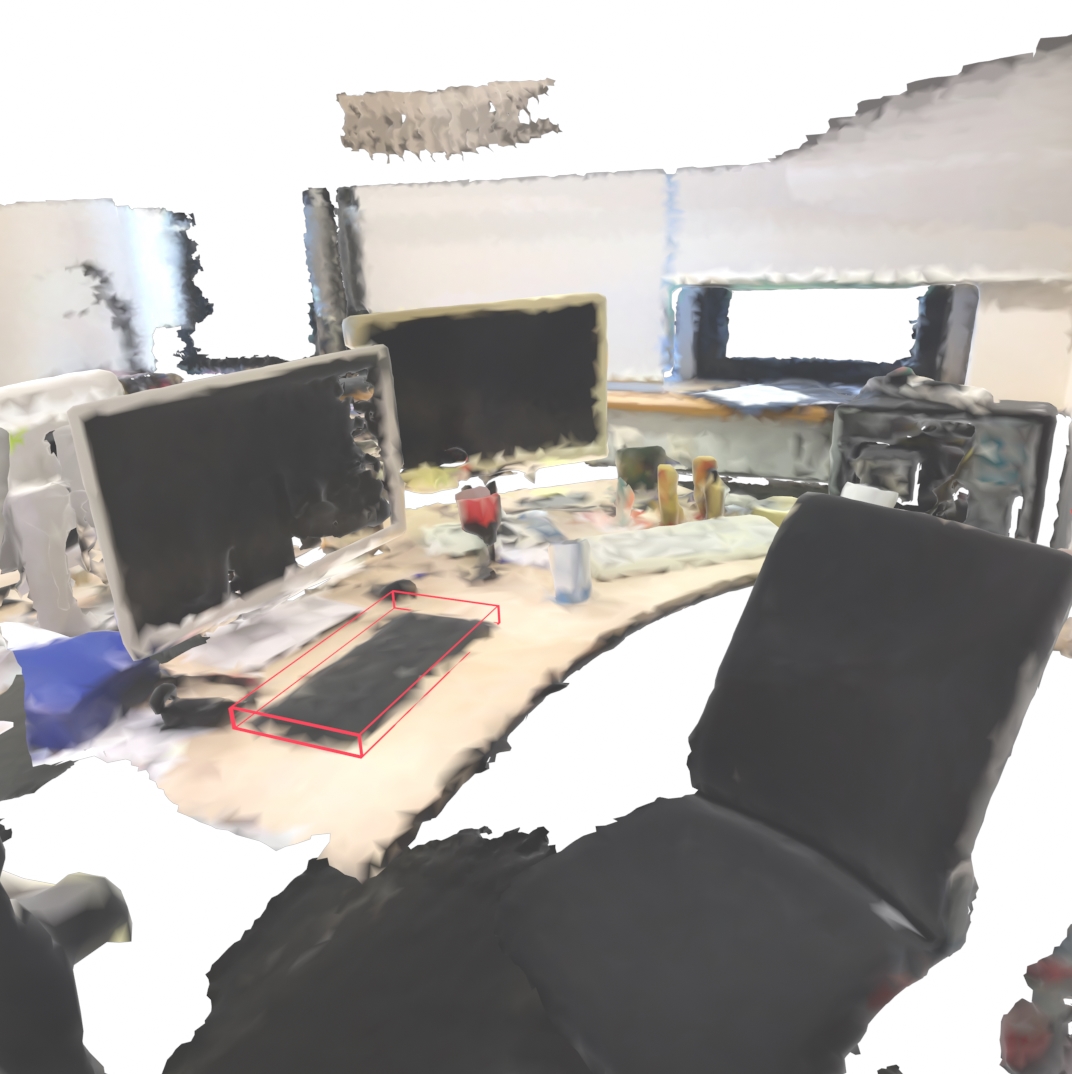}} & 
            Develop a description of the object at the specified position in the 3D backdrop. & This is a black keyboard. It is in front of two monitors. &
            The keyboard sits atop the workstation desk. It is located in front of the left-side monitor.\\
         \midrule
            \raisebox{-1\height}
            {\includegraphics[width=0.2\textwidth, ]{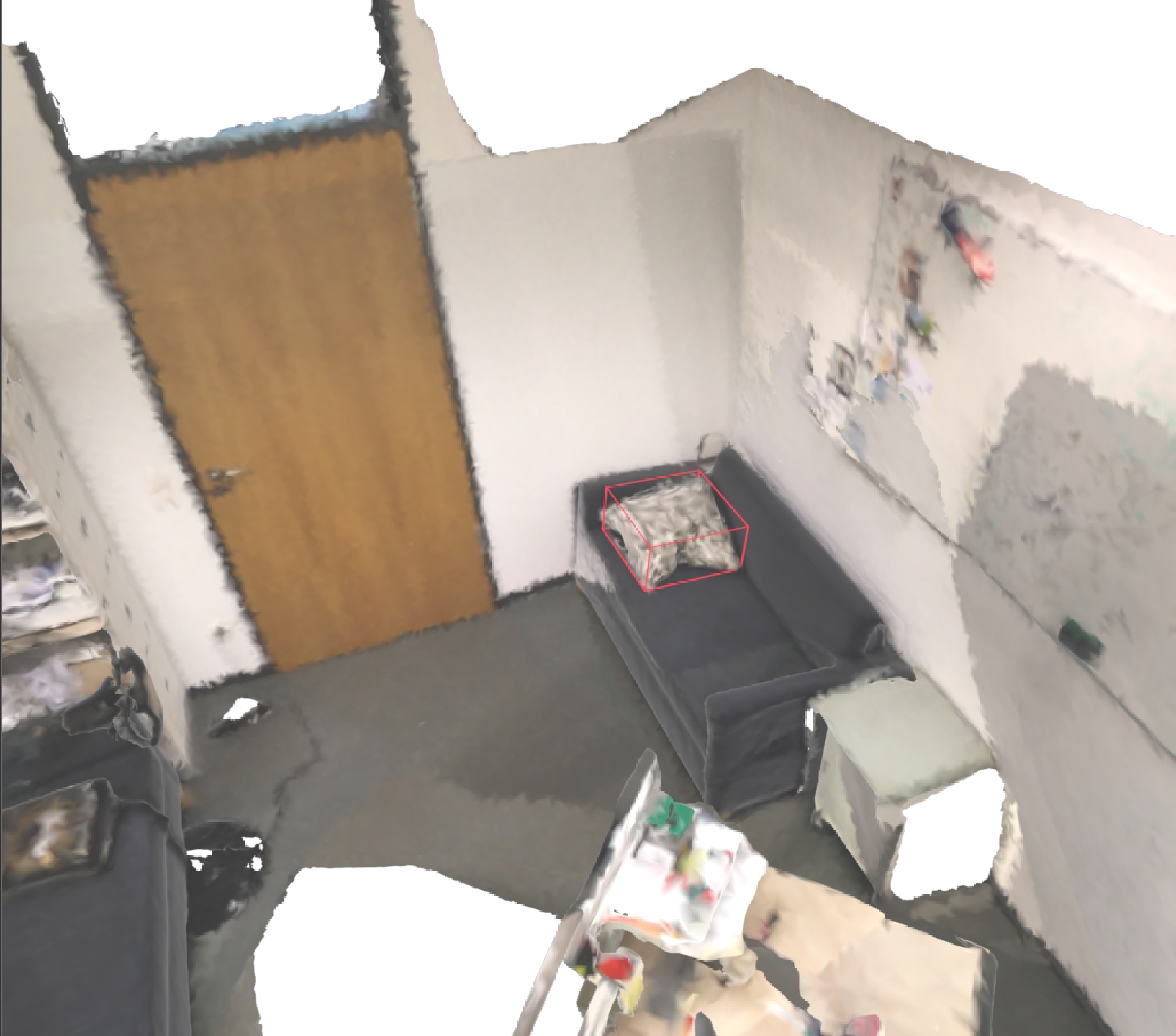}} &
            Focus on the object at the given 3D location and describe it. & This is a white pillow. It is on the couch. & It is a small pillow located on the couch.  You can notice it directly on your left when walking through the door into the room. \\
         \midrule
         \raisebox{-1\height}
         {\includegraphics[width=0.2\textwidth, ]{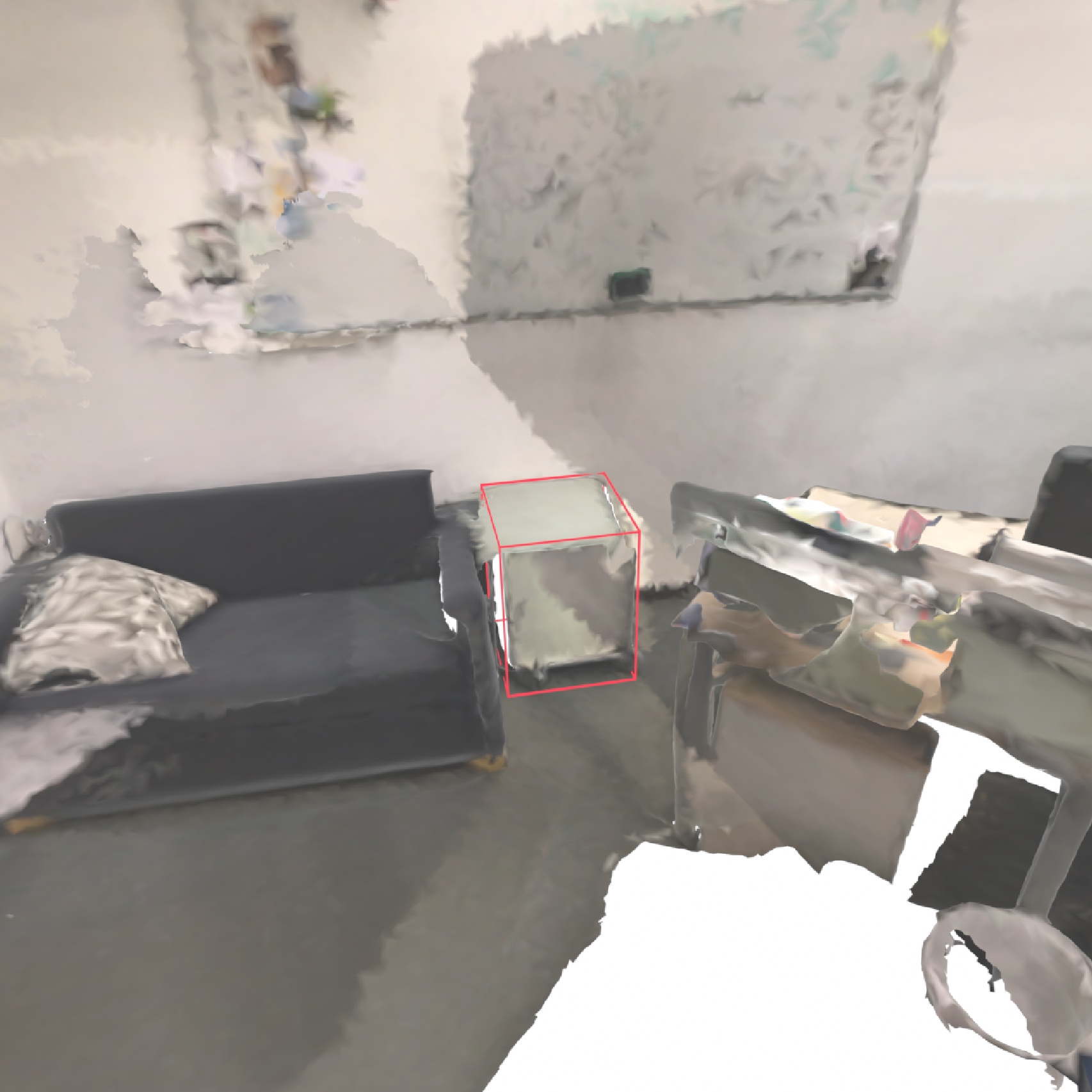}} & Provide a clearer description of the object at that specific 3D spot. &  This is a white cabinet. It is to the right of a couch. &
         This is a white ottoman. It is located to the right of the black couch. \\
        \midrule
            \raisebox{-1\height}
            {\includegraphics[width=0.2\textwidth]{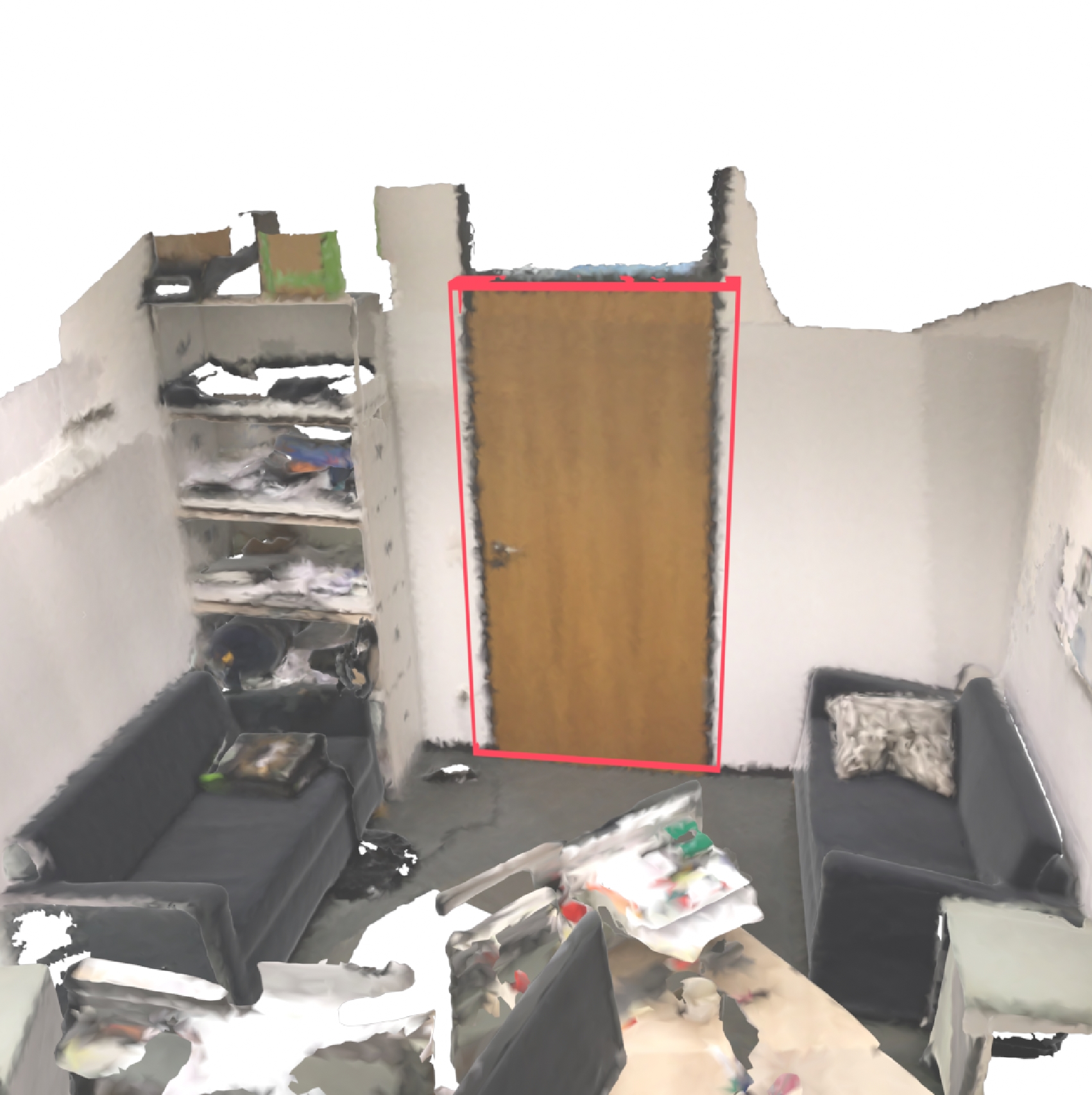}} &
            What does the object at the specified position in the 3D visualization look like? & There is a rectangular brown door. It is next to a couch. & There is a door in the middle of the northern wall next to a metal cabinet. The door is shaped like a brown rectangle and as a dark yellow colored doorknob. \\
        \bottomrule
\end{tabular}

\label{tab:Scan2Cap_add}
\end{table}

\subsection{ScanQA}

We provide additional qualitative results on ScanQA validation set in \cref{tab:ScanQA_add} and categorize the responses into several types:
\begin{itemize}[leftmargin=*]
    \item \textit{Wrong}. The response is inaccurate and deemed wrong.
    \item \textit{Wrong but reasonable}. The response is deemed wrong but is reasonable to some extent, probably due to ambiguities in the scene. Consider the second case in \cref{tab:ScanQA_add}. There are many objects such as a coat rack, a coat, and a mini fridge-shaped cabinet on the right side of the organizer. Though \agent's response ``mini fridge'' does not match the ground truth ``coat rack'', it is consistent with the 3D scene layout.
    \item \textit{Wrong but accurate}. The response is accurate according to the scene but is deemed wrong due to imperfect ground truth annotations.
    \item \textit{Correct}. The response is accurate and deemed correct.
    \item \textit{Correct and more accurate}. The response is more accurate than the ground truth annotations.
\end{itemize}

\begin{table}[t]
    \centering
    \caption{\textbf{Additional qualitative results on ScanQA validation set}.}
    \small
    \begin{tabular}{c|p{6em}|p{6em}|p{8em}|p{6em}}
    \toprule
         Scene  & Question & Response & Ground Truth & Response Type \\
         
         \midrule
         \raisebox{-1\height}
         {\includegraphics[width=0.2\textwidth, ]{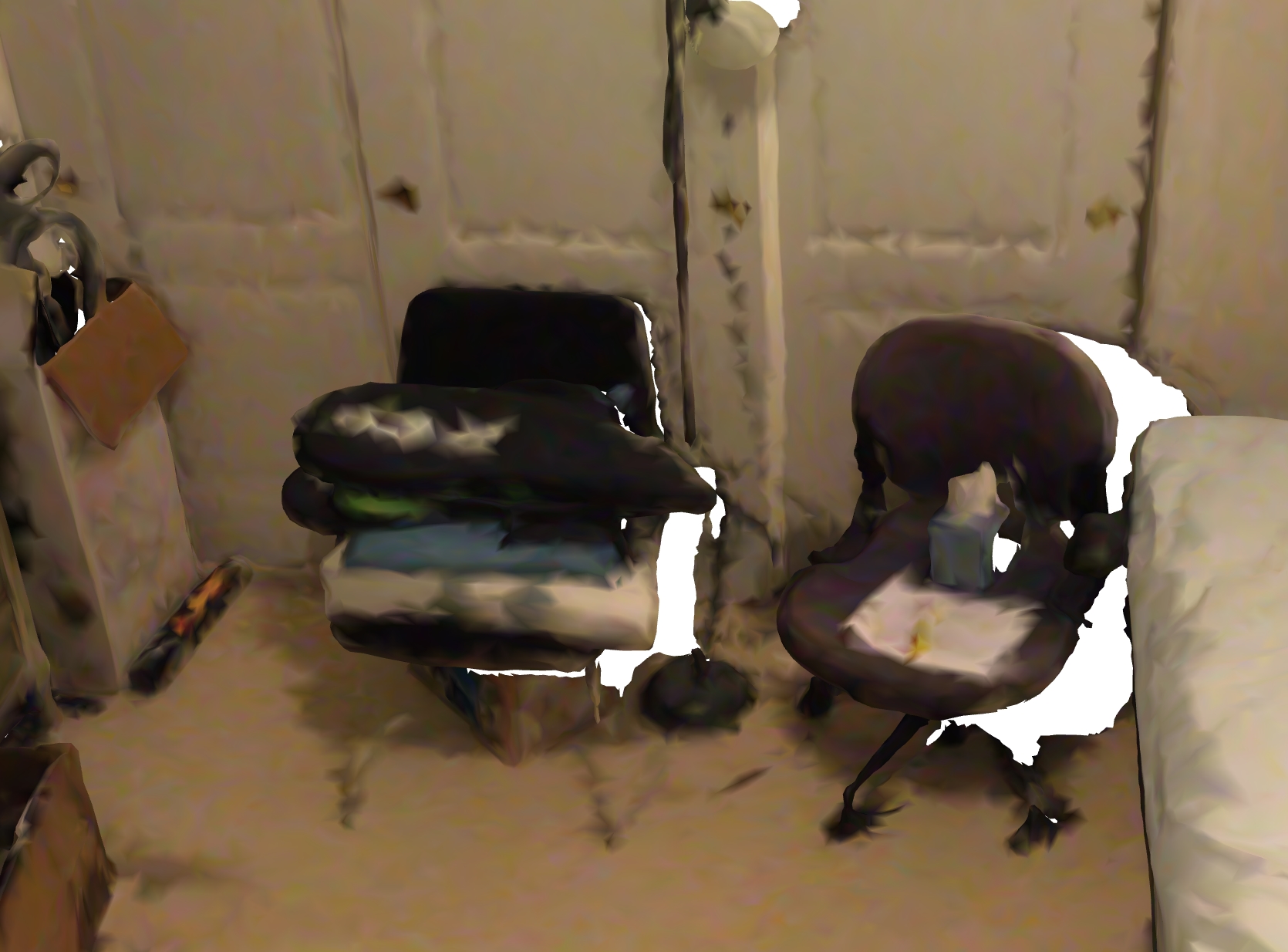}} & What is placed on the black chair out of the three chairs present in the room? & pillow & tan jacket, guitar box like object & \textit{Wrong} \\
         
         \midrule
         \raisebox{-1\height}
         {\includegraphics[width=0.2\textwidth, ]{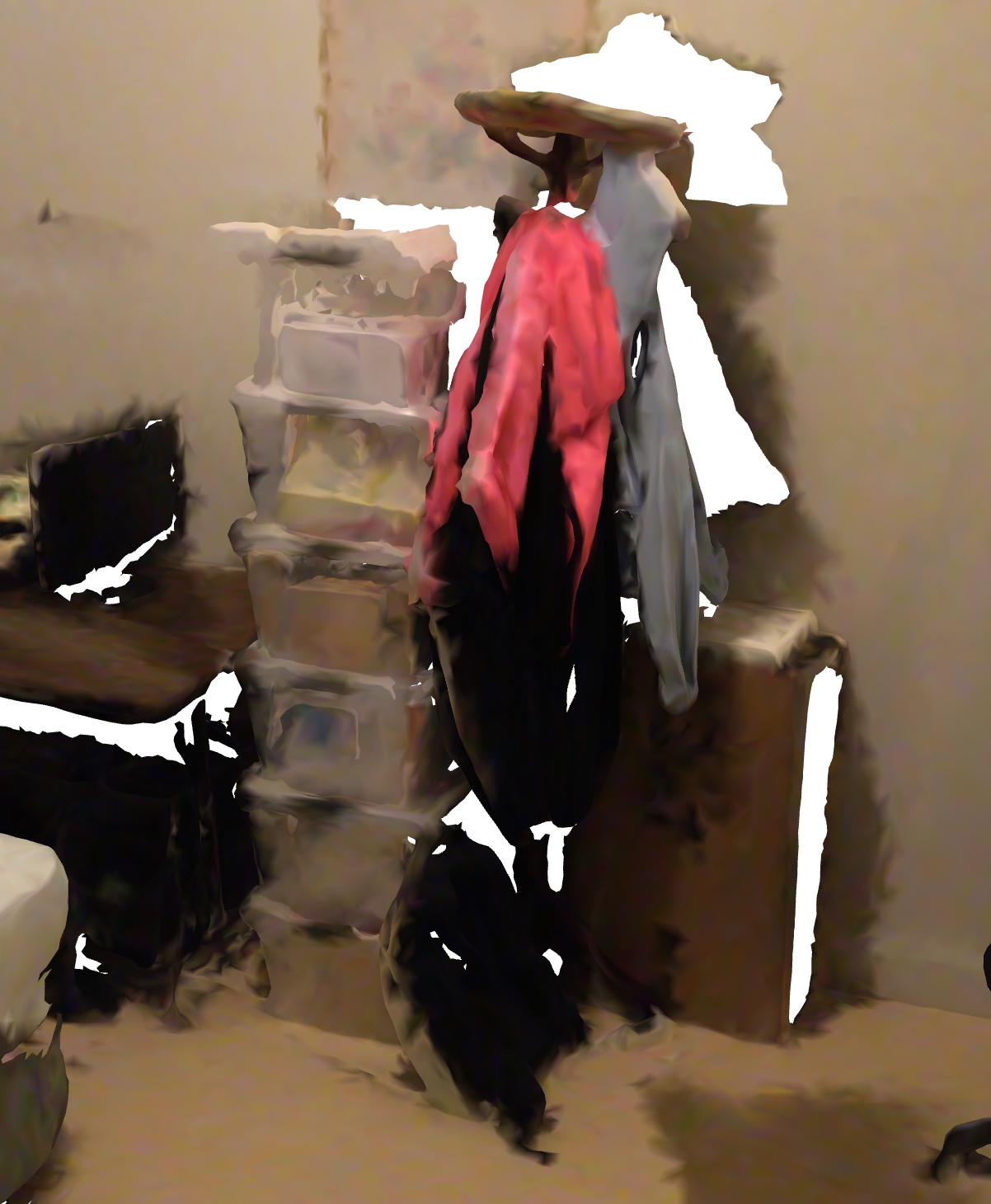}} & What is next to the organizer shelf? & mini fridge & coat rack & \textit{Wrong but reasonable} \\
         
         \midrule
         \raisebox{-1\height}
         {\includegraphics[width=0.2\textwidth, ]{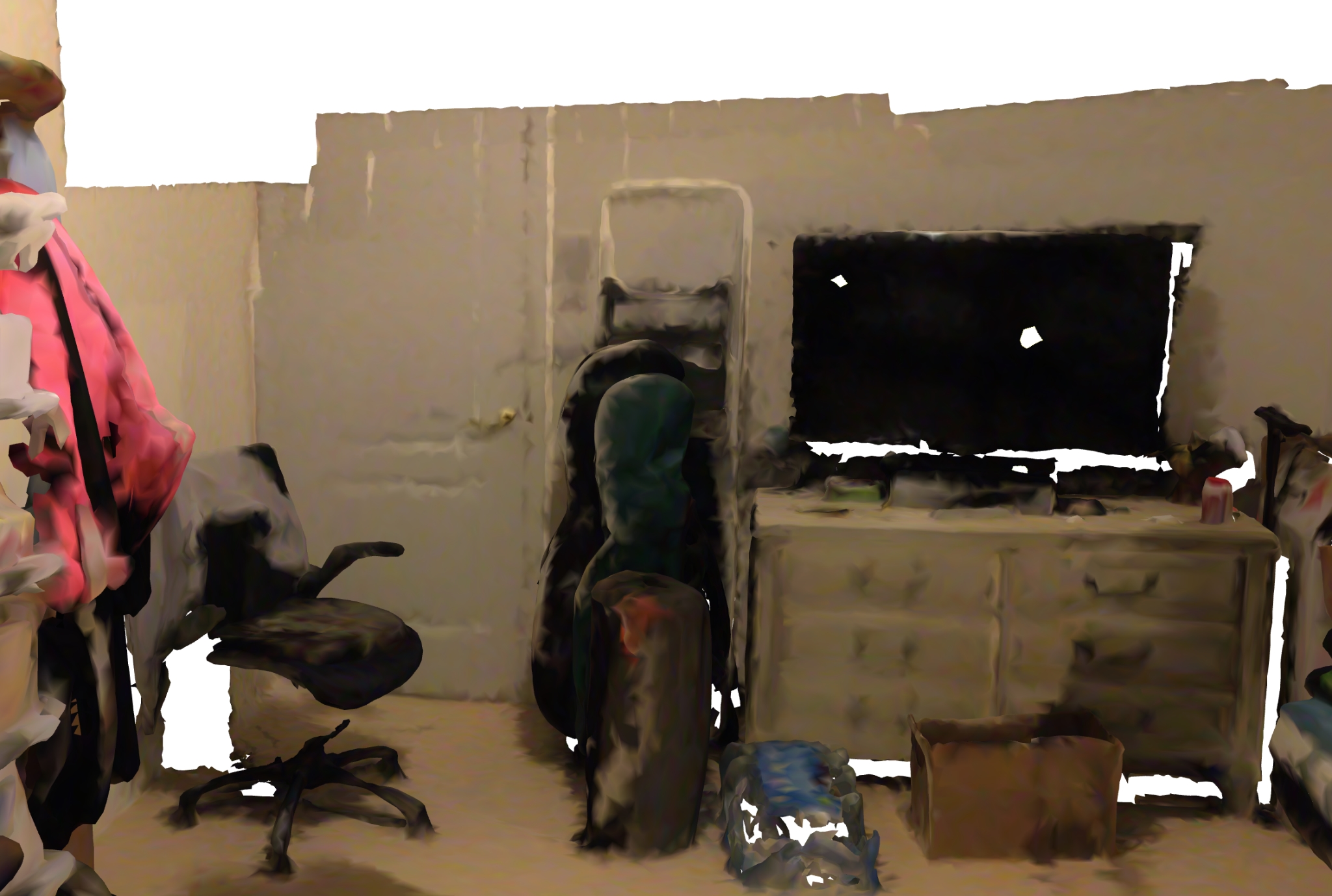}} & Where is the dresser located? &  to right of door &
         underneath television next to black guitar case and green guitar case, under tv set next to guitar cases & \textit{Wrong but accurate} \\

         \midrule
         \raisebox{-1\height}
         {\includegraphics[width=0.2\textwidth, ]{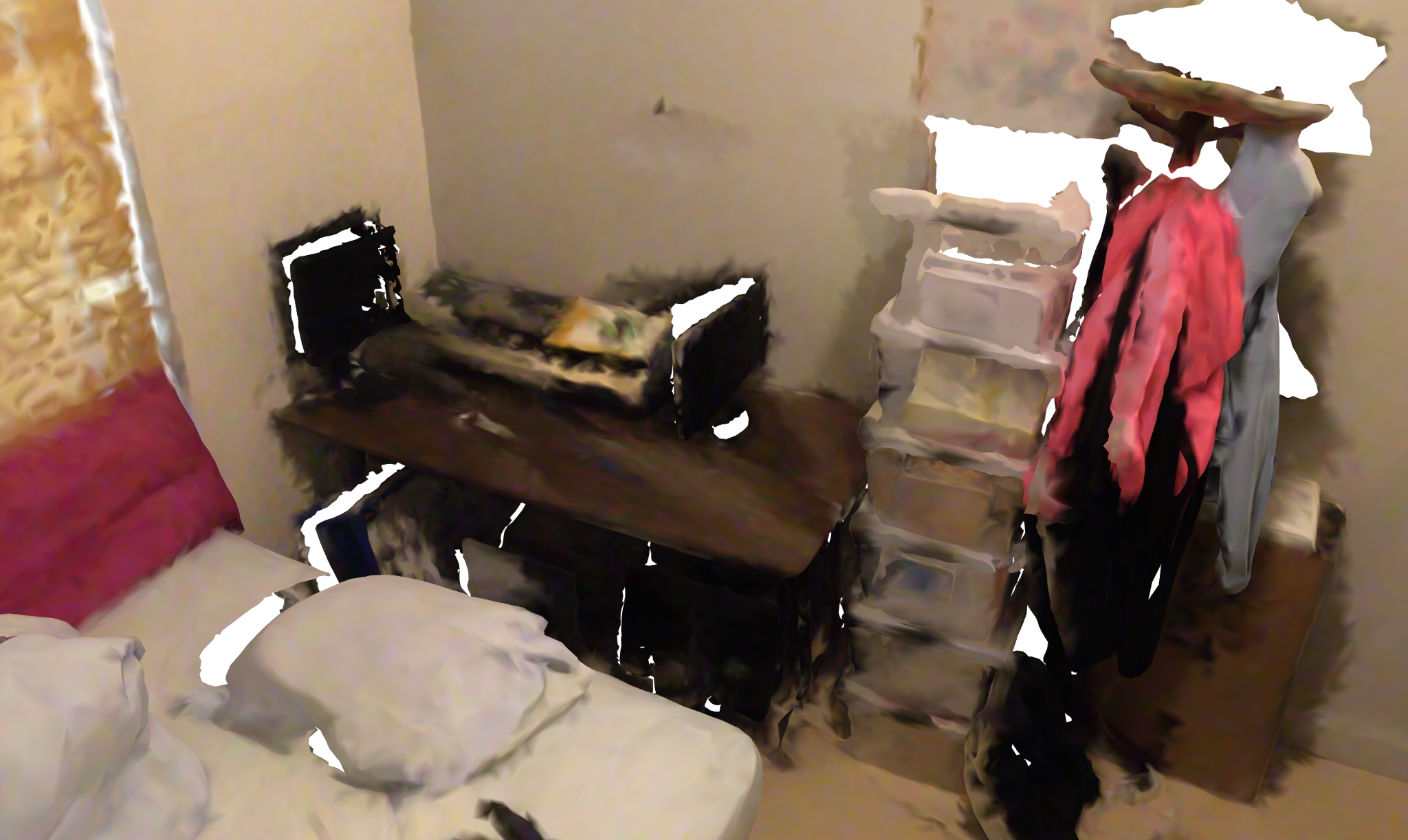}} &
         What kind of desk is the electric piano on? &
         wooden desk & wooden desk & \textit{Correct} \\
        
         \midrule
         \raisebox{-1\height}
         {\includegraphics[width=0.2\textwidth]{figs/vis_scanqa_4.jpg}} &
         How many monitor sits on top of the desk? & 2 & 2 & \textit{Correct} \\
         
         \midrule
            \raisebox{-1\height}
            {\includegraphics[width=0.2\textwidth]{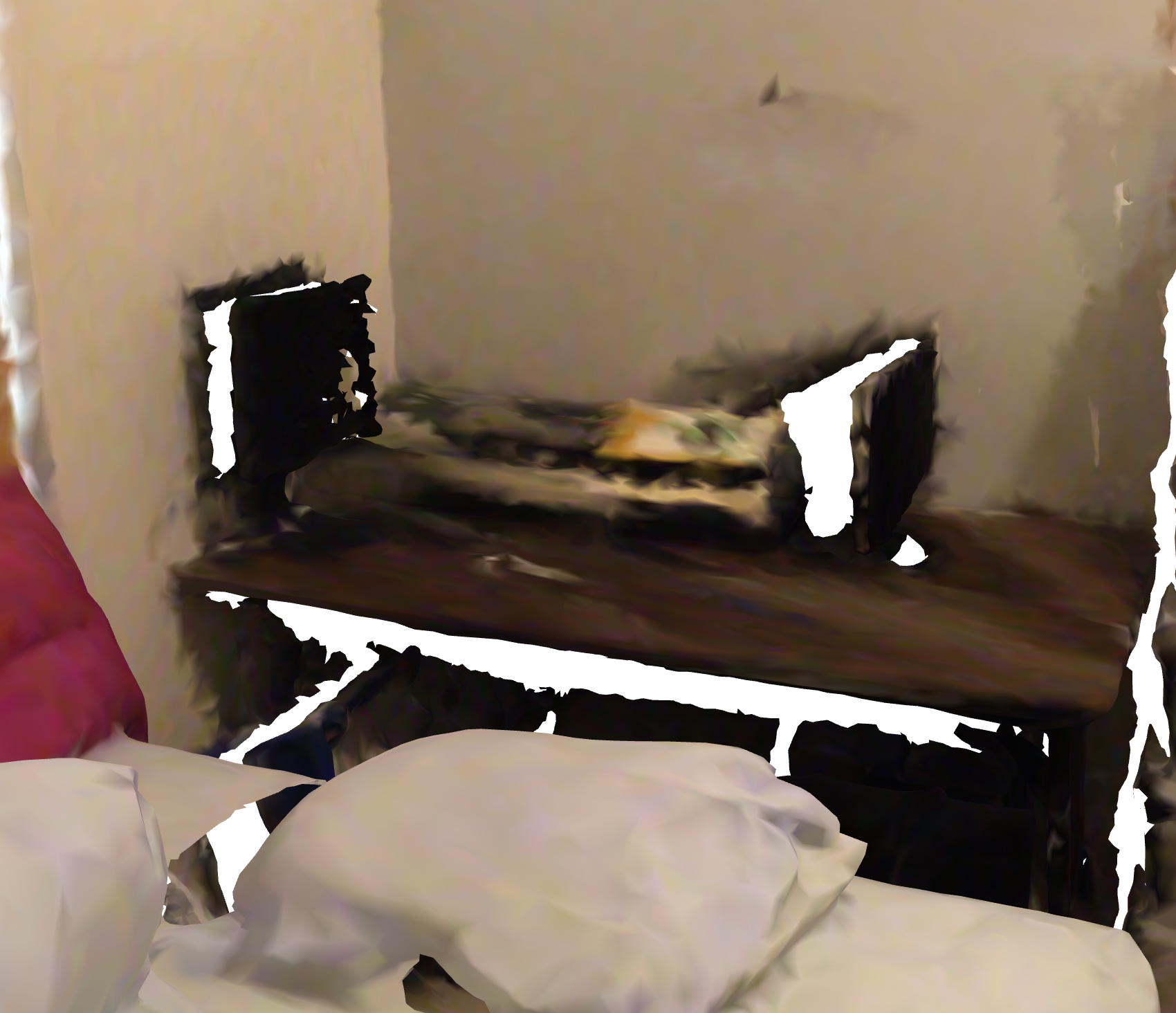}} & 
            What is on the left side of the brown desk? &
            black computer monitor & monitor / computer monitor &
            \textit{Correct and more accurate} \\
        \bottomrule
\end{tabular}

\label{tab:ScanQA_add}
\end{table}

\subsection{SQA3D}
We provide additional qualitative results on SQA3D test set in \cref{tab:sqa_add} and follow the aforementioned response types. The embodied reasoning in SQA3D requires the understanding of not only the scene but also the situation of embodiment. In \cref{tab:sqa_add}, answering ``What am I sitting at?'' necessitates that \agent accurately identifies the objects at its current location. And the response to ``How many beds are in front of me?'' indicates that \agent can reason based on the understanding of its orientation.

\begin{table}[t!]
    \centering
    \caption{\textbf{Additional qualitative results on SQA3D test set}.}
    \small
    \begin{tabular}{c|p{6em}|p{6em}|p{8em}|p{6em}}
        \toprule
         Scene & Question & Response & Ground Truth & Response Type \\
         \midrule
         \raisebox{-1\height}
         {\includegraphics[width=0.2\textwidth]{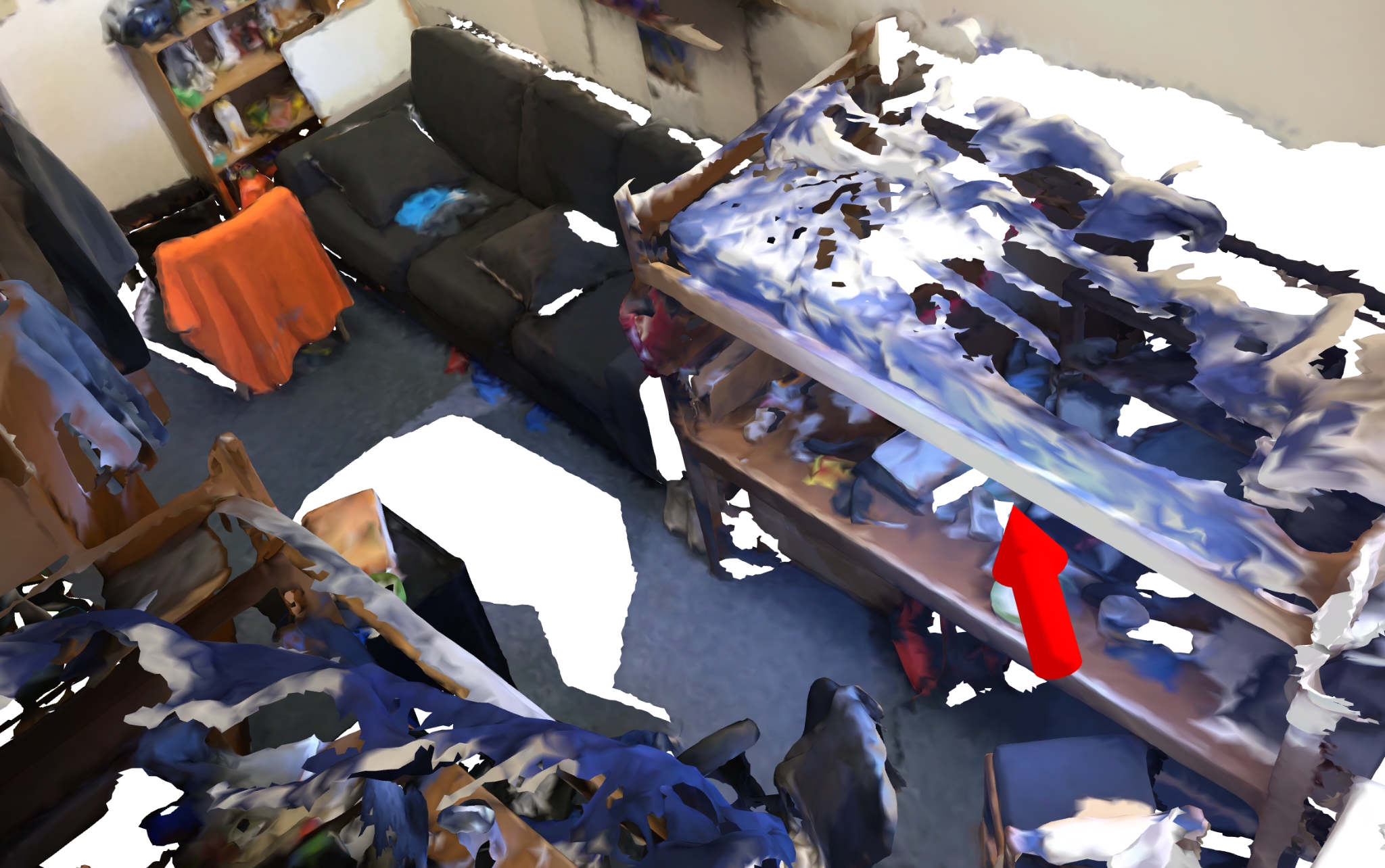}} & What is the only orange object on my left side? & backpack & jacket & \textit{Wrong} \\
         \midrule
         \raisebox{-1\height}
         {\includegraphics[width=0.2\textwidth]{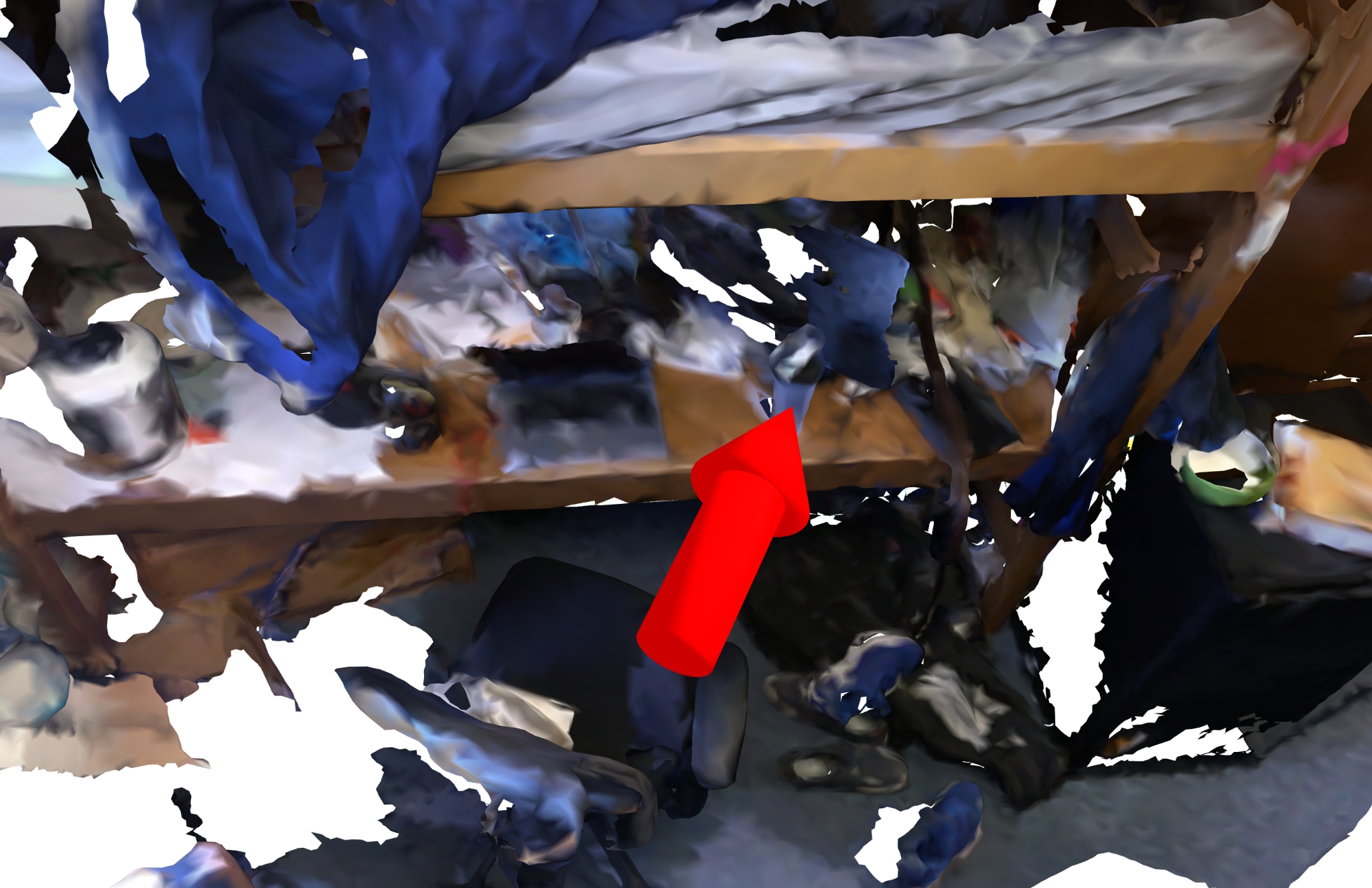}} & Is there anything on the desk? & yes & laptop & \textit{Wrong but accurate} \\
         \midrule
         \raisebox{-1\height}
         {\includegraphics[width=0.2\textwidth,]{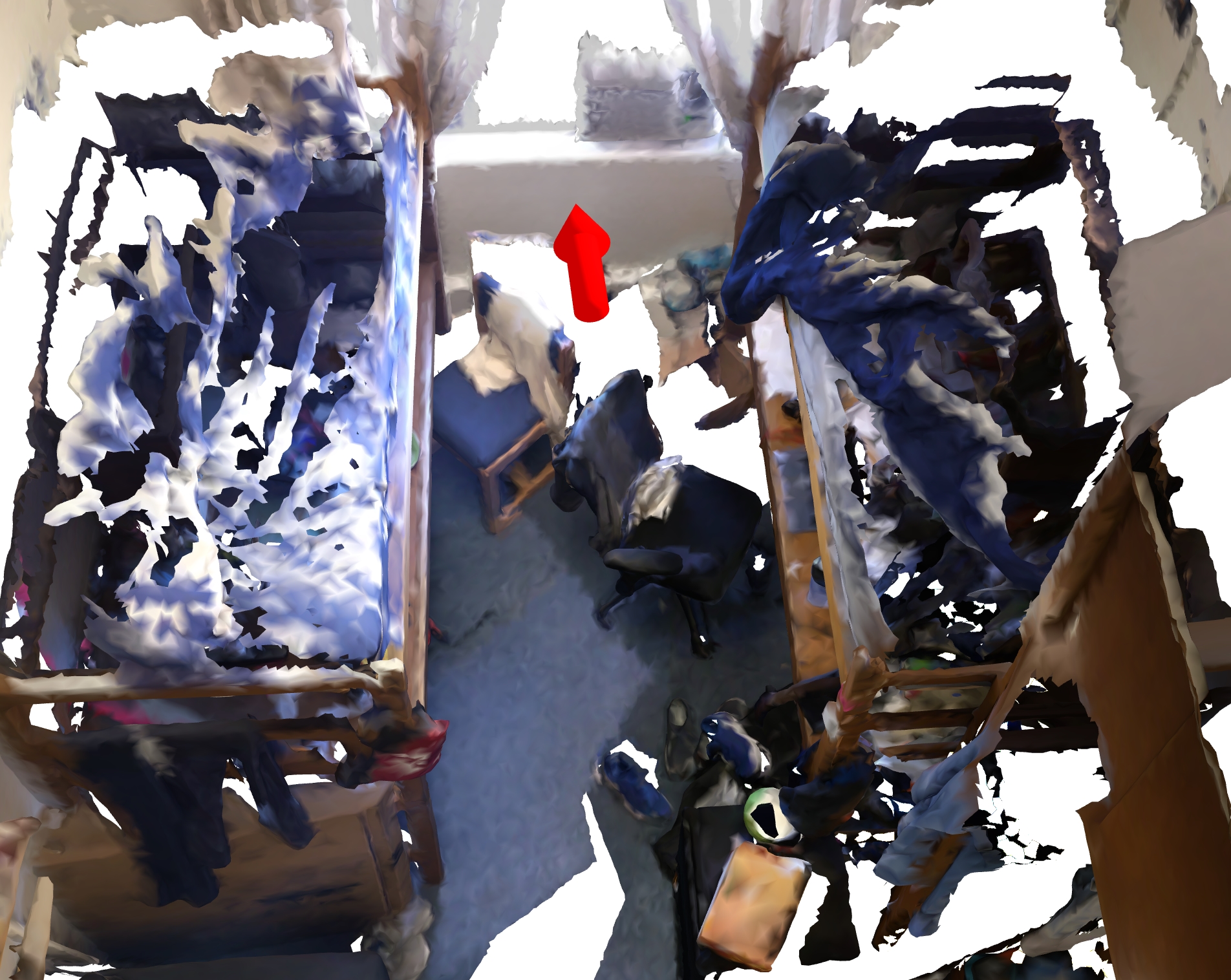}} & What type of bed is the one on the right? & loft bed & loft bed & \textit{Correct} \\
         \midrule
         \raisebox{-1\height}
         {\includegraphics[width=0.2\textwidth]{figs/vis_sqa_4.jpg}} & What am I sitting at? & desk & desk & \textit{Correct} \\
         \midrule
         \raisebox{-1\height}
         {\includegraphics[width=0.2\textwidth]{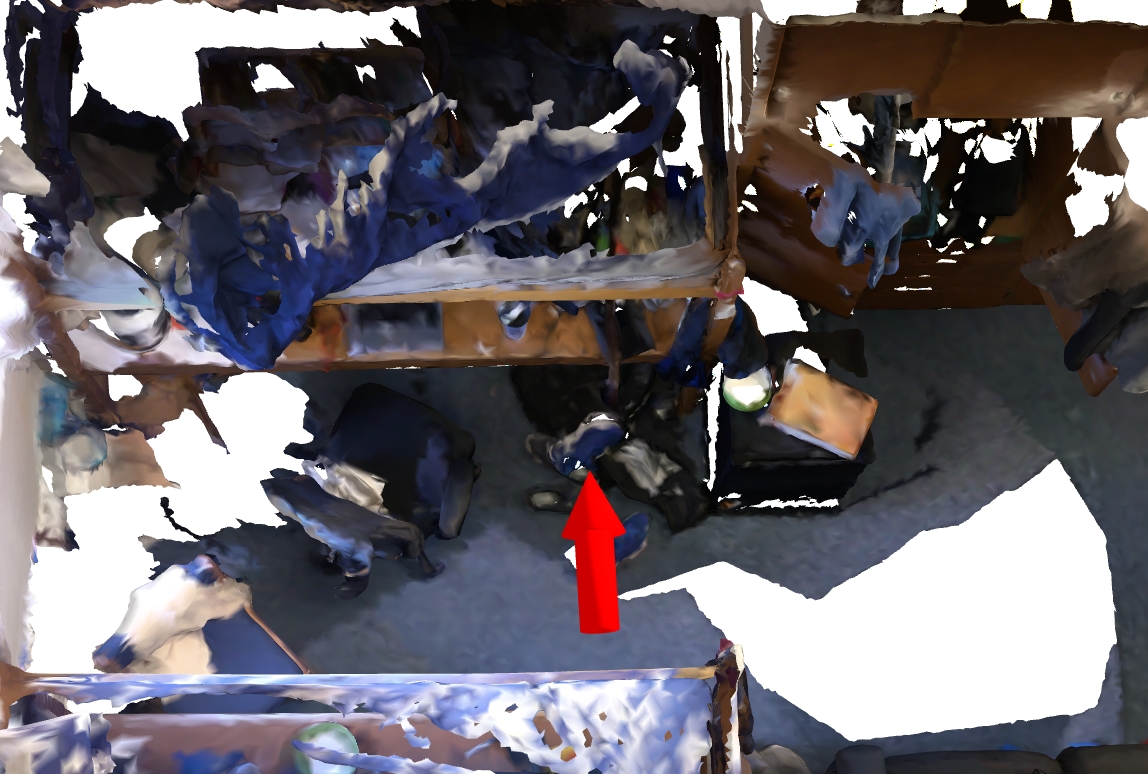}} & How many beds are in front of me? & one & one & \textit{Correct} \\
         \bottomrule
    \end{tabular}
    
    \label{tab:sqa_add}
\end{table}

\begin{table}[t!]
    \centering
    \caption{\textbf{Examples from our datasets}. We provide the tasks, 2D images (optional), 3D point clouds, and instructions as the input of \agent, as well as the expected responses of \agent.}
    \small
    \begin{tabular}{c|c|p{6em}|p{12em}}
    \toprule
         Dataset \& Task & 2D \& 3D input & Instruction & Response  \\
         \midrule
         \makecell{\agent-align\\object captioning} & \raisebox{-1\height}{\includegraphics[width=0.2\textwidth, ]{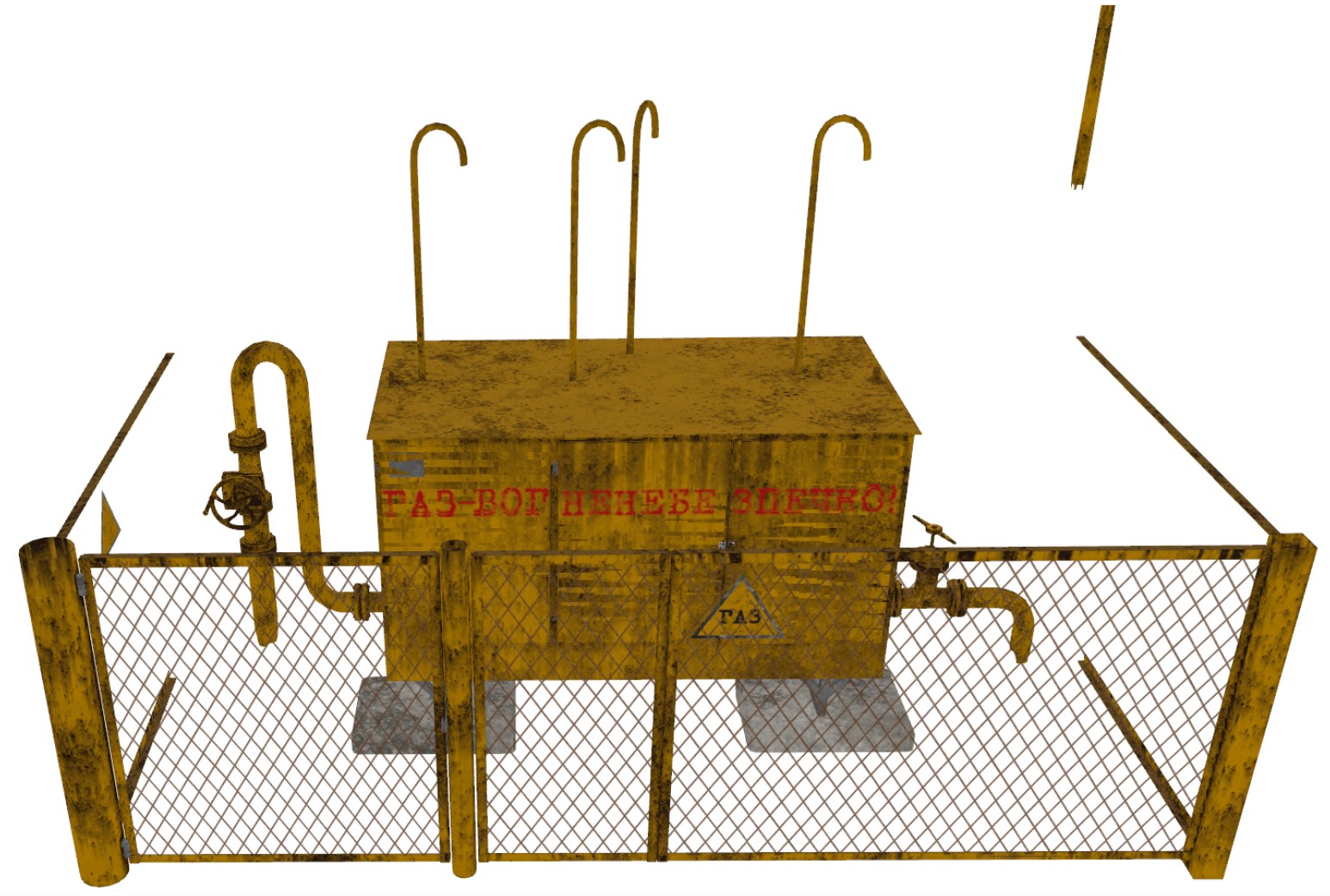}} & USER: Explain the object at the pinpointed location in the 3D setting. ASSISTANT: & a yellow metal cage with a fence around it. \\
         \midrule
         \makecell{\agent-align\\object referring} & \raisebox{-1\height}{\includegraphics[width=0.2\textwidth, ]{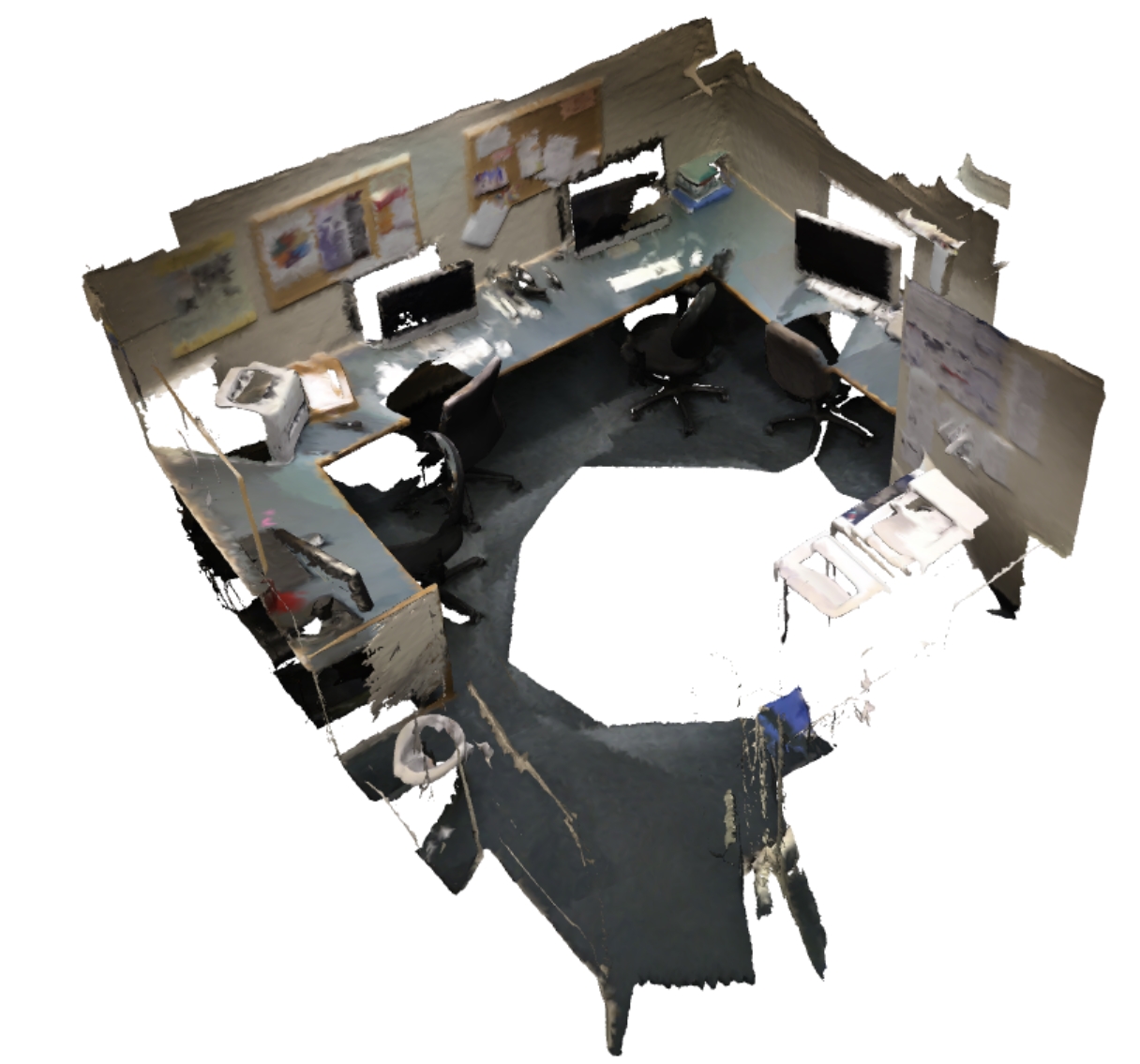}}& USER: Give me a snapshot of the object at the chosen 3D location. ASSISTANT:& find the office chair that is near the copier\\
         \midrule
         \makecell{\agent-align\\scene captioning} & \raisebox{-1\height}{\includegraphics[width=0.2\textwidth, ]{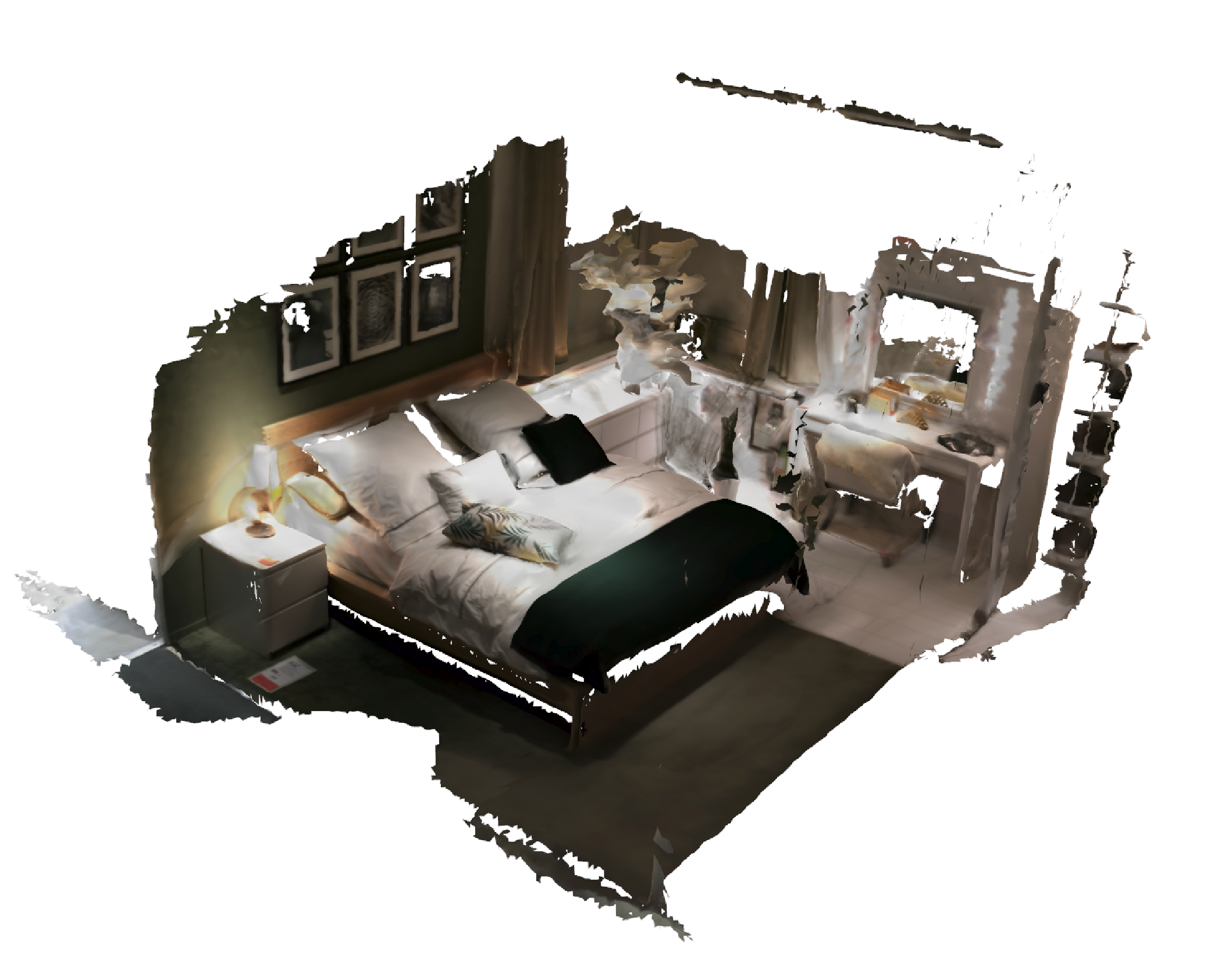}}& USER: Relay a brief summary of this scene's elements. ASSISTANT: &
         In this scene, there is a gray flat picture hanging on the wall, standing in front of a windowsill. There is also a rectangular white padded pillow lying on the bed, close to a big rectangular white padded pillow. The room has a white architectural wall and a flat overhead ceiling. The bed is tidy and white, located in front of a narrow tall commode. The curtains are hanging on the wall. The room has a calm and comfortable atmosphere, suitable for relaxation and sleep.\\ %
        \bottomrule
\end{tabular}

\label{tab:supp_data_example}
\end{table}

\begin{table}[t!]
\centering
\caption{\textbf{Examples from our datasets}. Continued.}
\small
\begin{tabular}{c|c|p{8em}|p{10em}}
\toprule     
         \makecell{\agent-instruct\\3D captioning} & \raisebox{-1\height}{\includegraphics[width=0.2\textwidth, ]{figs/scan2cap_door.jpg}}& USER: What does the object at the specified position in the 3D visualization look like? ASSISTANT:& there is a door in the middle of the northern wall next to a metal cabinet. the door is shaped like a brown rectangle and as a dark yellow colored doorknob.\\
         \midrule
         \makecell{\agent-instruct\\3D QA} & \raisebox{-1\height}{\includegraphics[width=0.2\textwidth, ]{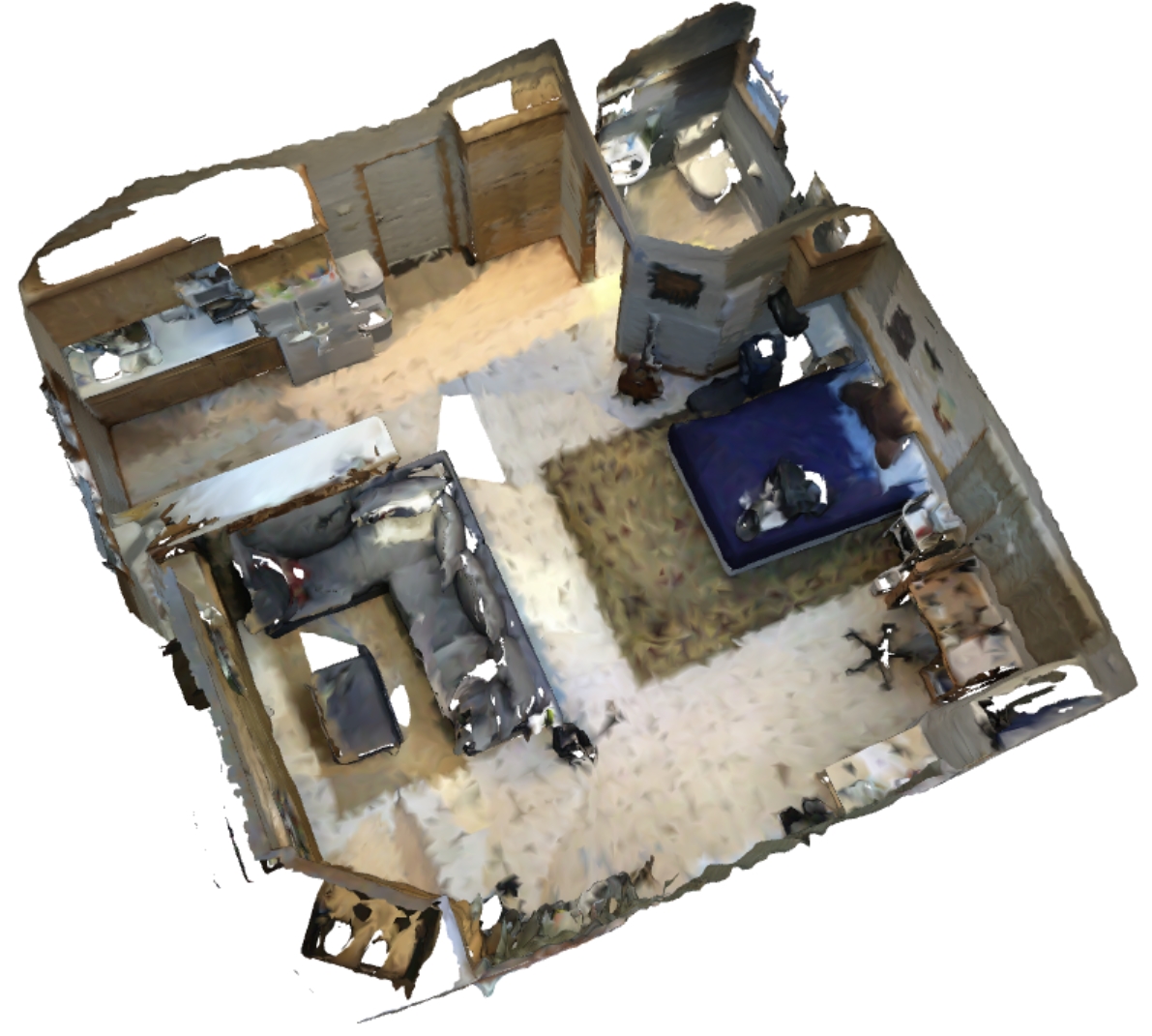}}& USER: What is in the right corner of room by curtains? ASSISTANT: & brown cabinet with tv sitting in it\\
         \midrule
         \makecell{\agent-instruct\\3D dialogue} &\raisebox{-1\height}{\includegraphics[width=0.2\textwidth, ]{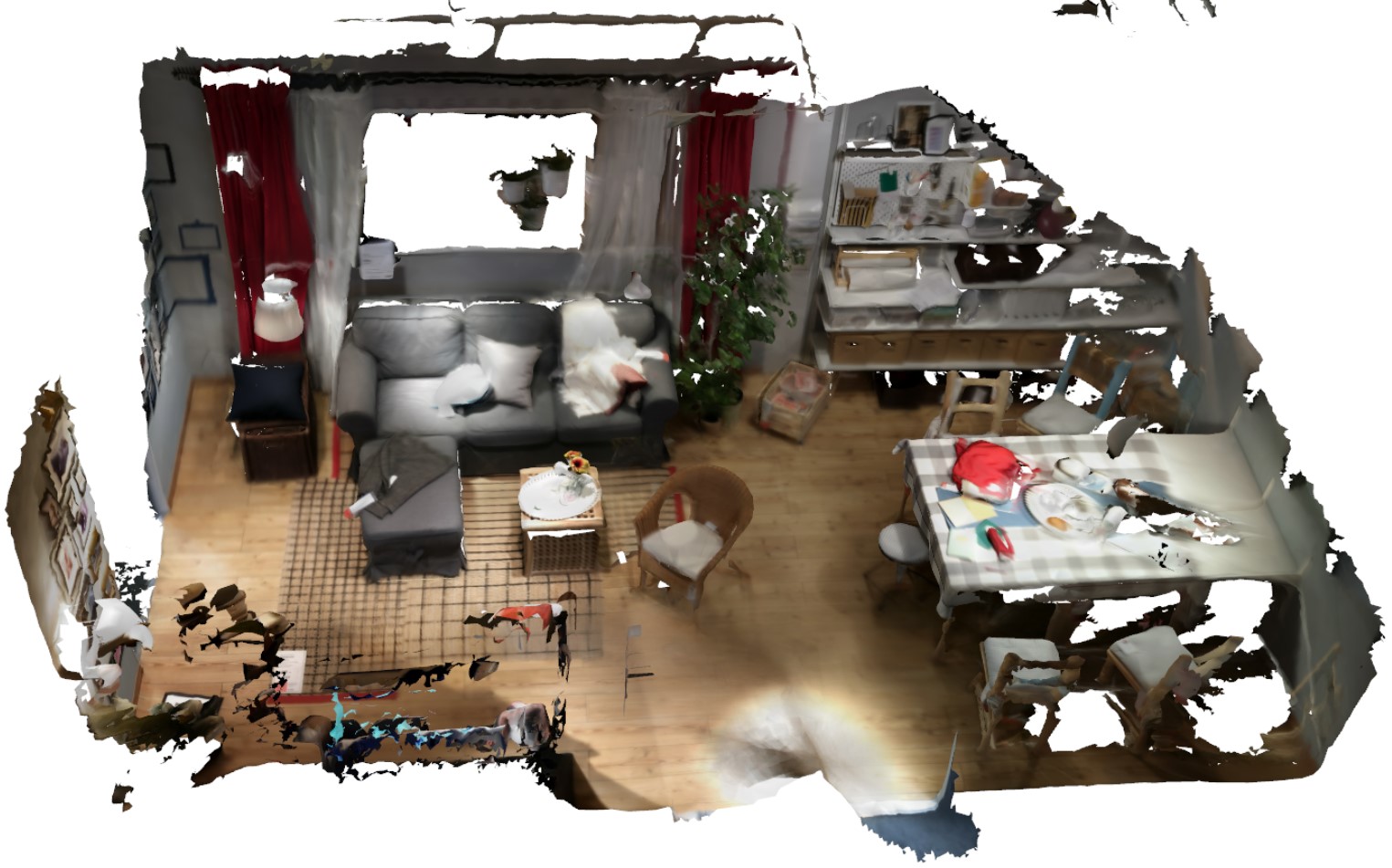}} & USER: I'm redecorating my living room. Can you give me some suggestions on how to arrange the furniture?
         
         ASSISTANT: Of course! I'd be happy to help. What furniture do you have in your living room?
         
         USER: I have a sofa, a table, a chair, and a couch table. ASSISTANT: &
         Great! Here's a suggestion: you can place the sofa against the wall, with the table in front of it. The chair can be placed to the right of the table, and the couch table can be placed to the right of the chair. This arrangement creates a cozy seating area. \\% 3rscan 95be45dd-a558-22da-9de4-002d61e13deb  
         \midrule
         \makecell{\agent-instruct\\task planning} & \raisebox{-1\height}{\includegraphics[width=0.2\textwidth, ]{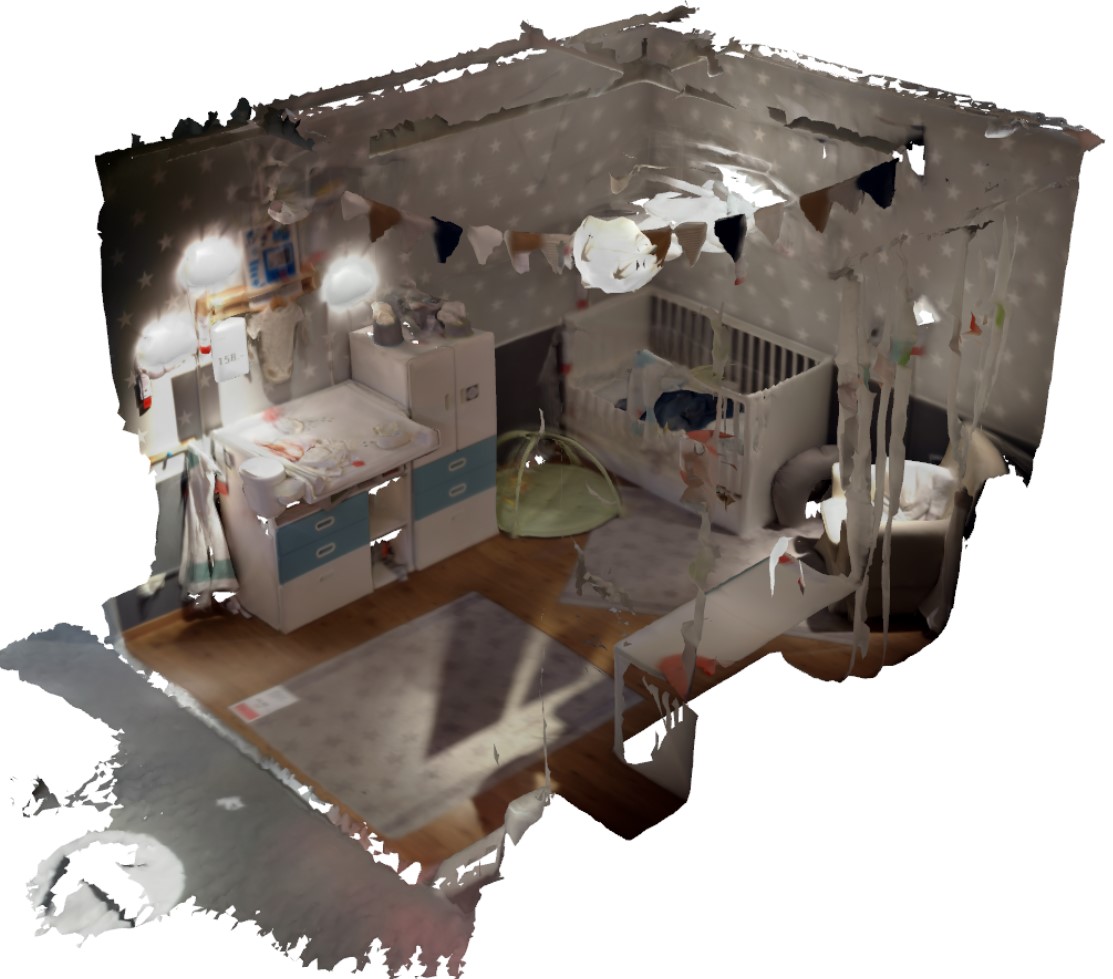}} & USER: Develop a plan for this: Arrange and embellish the family room. ASSISTANT: &
         1. Arrange and fluff the cushions and pillows on the armchair.
         
         2. Place decorations, such as figurines or vases, on the cabinet or shelf.
         
         3. Place hygiene products, such as wipes or lotions, on the commode.
         
         4. Turn on the lights and adjust their brightness if needed.
         
         5. Arrange diapers in a designated storage area.
         
         6. Place a blanket or pillow on the armchair for added comfort.
         
         7. Adjust the festoon or other decorative elements if desired.\\ %
    \bottomrule
    \end{tabular}
    
    \label{tab:supp_data_example_cont}
\end{table}

\begin{table}[t!]
\centering
\caption{\textbf{Examples from our datasets}. Continued.}
\small
\begin{tabular}{c|c|p{8em}|p{12em}}
\toprule     
         \makecell{\agent-instruct\\embodied navigation} &
         \raisebox{-1\height}{\makecell{\includegraphics[width=0.2\textwidth, ]{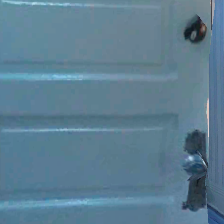} \\ \includegraphics[width=0.2\textwidth, ]{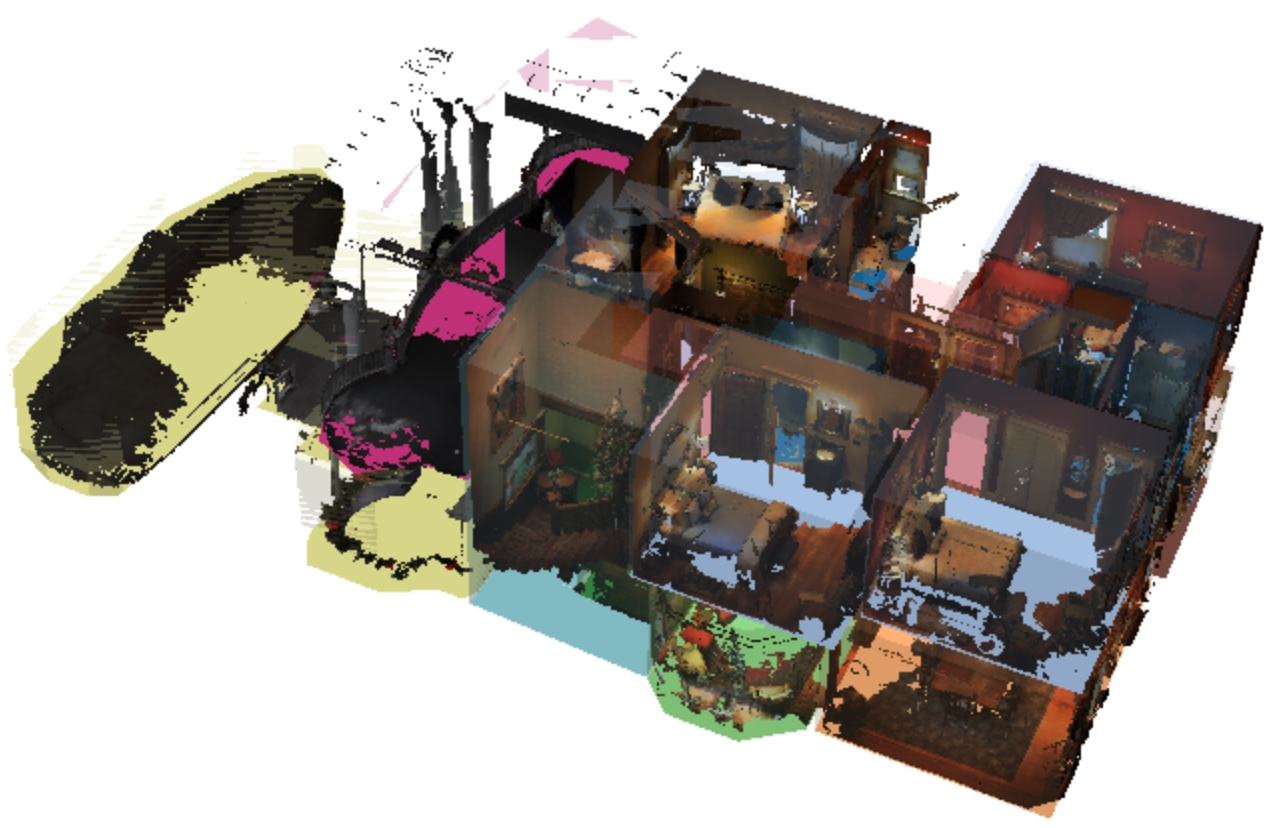}}}& USER: The task is navigation. Your goal is to find counter by moving around in the scene. Past actions: <31999> <31999> <31999> <31999>. ASSISTANT:& <31996>\\
         \midrule 
         \makecell{\agent-instruct\\robotic manipulation} & \raisebox{-1\height}{\makecell{\includegraphics[width=0.2\textwidth, ]{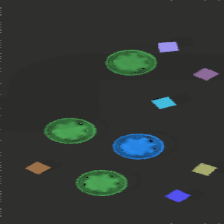}\\\includegraphics[width=0.2\textwidth, ]{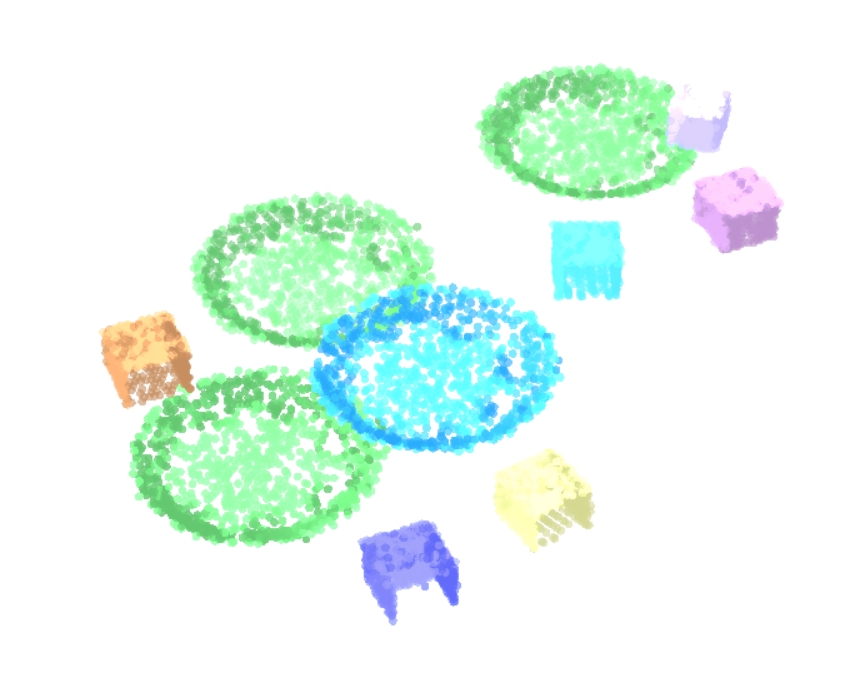}}}& USER: The task is manipulation. Your goal is to put the blue blocks in a green bowl. Past actions: <31991> <31671> <31511> <31991> <31671> <31511> <31991> <31671> <31511> <31991> <31671> <31511> <31991> <31671> <31511> <31991> <31671> <31511> <31991> <31671> <31511> <31991> <31671> <31511>. ASSISTANT:& <31748> <31644> <31511> <31736> <31595> <31500> \\     
    \bottomrule
    \end{tabular}
    
    \label{tab:supp_data_example_cont2}
\end{table}

\end{document}